%% file: Paper.tex
\documentclass{article}
\pdfoutput=1

\PassOptionsToPackage{square, numbers}{natbib}

\usepackage{Template/iclr2022_conference,times}

\usepackage{microtype}
\usepackage[utf8]{inputenc}
\usepackage[T1]{fontenc}    
\usepackage[hidelinks]{hyperref}
\usepackage{url}
\usepackage{rmclarke}
\usepackage{amsbsy}
\usepackage{pgfplots}
\pgfplotsset{compat=newest}
\usetikzlibrary{calc, arrows.meta, math, positioning}
\usepackage{graphicx}
\usepackage{caption}
\usepackage{subcaption}
\usepackage{booktabs}
\def\extrarulespace{-0.5\dimexpr \aboverulesep + \belowrulesep + \cmidrulewidth}
\usepackage{multirow}
\usepackage{makecell}
\usepackage{siunitx}
\usepackage{etoolbox}
\robustify\bfseries
\newcolumntype{U}{@{\,}>{\scriptsize}l}
\usepackage{enumitem}
\usepackage{algorithm}
\usepackage{algorithmic}
\usepackage{wrapfig}

\newcommand{\at}[2]{\ensuremath{\left[ #1 \right]}_{#2}}
\newtheorem{theorem}{Theorem}

\newcommand{\relabel}[1]{\tag{\ref{#1} revisited}}
\newcommand{\punc}[1]{\,#1}

\definecolor{viridisMin}{HTML}{440154}
\definecolor{solarizedBlack}{HTML}{000000}
\definecolor{solarizedGrey}{HTML}{B2B2B2}
\definecolor{solarizedHatchGrey}{HTML}{93A1A1}
\DeclareRobustCommand{\linepatch}[2][solid]{%
  \tikzset{dashstyle/.style={#1}}%
  \def\instruction{#1}%
  \def\dashedref{dashed}%
  \ifx\instruction\dashedref%
    \tikzset{dashstyle/.style={dash pattern=on 0.225cm off 0.05cm}}%
  \fi%
  \begin{tikzpicture}[baseline=-0.75ex, scale=0.1]
    \draw[very thick, color=solarized#2, dashstyle] (-2.5, 0.25) -- (2.5, 0.25);
  \end{tikzpicture}%
}
\DeclareRobustCommand\weightupdate{\tikz[baseline=0.75ex] \draw [weight update] (0, 0.25) -- +(0.5, 0) ;}
\DeclareRobustCommand\hpupdate{\tikz[baseline=0.75ex] \draw [hp update] (0, 0.25) -- +(0.5, 0) ;}
\DeclareRobustCommand\approxhpupdate{\tikz[baseline=0.75ex] \draw [approx hp update] (0, 0.25) -- +(0.5, 0) ;}
\DeclareRobustCommand\envelopekey{%
  \tikz[baseline=-0.75ex, scale=0.1]{
    \draw[draw=none, fill=black, opacity=0.4] (-2.5, -1.5) rectangle (2.5, 1.5);
    \draw[very thick, color=black] (-2.5, 0) -- (2.5, 0);}}

\title{Scalable One-Pass Optimisation of\\ High-Dimensional Weight-Update\\ Hyperparameters by Implicit Differentiation}

\author{%
  Ross M.\ Clarke\\
  University of Cambridge\\
  \texttt{rmc78@cam.ac.uk}\\
  \And
  Elre T.\ Oldewage\\
  University of Cambridge\\
  \texttt{etv21@cam.ac.uk}
  \And
  José Miguel Hernández-Lobato\\
  University of Cambridge\\
  Alan Turing Institute\\
  \texttt{jmh233@cam.ac.uk}
}

\iclrfinalcopy 

\begin{document}

\maketitle

\vspace*{-3.5ex}
\begin{abstract}
  Machine learning training methods depend plentifully and intricately on
  hyperparameters, motivating automated strategies for their optimisation.
  Many existing algorithms restart training for each new hyperparameter
  choice, at considerable computational cost. Some hypergradient-based one-pass
  methods exist, but these either cannot be applied to arbitrary optimiser hyperparameters
  (such as learning rates and momenta)
  or take several times longer to train than their base models. We extend these
  existing methods to develop an approximate hypergradient-based hyperparameter optimiser which is
  applicable to any continuous hyperparameter appearing in a differentiable model weight
  update, yet requires only one training episode, with no restarts. We also
  provide a motivating argument for convergence to the true hypergradient, and
  perform tractable gradient-based optimisation of independent learning rates
  for each model parameter. Our
  method performs competitively from varied random hyperparameter initialisations on
  several UCI datasets and Fashion-MNIST (using a one-layer MLP), Penn Treebank (using an LSTM) and CIFAR-10
  (using a ResNet-18), in time only 2--3x greater than vanilla training.
\end{abstract}
\vspace*{-1.5ex}

\section{Introduction}
Many machine learning methods are governed by \emph{hyperparameters}:
quantities other than model \emph{parameters} or \emph{weights} which
nonetheless influence training (e.g.\ optimiser settings,
dropout probabilities and dataset configurations). As suitable
hyperparameter selection is crucial to system performance (e.g.\ \citet{kohaviAutomaticParameterSelection1995}),
it is a pillar of efforts to automate machine learning \citep[Chapter 1]{hutterAutomatedMachineLearning2018}, spawning
several hyperparameter optimisation (HPO) algorithms (e.g.\
\citet{bergstraRandomSearchHyperParameter2012, snoekPracticalBayesianOptimization2012, snoekScalableBayesianOptimization2015, falknerBOHBRobustEfficient2018}).
However, HPO is computationally intensive and random search is an
unexpectedly strong (but beatable;
\citet{turnerBayesianOptimizationSuperior2021}) baseline; beyond random or grid
searches, HPO is relatively underused in research
\citep{bouthillierSurveyMachinelearningExperimental2020}.  

Recently, \citet{lorraineOptimizingMillionsHyperparameters2020} used gradient-based
updates to adjust hyperparameters \emph{during} training, displaying
impressive optimisation performance and scalability to high-dimensional
hyperparameters. Despite their computational efficiency (since updates occur before final
training performance is known),
\citeauthor{lorraineOptimizingMillionsHyperparameters2020}'s algorithm only
applies to hyperparameters on which the loss function depends explicitly (such
as $\ell_{2}$ regularisation), notably excluding optimiser hyperparameters.

Our work extends
\citeauthor{lorraineOptimizingMillionsHyperparameters2020}'s algorithm to support arbitrary
continuous inputs to a differentiable weight update formula, including
learning rates and momentum factors. We demonstrate our algorithm handles a range of
hyperparameter initialisations and datasets,
improving test loss after a single training episode (`one pass').
Relaxing differentiation-through-optimisation
\citep{domkeGenericMethodsOptimizationBased2012} and
hypergradient descent's \citep{baydinOnlineLearningRate2018} exactness allows us to improve
computational and memory efficiency. Our scalable one-pass
method improves performance from arbitrary hyperparameter
initialisations, and could be augmented with a further search over those
initialisations if desired.

  \tikzset{
    weight update/.style={solarizedMagenta, arrows={-Latex}},
    hp update/.style={solarizedGreen, arrows={-Latex}},
    approx hp update/.style={solarizedBlue, arrows={-Latex}},
    best response/.style={solarizedYellow, mark=none},
    naive training/.style={solarizedMagenta, dashed, mark=none},
    training min/.style={draw, fill=solarizedMagenta, color=solarizedMagenta, circle, inner sep=0.05cm},
    validation min/.style={draw, fill=solarizedMagenta, color=solarizedGreen, circle, inner sep=0.05cm},
  }
  \pgfplotsset{
    loss surface/.style={surf, colormap={SolarizedLight}{color=(solarizedBase2!80!black)
        color=(solarizedBase2!30!solarizedBase3) color=(solarizedBase3)}, shader=interp},
    best response/.style={solarizedYellow, mark=none},
    naive training/.style={solarizedMagenta, dashed, mark=none, samples y=0, domain=-3.5:0},
    training min/.style={mark=*, color=solarizedMagenta},
    validation min/.style={mark=*, color=solarizedGreen},
  }
\begin{figure*}
  \centerfigure
  \tikzmath{
    \lr = 0.1;
    \hyperlr = 0.1;
    \weightsteps = 2;
    \approxsteps = 5;
    \epochs = 10;
    \x{0} = -3.5;
    \y{0} = -3.0;
    \startx = \x{0};
    \starty = \y{0};
    function f(\x,\y) {return (\x + sin(\y r))^2 + 0.2*(\y + 2)^2;};
    function g(\x,\y) {return (\x - 1)^2 + (\y -1)^2;};
    function br(\y) {return -sin(\y r);};
    function dbr(\y) {return -cos(\y r) ;};
    function dx(\x,\y) {return 2*(\x + sin(\y r));};
    function dy(\x,\y) {2*(\x + sin(\y r))*cos(\y r) + 0.4*(\y + 2);};
    function ddx(\x, \y) {return 2;};
    function gdy(\x,\y) {return 2*(\y - 1);};
    function ddirect(\x,\y) {return 2 * (\y - 1) ;};
    function dindirect(\x,\y) {return 2 * (\x - 1) * dbr(\y);};
    function hypergrad(\x,\y) {ddirect(\x, \y) + dindirect(\x, \y);};
    int \step, \laststep, \epoch, \lastepoch;
    for \epoch in {1,2,...,\epochs}{
      \lastepoch = \epoch - 1 ;
      for \n in {1,2,...,\weightsteps}{
        \step = \weightsteps * (\epoch - 1) + \n ;
        \laststep = \step - 1 ;
        \x\step = \x\laststep - \lr*dx(\x\laststep, \y\lastepoch) ;
      };
      \y\epoch = \y\lastepoch - \hyperlr * hypergrad(\x\step, \y\lastepoch) ;
      \approxsum = 0.0;
      for \j in {0,1,...,\approxsteps-1}{
        \approxsum = \approxsum + (1 - \lr*ddx(\x\step, \y\lastepoch))^\j;
      };
      \approxy\epoch = \y\lastepoch - \hyperlr*(ddirect(\x\step,
      \y\lastepoch) - \lr*gdy(\x\step, \y\lastepoch) * \approxsum);
    };
  }
  \begin{subfigure}[m]{0.35\textwidth}
    \centering
    \begin{tikzpicture}
      \begin{axis}[
        view/v=45,
        ticks=none,
        label style={sloped},
        xlabel={Weight $\vec{w}$},
        ylabel={Hyperparameter $\vec{\lambda}$},
        zlabel={Training Loss $\mathcal{L}_{T}(\vec{\lambda}, \vec{w})$},
        ymax=1,
        zmin=0,
        scale=0.56
        ]
        \addplot3 [loss surface] {f(x, y)};
        \addplot3 [best response] ({br(y)}, y, {f(br(y), y)});
        \addplot3 [naive training] (x, \starty, {f(x, \starty)}) ;
        \addplot3 [training min] (1, -2, 0);
        \addplot3 [validation min] (1, 1, 0);
        \tikzmath{
          int \lasts, \curs, \epoch, \lastepoch;
          for \epoch in {1,2,...,\epochs}{
            \lastepoch = \epoch - 1 ;
            \arrowscale = 1.1 - 0.1 * \epoch;
            for \n in {1,2,...,\weightsteps}{
              \curs = \weightsteps * (\epoch - 1) + \n ;
              \lasts = \curs - 1 ;
              {\edef\d{
                  \noexpand\draw[weight update, arrows={-Latex[scale=\arrowscale]}]
                  (\x\lasts, \y\lastepoch,
                  {f(\x\lasts, \y\lastepoch)})
                  -- (\x\curs, \y\lastepoch, {f(\x\curs, \y\lastepoch)}) ;}
                \d};
            };
            {\edef\d{\noexpand\draw[hp update, arrows={-Latex[scale=\arrowscale]}] (\x\curs, \y\lastepoch,
                {f(\x\curs, \y\lastepoch)}) -- (\x\curs, \y\epoch, {f(\x\curs,
                  \y\epoch)});
              \noexpand\draw[approx hp update, arrows={-Latex[scale=\arrowscale]}] (\x\curs, \y\lastepoch, {f(\x\curs,
                \y\lastepoch)}) -- (\x\curs, \approxy\epoch, {f(\x\curs,
                \approxy\epoch)});}
            \d};
          };
        }
      \end{axis}
    \end{tikzpicture}
    \caption{Training Space}
  \end{subfigure}
  \hfill
  \begin{subfigure}[m]{0.35\textwidth}
    \centering
    \begin{tikzpicture}
      \begin{axis}[
        view/v=45,
        ticks=none,
        label style={sloped},
        xlabel={Weight $\vec{w}$},
        ylabel={Hyperparameter $\vec{\lambda}$},
        zlabel={Validation Loss $\mathcal{L}_{V}(\vec{\lambda}, \vec{w})$},
        ymax=1,
        scale=0.56
        ]
        \addplot3 [loss surface] {g(x, y)};
        \addplot3 [best response] ({br(y)}, y, {g(br(y), y)});
        \addplot3 [naive training] (x, \starty, {g(x, \starty)}) ;
        \addplot3 [training min] (1, -2, {g(1, -2)});
        \addplot3 [validation min] (1, 1, 0);
        \tikzmath{
          int \lasts, \curs, \epoch, \lastepoch;
          for \epoch in {1,2,...,\epochs}{
            \lastepoch = \epoch - 1;
            \arrowscale = 1.1 - 0.1 * \epoch;
            for \n in {1,2,...,\weightsteps}{
              \curs = \weightsteps * (\epoch - 1) + \n;
              \lasts = \curs - 1;
              { \edef\d{\noexpand\draw[weight update, arrows={-Latex[scale=\arrowscale]}]
                  (\x\lasts, \y\lastepoch, {g(\x\lasts, \y\lastepoch)})
                  -- (\x\curs, \y\lastepoch, {g(\x\curs, \y\lastepoch)});} \d };
            };
            { \edef\d{\noexpand\draw[hp update, arrows={-Latex[scale=\arrowscale]}]
                (\x\curs, \y\lastepoch, {g(\x\curs, \y\lastepoch)})
                -- (\x\curs, \y\epoch, {g(\x\curs, \y\epoch)});
              \noexpand\draw[approx hp update, arrows={-Latex[scale=\arrowscale]}] (\x\curs, \y\lastepoch, {g(\x\curs,
                \y\lastepoch)}) -- (\x\curs, \approxy\epoch, {g(\x\curs, \approxy\epoch)});} \d };
          };
        }
      \end{axis}
    \end{tikzpicture}
    \caption{Validation Space}
  \end{subfigure}
  \hfill
  \DeclareRobustCommand\bestresponse{\tikz[baseline=0.75ex] \draw [best response, thick] (0, 0.25) -- +(0.5, 0) ;}
  \DeclareRobustCommand\naivetraining{\tikz[baseline=0.75ex] \draw [naive training] (0, 0.25) -- +(0.5, 0) ;}
  \DeclareRobustCommand\trainingmin{\tikz \node [training min] (0, 0) {};}
  \DeclareRobustCommand\validationmin{\tikz \node [validation min] (0, 0) {};}
  \begin{subfigure}{0.30\textwidth}
    \centering
    \begin{tabular}{cl}
      \weightupdate & $\Delta\vec{w} \propto \pderiv{\mathcal{L}_{T}}{\vec{w}}$ \\
      \hpupdate & $\Delta\vec{\lambda} \propto \deriv{\mathcal{L}_{V}}{\vec{\lambda}}$ \\
      \approxhpupdate &
      $\widehat{\Delta\vec{\lambda}} \approxpropto \deriv{\mathcal{L}_{V}}{\vec{\lambda}}$ \\
      \bestresponse & $\vec{w}^{*}(\vec{\lambda})$  $^\dagger$\\
      \naivetraining & $\mathcal{L}_{T}(\vec{\lambda} = \vec{\lambda}_{0}, \vec{w})$\\
      \trainingmin &
      $\min_{\vec{\lambda}, \vec{w}} \mathcal{L}_T(\vec{\lambda}, \vec{w})$ $^\dagger$\\
      \validationmin &
      $\min_{\vec{\lambda}, \vec{w}} \mathcal{L}_V(\vec{\lambda}, \vec{w})$ $^\dagger$\\[1ex]
    \end{tabular}
    \raggedright $^\dagger$ projected onto corresponding loss surface
  \end{subfigure}

    \caption{Summary of our derivation (Section~\ref{sec:Derivations}): an example of online
    hypergradient descent using exact hypergradients from the
    implicit function theorem (IFT) (\hpupdate), with our method's approximate
    hypergradients (\approxhpupdate) superimposed. We target
      optimal validation loss (\validationmin),
    adjusting weights $\vec{w}$ based on the training loss. Classical
    weight updates (for fixed hyperparameters $\vec{\lambda}$) converge (\weightupdate,
    \naivetraining) to the
    best-response line $\vec{w}^{*}(\vec{\lambda})$ (\bestresponse);
    the IFT gives hyperparameter updates (\hpupdate)
    leading to a minimum of validation loss along $\vec{w}^{*}(\vec{\lambda})$.
    Our approximate hyperparameter updates (\approxhpupdate)
    differ in magnitude from these exact updates, but still give useful guidance.}
\label{fig:SummaryFigure}
\end{figure*}
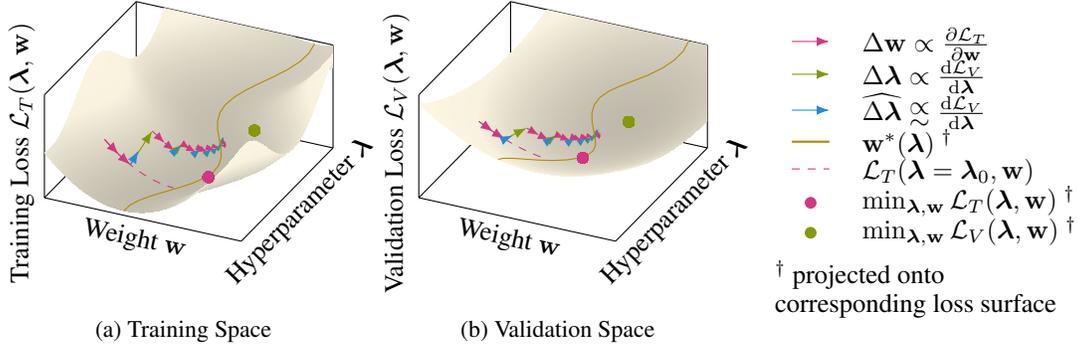

\section{Weight-Update Hyperparameter Tuning}
\label{sec:Derivations}
In this section, we develop our method. Expanded derivations and a summary of differences from \citet{lorraineOptimizingMillionsHyperparameters2020} are given in Appendix~\ref{sec:VerboseDerivations}.

\subsection{Implicit Function Theorem in Bilevel Optimisation}
Consider some model with learnable parameters $\vec{w}$, training loss
$\mathcal{L}_{T}$, optimisation hyperparameters $\vec{\lambda}$ and validation
loss $\mathcal{L}_{V}$. We use $\vec{w}^{*}, \vec{\lambda}^{*}$ to represent the
optimal values of these quantities, found by solving the following \emph{bilevel optimisation problem}:
\begin{align}\label{eq:BilevelOptimisation}
  ^\text{(a)}\  \vec{\lambda}^{*} = \argmin_{\vec{\lambda}} \mathcal{L}_{V}(\vec{\lambda}, \vec{w}^{*}(\vec{\lambda})) \punc{,}
  &&
    \text{such that}
  &&
  ^\text{(b)}\  \vec{w}^{*}(\vec{\lambda}) = \argmin_{\vec{w}} \mathcal{L}_{T} (\vec{\lambda}, \vec{w}) \punc{.}
\end{align}
The optimal model parameters $\vec{w}^{*}$ may vary with
$\vec{\lambda}$, making $\mathcal{L}_{V}$ an \emph{implicit} function of
$\vec{\lambda}$ alone.

We approach the outer optimisation (\ref{eq:BilevelOptimisation}a) similarly to
\citet{lorraineOptimizingMillionsHyperparameters2020} and \citet{majumderLearningLearningRate2019}, using the
\emph{hypergradient}: the total derivative of $\mathcal{L}_V$ with respect to the hyperparameters $\vec{\lambda}$.
One strategy for solving \eqref{eq:BilevelOptimisation} is therefore to
alternate between updating $\vec{w}$ for several steps using
$\pderiv{\mathcal{L}_T}{\vec{w}}$ and  updating $\vec{\lambda}$ using the
hypergradient, as shown in
Figure~\ref{fig:SummaryFigure} (by \weightupdate{} and \hpupdate).

Carefully distinguishing the total differential
$\mathrm{d}\vec{\lambda}$ and the partial differential $\partial \vec{\lambda}$, we
have
\begin{equation}
  \label{eq:TotalDerivative}
  \deriv{\mathcal{L}_{V}}{\vec{\lambda}} = \pderiv{\mathcal{L}_{V}}{\vec{\lambda}} + \pderiv{\mathcal{L}_{V}}{\vec{w}^{*}} \pderiv{\vec{w}^{*}}{\vec{\lambda}} \punc{.}
\end{equation}

While derivatives of $\mathcal{L}_{V}$ are easily computed for typical loss functions, the final derivative of the optimal model
parameters ($\pderiv{\vec{w}^{*}}{\vec{\lambda}}$) presents some difficulty. Letting square brackets indicate the evaluation of the interior at the
  subscripted values,
we may rewrite $\pderiv{\vec{w}^{*}}{\vec{\lambda}}$ as follows:
\begin{theorem}[Cauchy's Implicit Function Theorem (IFT)]
  \label{thm:ImplicitFunctionTheorem}
  Suppose for some $\vec{\lambda}'$ and $\vec{w}'$ that
  $\at{\pderiv{\mathcal{L}_{T}}{\vec{w}}}{\vec{\lambda}', \vec{w}'} = \vec{0}$.
  If $\pderiv{\mathcal{L}_{T}}{\vec{w}}$ is a continuously differentiable
  function with invertible Jacobian, then there exists a function
  $\vec{w}^{*}(\vec{\lambda})$ over an open subset of hyperparameter space, such
  that $\vec{\lambda}'$ lies in the open subset,
  $\at{\pderiv{\mathcal{L}_{T}}{\vec{w}}}{\vec{\lambda}, \vec{w}^{*}(\vec{\lambda})} = \vec{0}$
  and
  \begin{equation}\label{eq:ImplicitFunctionTheorem}
    \pderiv{\vec{w}^{*}}{\vec{\lambda}} = - \left( \frac{\partial^{2} \mathcal{L}_{T}}{\partial \vec{w} \partial \vec{w}\trans} \right)^{-1} \frac{\partial^{2} \mathcal{L}_{T}}{\partial \vec{w} \partial \vec{\lambda}\trans} \punc{.}
  \end{equation}
\end{theorem}
$\vec{w}^{*}(\vec{\lambda})$ is called the \emph{best response} of $\vec{w}$ to
$\vec{\lambda}$ (Figure~\ref{fig:SummaryFigure}). While
\eqref{eq:ImplicitFunctionTheorem} suggests a route to computing
$\pderiv{\vec{w}^{*}}{\vec{\lambda}}$, inverting a potentially
high-dimensional Hessian in $\vec{w}$ is not computationally tractable.

\subsection{Approximate Best-Response Derivative}
To develop and justify a computationally tractable approximation to
 \eqref{eq:ImplicitFunctionTheorem}, we mirror the strategy of \citet{lorraineOptimizingMillionsHyperparameters2020}.
Consider the broad class of weight optimisers with updates of the form
\begin{equation}\label{eq:GeneralWeightUpdate}
  \vec{w}_{i}(\vec{\lambda}) = \vec{w}_{i-1}(\vec{\lambda}) - \vec{u}(\vec{\lambda}, \vec{w}_{i-1}(\vec{\lambda}))
\end{equation}
for some arbitrary differentiable function $\vec{u}$, with $i$ indexing each
update iteration. We deviate here from the approach of
\citet{lorraineOptimizingMillionsHyperparameters2020} by admitting general
functions $\vec{u}(\vec{\lambda}, \vec{w})$, rather than assuming the
particular choice
$\vec{u}_{\textrm{SGD}} = \eta \pderiv{\mathcal{L}_{T}}{\vec{w}}$ (see
Appendix~\ref{sec:ExpandedUpdateFunction} for more details). In particular,
this allows $\vec{\lambda}$ to include optimiser hyperparameters. Differentiating
\eqref{eq:GeneralWeightUpdate} and unrolling the recursion gives
\begin{equation}\label{eq:UnrolledUpdate}
    \at{\pderiv{\vec{w}_{i}}{\vec{\lambda}}}{\vec{\lambda}'}
    =
    - \sum_{0 \leq j < i} \Bigg( \prod_{0 \leq k < j} \at{\vec{I} - \pderiv{\vec{u}}{\vec{w}}}{\vec{\lambda}', \vec{w}_{i-1-k}} \Bigg) \at{\pderiv{\vec{u}}{\vec{\lambda}}}{\vec{\lambda}', \vec{w}_{i-1-j}} \punc{,}
\end{equation}
where $j$ indexes our steps back through time from $\vec{w}_{i-1}$ to
$\vec{w}_{0}$, and all $\vec{w}$ depend on the current hyperparameters $\vec{\lambda}'$ --- see Appendix~\ref{sec:AppendixApproximateBestResponseDerivative} for the full derivation.
Now, we follow \citeauthor{lorraineOptimizingMillionsHyperparameters2020} and assume
 $\vec{w}_{0},\ldots,\vec{w}_{i-1}$ to be
equal to $\vec{w}_{i}$.
With this, we simplify the product in \eqref{eq:UnrolledUpdate} to
a $j$th power by evaluating all derivatives at $\vec{w}_i$. This result
is then used to approximate $\pderiv{\vec{w}^{*}}{\vec{\lambda}}$ by further assuming that
$\vec{w}_{i} \approx \vec{w}^{*}$. These two approximations lead to
\begin{equation}\label{eq:BestResponseApproximation}
  \at{\pderiv{\vec{w}_{i}}{\vec{\lambda}}}{\vec{\lambda}'}
  \approx - \at{ \sum_{0 \leq j < i} \left( \vec{I} - \pderiv{\vec{u}}{\vec{w}} \right)^{j} \pderiv{\vec{u}}{\vec{\lambda}}}{\vec{\lambda}', \vec{w}_{i}(\vec{\lambda}')}
  \approx \at{\pderiv{\vec{w}^{*}}{\vec{\lambda}}}{\vec{\lambda}'} 
  \punc{.}
\end{equation}
We reinterpret $i$ as a predefined look-back distance, trading off
accuracy and computational efficiency.

The combination of these approximations implies $\vec{w}_{i} = \vec{w}^{*}$ for all $i$, which is
initially inaccurate, but which we would expect to become more correct as 
training proceeds. In
mitigation, we perform several weight updates prior to each
hyperparameter update.
This means derivatives in earlier terms of the series of
\eqref{eq:BestResponseApproximation} (which are likely the largest, dominant terms) are evaluated at weights closer to $\vec{w}^*$, therefore making the summation more accurate. 
In Section~\ref{sec:Experiments}, we show that the approximations described here result in an algorithm that is both practical and effective.

Our approximate result \eqref{eq:BestResponseApproximation} combines the general weight
update of \citet{majumderLearningLearningRate2019} with the
overall approach and constant-weight assumption of
\citet{lorraineOptimizingMillionsHyperparameters2020}. The latter empirically show that an approximation similar to \eqref{eq:BestResponseApproximation} leads to a directionally-accurate
approximate hypergradient; we illustrate the approximate updates from our
derivations in Figure~\ref{fig:SummaryFigure} (by \approxhpupdate).

\subsection{Convergence to Best-Response Derivative}
To justify the approximations in \eqref{eq:BestResponseApproximation}, note that
the central part of that equation is a truncated Neumann series. 
Taking the limit $i \to \infty$ , when such a limit exists, results in the closed form
\begin{equation}\label{eq:InfiniteLimit}
  \at{\pderiv{\vec{w}^{*}}{\vec{\lambda}}}{\vec{\lambda}'}
  \approx - \at{\left( \pderiv{\vec{u}}{\vec{w}} \right)^{-1} \pderiv{\vec{u}}{\vec{\lambda}}}{\vec{\lambda}', \vec{w}^{*}(\vec{\lambda}')} \punc{.}
\end{equation}
This is precisely the result of the IFT (Theorem~\ref{thm:ImplicitFunctionTheorem}) applied to
$\vec{u}$ instead of $\pderiv{\mathcal{L}_T}{\vec{w}}$; that is, substituting the simple SGD update
$\vec{u}_{\textrm{SGD}}(\vec{\lambda}, \vec{w}) = \eta \pderiv{\mathcal{L}_{T}}{\vec{w}}$ into \eqref{eq:InfiniteLimit} recovers
\eqref{eq:ImplicitFunctionTheorem} exactly. Thus, under certain conditions, our approximation
\eqref{eq:BestResponseApproximation} converges to the true best-response
Jacobian in the limit of infinitely long look-back windows. 

\subsection{Hyperparameter Updates}
Substituting \eqref{eq:BestResponseApproximation} into
\eqref{eq:TotalDerivative} yields a tractable approximation for the hypergradient
$\deriv{\mathcal{L}_{V}}{\vec{\lambda}}$, with which we can update hyperparameters
by gradient descent.
Our implementation
in Algorithm~\ref{alg:OurAlgorithm} (Figure \ref{fig:MainAnalyticalToyAblation})
closely parallels \citeauthor{lorraineOptimizingMillionsHyperparameters2020}'s  algorithm, invoking
Jacobian-vector products \citep{pearlmutterFastExactMultiplication1994} during gradient computation for memory efficiency via
the \verb|grad_outputs| argument, which also provides the repeated
multiplication for the $j$th power in \eqref{eq:BestResponseApproximation}.
Thus, we retain the $\mathcal{O}(|\vec{w}| + |\vec{\lambda}|)$ time and
memory cost of
\citet{lorraineOptimizingMillionsHyperparameters2020}, where $|\cdot|$ denotes cardinality.
The core loop to compute the summation in \eqref{eq:BestResponseApproximation} comes from
an algorithm of \citet{liaoRevivingImprovingRecurrent2018}.
Note that Algorithm~\ref{alg:OurAlgorithm}
approximates $\deriv{\mathcal{L}_{V}}{\vec{\lambda}}$ retrospectively by considering
only the last weight update rather than any future weight updates.

Unlike differentiation-through-optimisation
\citep{domkeGenericMethodsOptimizationBased2012}, Algorithm~\ref{alg:OurAlgorithm} crucially estimates
hypergradients without reference to old model parameters, thanks to the approximate-hypergradient
construction of \citet{lorraineOptimizingMillionsHyperparameters2020} and
\eqref{eq:BestResponseApproximation}. We thus do not store network weights
at multiple time steps, so gradient-based HPO becomes possible on
previously-intractable large-scale problems. In essence, we develop an approximation to online
hypergradient descent \citep{baydinOnlineLearningRate2018}.

Optimiser hyperparameters generally do not affect the optimal weights, suggesting their
hypergradients should be zero. However, in practice,
$\vec{w}^{*}$ is better reinterpreted as the \emph{approximately} optimal
weights obtained after a finite training episode. These certainly depend on the optimiser
hyperparameters, which govern the convergence of $\vec{w}$, thus justifying our use of the
bilevel framework.

We emphasise training is not reset after each hyperparameter update --- we
simply continue training from where we left off, using the new hyperparameters.
Consequently, Algorithm~\ref{alg:OurAlgorithm}
avoids the time cost of multiple training restarts.
While our locally greedy hyperparameter updates
threaten a short-horizon bias \citep{wuUnderstandingShortHorizonBias2018a},
we still realise practical improvements in our experiments.

\subsection{Reinterpretation of Iterative Optimisation}
\label{sec:ReinterpretationOfIterativeOptimisation}
Originally, we stated the IFT
\eqref{eq:ImplicitFunctionTheorem} in terms of minima of $\mathcal{L}_{T}$
(zeros of $\pderiv{\mathcal{L}_{T}}{\vec{w}}$), and
substituting $\vec{u}_{\textrm{SGD}} = \eta \pderiv{\mathcal{L}_{T}}{\vec{w}}$ into \eqref{eq:InfiniteLimit} recovers
this form of \eqref{eq:ImplicitFunctionTheorem}. However, in general, \eqref{eq:InfiniteLimit} recovers the Theorem
for zeros of $\vec{u}$, which are not necessarily minima of the training loss.
Despite this, our development can be compared to
\citet{lorraineOptimizingMillionsHyperparameters2020} by expressing $\vec{u}$
as the derivative of an augmented `pseudo-loss' function.
Consider again the simple SGD update $\vec{u}_{\textrm{SGD}}$, which provides
the weight update rule $\vec{w}_{i} = \vec{w}_{i-1} - \eta \pderiv{\mathcal{L}_{T}}{\vec{w}}$.
By trivially defining a pseudo-loss
$\overline{\mathcal{L}} = \eta \mathcal{L}_{T}$, we may absorb
$\eta$ into a loss-like derivative, yielding
  $\vec{w}_{i} = \vec{w}_{i-1} - \pderiv{\overline{\mathcal{L}}}{\vec{w}}$. More
  generally, we may write
  $\overline{\mathcal{L}} = \int \vec{u}(\vec{\lambda}, \vec{w}) \,\mathrm{d}\vec{w}$.

\begin{figure}
\begin{minipage}[b]{0.48\textwidth}
    \def\start{-2}
    \centering
    \begin{subfigure}[b]{\linewidth}
      \centering
      \begin{tikzpicture}[
        declare function={f(\x) = (\x - 2)^2;},
        declare function={dx(\x) = 2*(\x - 2);},
        declare function={lr = 0.5;},
        ]
        \begin{axis}[
          axis x line=none,
          axis y line=none,
          xmin=-3,
          ymin=-2,
          ymax=40,
          scale=0.65,
          enlargelimits=false,
          scale only axis,
          height=4cm,
          width=\linewidth
          ]
          \addplot [black, thick] {f(x)};
          \coordinate (start) at ({\start}, {f(\start)});
          \foreach \lr / \lri / \pos  in {0.1 / 1 / below left=3pt and 1pt of step, 0.2 / 2 / above, 0.4 / 3 / right} {
            \pgfmathsetmacro\new{\start - \lr*dx(\start)}
            \edef\body{
              \noexpand\draw[approx hp update] (start) -- (\new, {f(\start)}) coordinate
              (step) node [\pos] {$\eta_{\lri} \pderiv{\noexpand\mathcal{L}}{\noexpand\vec{w}}$};
              \noexpand\draw[approx hp update, dotted, -] (step) -- (\new, {f(\new)})  coordinate (end) ;
              \noexpand\draw[weight update] (start) -- (end) ;
            }
            \body
          }
          \draw[black] (4, 8) node {$\mathcal{L}$};
        \end{axis}
      \end{tikzpicture}
      \caption{$\vec{w} \gets \vec{w} - \eta_{i} \pderiv{\mathcal{L}}{\vec{w}}$}\label{fig:ClassicInterpretation}
    \end{subfigure}

    \begin{subfigure}[b]{\linewidth}
      \centering
      \begin{tikzpicture}[
        declare function={f(\x) = (\x - 2)^2;},
        declare function={dx(\x) = 2*(\x - 2);},
        ]
        \begin{axis}[
          axis x line=none,
          axis y line=none,
          xmin=-3,
          xmax=6,
          ymin=-0.5,
          scale=0.65,
          enlargelimits=false,
          scale only axis,
          height=4cm,
          width=\linewidth
          ]
          \foreach \lr / \lri / \pos in {0.1 / 1 / below left=-3pt and 5pt of step, 0.2 / 2 / right=2pt, 0.4 / 3 / right} {
            \pgfmathsetmacro\new{\start - \lr*dx(\start)}
            \edef\body{
              \noexpand\coordinate (start) at ({\start}, {\lr*f(\start)});
              \noexpand\addplot [black, thick] {\lr * f(x)};
              \noexpand\draw[approx hp update] (start) -- (\new, {\lr*f(\start)}) coordinate (step)
              node [\pos] {$\pderiv{\overline{\noexpand\mathcal{L}}_{\lri}}{\noexpand\vec{w}}$};
              \noexpand\draw[weight update] (start) -- (\new, {\lr*f(\new)}) coordinate (new);
              \noexpand\draw[approx hp update, dotted, -] (new) -- (step) ;
            }
            \body
          }
          \draw[approx hp update, dotted, -] (\start, {0.1*f(\start)}) -- (\start, {0.4*f(\start)});
          \draw[black] (5.5, 4.4) node {$\overline{\mathcal{L}}_{3}$};
          \draw[black] (5.5, 2.3) node {$\overline{\mathcal{L}}_{2}$};
          \draw[black] (5.5, 0.5) node {$\overline{\mathcal{L}}_{1}$};
        \end{axis}
      \end{tikzpicture}
      \caption{$\vec{w} \gets \vec{w} - \pderiv{\overline{\mathcal{L}}_{i}}{\vec{w}}, \quad \overline{\mathcal{L}}_{i} = \eta_{i} \mathcal{L}$}\label{fig:OurInterpretation}
    \end{subfigure}

    \DeclareRobustCommand\weightupdatekey{
      \tikz[baseline=-0.75ex, scale=0.1]{
        \draw[approx hp update] (0, 0) -- (10, 0);}}
    \DeclareRobustCommand\lossupdatekey{
      \tikz[baseline=-0.75ex, scale=0.1]{
        \draw[weight update] (0, 0) -- (10, 0);}}
  \vspace*{2ex}
  \begin{tabular}{cl}
    \weightupdatekey & Weight updates\\
    (\lossupdatekey) & (projected onto $\mathcal{L}$)
  \end{tabular}
  \caption{Reinterpreting the role of learning rate $\eta$ over a loss function $\mathcal{L}$. (a)~In the classical
    setting, $\eta$ scales the gradient of some fixed $\mathcal{L}$. (b)~In our
    setting, $\eta$ scales $\mathcal{L}$ to form a `pseudo-loss'
    $\overline{\mathcal{L}}$, whose gradient is used as-is: our loss function
    has become dependent on $\eta$. The same
    weight updates are obtained in both (a) and (b).}\label{fig:ReinterpretedHyperparameters}
\end{minipage}\hfill%
\begin{minipage}[b]{0.48\textwidth}
  \centering
  \includegraphics[width=\linewidth]{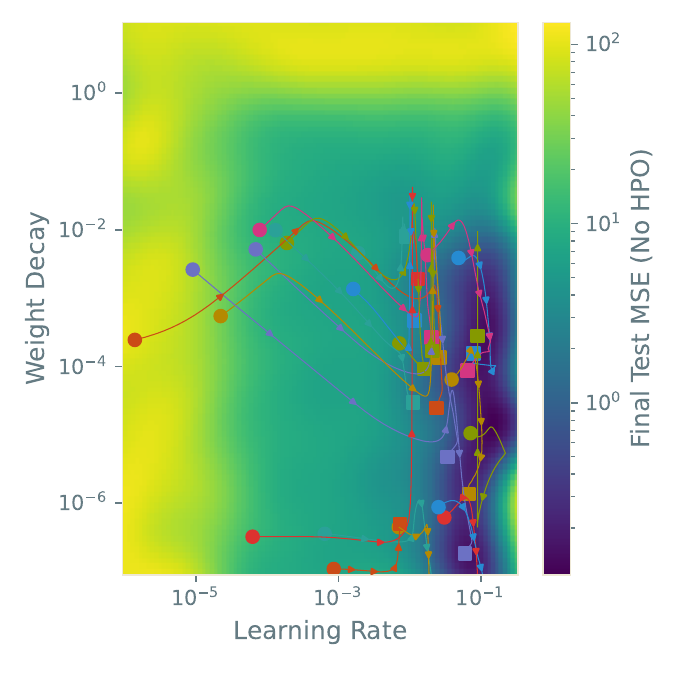}
  \DeclareRobustCommand\evolutionkey{
    \tikz[baseline=-0.75ex, scale=0.1]{
      \foreach \c / \y in {solarizedRed/2.5, solarizedGreen/0, solarizedBlue/-2.5} {
        \draw[fill=\c, color=\c] (0, \y) circle (1);
        \draw[thick, color=\c, arrows={-Triangle[round]}, shorten >=-3.5pt] (0, \y) -- +(5, 0);
        \draw[thick, color=\c] (5, \y) -- +(5, 0);
        \draw[fill=\c, color=\c, rounded corners=0.2pt] (9, \y-1) rectangle +(2, 2);
      };}}
  \DeclareRobustCommand\nankey{
    \tikz[baseline=-0.75ex, scale=0.1]{
      \draw[black, very thick] (-1, -1) -- (1, 1) (-1, 1) -- (1, -1);
      \draw[very thick, black, arrows={-Triangle[round]}] (0, 0) -- (10, 0);}}
  \begin{tabular}{cl}
    \evolutionkey & \makecell[l]{Initialisation / trajectory / end\\of multiple runs} \\
  \end{tabular}

  \caption{Sample hyperparameter trajectories from training UCI Energy under
    \emph{Our$^\text{WD+LR}$} configuration.
    Background shading gives a non-HPO baseline with hyperparameters fixed at the
    corresponding initial point; these results are interpolated by a Gaussian process.
    Note the trajectories are generally attracted to the valley of high
    performance at learning rates around $10^{-1}$ and weight decays below $10^{-2}$.}
  \label{fig:ToyTrajectories}
\end{minipage}
\end{figure}

Expressing the update in this form suggests a reinterpretation of the role of
optimiser hyperparameters. Conventionally, our visualisation of gradient descent
has the learning rate control the size of steps over some undulating
landscape. Instead, we propose fixing a unit step size, with the `learning
rate' scaling the landscape underneath. Similarly, we suppose a `momentum' could,
at every point, locally squash the loss surface in the negative-gradient direction
and stretch it in the positive-gradient direction. In aggregate, these transformations
straighten out optimisation trajectories and bring local optima closer to the
current point. While more complex hyperparameters lack a clear visualisation in
this framework, it nevertheless allows
a broader class of hyperparameters to `directly alter the loss function' instead
of remaining completely independent,
circumventing the problem with optimiser hyperparameters noted by
\citet{lorraineOptimizingMillionsHyperparameters2020}.
Figure~\ref{fig:ReinterpretedHyperparameters} and
Appendix~\ref{sec:ExpandedIntuitiveReinterpretation} support this
argument.

\section{Related Work}
\citet{kohaviAutomaticParameterSelection1995} first noted different problems
respond optimally to different hyperparameters; \citet[Chapter 1]{hutterAutomatedMachineLearning2018} summarise the resulting 
\emph{hyperparameter optimisation} (HPO) field.

Black-box HPO treats the training process as atomic, selecting trial
configurations by grid search (coarse or intractable at scale), random search (\citet{bergstraRandomSearchHyperParameter2012};
often more efficient) or population-based searches
(mutating promising trials). Pure Bayesian Optimisation
\citep{mockusApplicationBayesianMethods1978, snoekPracticalBayesianOptimization2012}
guides the search with a predictive model; many works seek to
exploit its sample efficiency (e.g.\
\citet{swerskyRaidersLostArchitecture2014,
  snoekScalableBayesianOptimization2015, wangBayesianOptimizationBillion2016,
  kandasamyMultifidelityBayesianOptimisation2017,
  levesqueBayesianOptimizationConditional2017,
  perroneLearningSearchSpaces2019}). However, these methods require each proposed
configuration to be fully trained, incurring considerable computational expense.
Other techniques infer information \emph{during} a training run --- from learning
curves \citep{provostEfficientProgressiveSampling1999,
  swerskyFreezeThawBayesianOptimization2014, domhanSpeedingAutomaticHyperparameter2015,
  chandrashekaranSpeedingHyperparameterOptimization2017,
  kleinLearningCurvePrediction2017}, smaller surrogate problems
\citep{petrakFastSubsamplingPerformance2000, vandenboscha.WrappedProgressiveSampling2004,
  kruegerFastCrossValidationSequential2015, sparksAutomatingModelSearch2015,
  thorntonAutoWEKACombinedSelection2013,
  sabharwalSelectingNearoptimalLearners2016} or intelligent resource allocation
\citep{jamiesonNonstochasticBestArm2016, liHyperbandNovelBanditbased2017, falknerBOHBRobustEfficient2018,
  bertrandHyperparameterOptimizationDeep2017,
  wangCombinationHyperbandBayesian2018}. Such techniques could be applied on top of our algorithm to improve performance.

In HPO with nested \emph{bilevel} optimisation, hyperparameters
are optimised conditioned on optimal weights, enabling
updates \emph{during} training, with a separate validation set mitigating overfitting risk; this relates to meta-learning \citep{franceschiBilevelProgrammingHyperparameter2018}.
Innovations include differentiable unrolled stochastic gradient descent (SGD) updates
\citep{domkeGenericMethodsOptimizationBased2012,
  maclaurinGradientbasedHyperparameterOptimization2015, baydinOnlineLearningRate2018,
  shabanTruncatedBackpropagationBilevel2019, majumderLearningLearningRate2019}, conjugate gradients or
hypernetworks \citep{lorraineStochasticHyperparameterOptimization2018,
  lorraineOptimizingMillionsHyperparameters2020,
  mackaySelfTuningNetworksBilevel2019, fuNeuralOptimizersHypergradients2017}, solving one level while
penalising suboptimality of the other
\citep{mehraPenaltyMethodInversionFree2020}, and deploying Cauchy's implicit function theorem
\citep{larsenDesignRegularizationNeural1996,
  bengioGradientBasedOptimizationHyperparameters2000,
  luketinaScalableGradientBasedTuning2016,
  lorraineOptimizingMillionsHyperparameters2020}.
\citet{doniniMARTHESchedulingLearning2020} extrapolate training to optimise
arbitrary learning rate schedules, extending earlier work on gradient
computation \citep{franceschiForwardReverseGradientBased2017}, but do not
immediately accommodate other hyperparameters.
  While non-smooth procedures exist
  \citep{lopez-ramosOnlineHyperparameterSearch2021}, most methods focus on
  smooth hyperparameters which augment the loss function (e.g.\ weight
  regularisation) and work with the augmented loss directly. Consequently, they cannot handle optimiser
hyperparameters not represented in the loss function
(e.g.\ learning rates and momentum factors; \citet{lorraineOptimizingMillionsHyperparameters2020}).
Many of these methods compute hyperparameter updates locally (not over the entire
training process), which may induce short-horizon bias
\citep{wuUnderstandingShortHorizonBias2018a}, causing myopic convergence to
local optima.

Modern developments include theoretical reformulations of bilevel
optimisation to improve performance \citep{liuGenericFirstOrderAlgorithmic2020,
  liImprovedBilevelModel2020}, optimising distinct hyperparameters for each
model parameter \citep{lorraineOptimizingMillionsHyperparameters2020,
  jieAdaptiveMultilevelHypergradient2020}, and computing forward-mode
hypergradient averages using more exact techniques than we do
\citep{micaelliNongreedyGradientbasedHyperparameter2020}. Although these
approaches increase computational efficiency and the range of tunable
parameters, achieving both benefits at once remains challenging. Our
algorithm accommodates a diverse range of
differentiable hyperparameters, but retains the efficiency of existing
approaches (specifically \citet{lorraineOptimizingMillionsHyperparameters2020}).
Inevitably, we also inherit gradient-based methods' difficulties with robustness and
discrete hyperparameters.


\section{Experiments}
\label{sec:Experiments}

\begin{figure}
  \begin{minipage}{0.60\linewidth}
    \vspace*{-3ex}
    \begin{algorithm}[H]
      \caption{\\Scalable One-Pass Optimisation of High-Dimensional Weight-Update Hyperparameters by
        Implicit Differentiation}
      \label{alg:OurAlgorithm}
      \def\func#1{\texttt{\small #1}}
      \begin{algorithmic}
        \WHILE{training continues} \FOR[$T$ steps of weight updates]{$t \gets 1$ \TO
          $T$} \STATE $\vec{w} \gets \vec{w} - \vec{u}(\vec{\lambda}, \vec{w})$ \ENDFOR
        \STATE
        $\vec{p} = \vec{v} = \at{\pderiv{\mathcal{L}_{V}}{\vec{w}}}{\vec{\lambda}, \vec{w}}$
        \COMMENT{Initialise accumulators} \FOR[Accumulate first summand in
        \eqref{eq:BestResponseApproximation}]{$j \gets 1$ \TO $i$}
        \STATE
        $\vec{v} \gets \vec{v} - \func{grad}(\vec{u}(\vec{\lambda}, \vec{w}), \vec{w}, \func{grad\_outputs} = \vec{v})$
        \STATE $\vec{p} \gets \vec{p} + \vec{v}$ \ENDFOR \COMMENT{Now
          $\vec{p} \approx \at{\pderiv{\mathcal{L}_{V}}{\vec{w}} \left( \pderiv{\vec{u}}{\vec{w}} \right)^{-1}}{\vec{\lambda}, \vec{w}}$}
        \STATE
        $\vec{g}_{\textrm{indirect}} = - \func{grad}(\vec{u}(\vec{\lambda}, \vec{w}), \vec{\lambda}, \func{grad\_outputs} = \vec{p})$
        \\ \COMMENT{Now
          $\vec{g}_{\textrm{indirect}} \approx - \at{\pderiv{\mathcal{L}_{V}}{\vec{w}} \left( \pderiv{\vec{u}}{\vec{w}} \right)^{-1} \pderiv{\vec{u}}{\vec{\lambda}}}{\vec{\lambda}, \vec{w}}$}
        \STATE
        $\vec{\lambda} \gets \vec{\lambda} - \kappa \left( \at{\pderiv{\mathcal{L}_{V}}{\vec{\lambda}}}{\vec{\lambda}, \vec{w}} + \vec{g}_{\textrm{indirect}} \right)$
        \\ \COMMENT{Any gradient-based optimiser (SGD shown, LR $\kappa$)} \ENDWHILE
      \end{algorithmic}
    \end{algorithm}
  \end{minipage}
\hspace{0.1cm}
\begin{minipage}{0.40\textwidth}

\hspace{0.25cm}
  \begin{subfigure}[m]{1.00\textwidth}
  \centering
  \vspace*{-1ex}
  \includegraphics[width=0.225\textwidth]{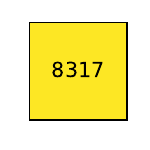}
  \caption{Baseline: \emph{Random} ($\times 1\,000$)}
  \end{subfigure}

\centering

  \begin{subfigure}[m]{1.00\textwidth}
    \centering
    \includegraphics[width=\textwidth]{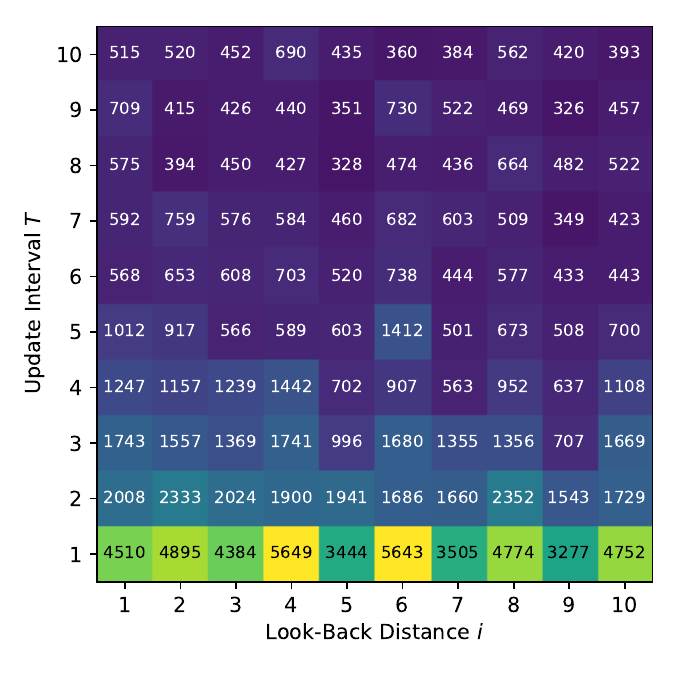}
    \vspace{-0.75cm}
    \caption{Our approach: \emph{Ours$^\textrm{WD+LR+M}$} ($\times 1\,000$)}
  \end{subfigure}
    \end{minipage}
\vspace*{-1.5ex}
\caption{Left: Pseudocode for our method. Right: Median final test loss on UCI
  Energy after 400 hyperparameter updates from random initialisations, for
  various update intervals $T$ and look-back distances $i$, with (a) no
  hyper-parameter tuning, \emph{Random}, and (b) our proposed method, \emph{Ours$^\text{WD+LR+M}$}.}\label{fig:MainAnalyticalToyAblation}
  \vspace{-2.5ex}
\end{figure}

Our empirical evaluation uses the hardware and software
detailed in Appendix~\ref{sec:ExperimentalConfiguration}, with code available at
{\small\url{https://github.com/rmclarke/OptimisingWeightUpdateHyperparameters}}.

Throughout, we train models using SGD with weight decay and momentum. We uniformly
sample initial learning rates, weight decays and momenta, using logarithmic and
sigmoidal transforms (see Appendix~\ref{sec:ExperimentDetails}), applying each
initialisation in the following eight settings:
\newpage
\begin{description}[noitemsep]
\item[Random] No HPO; hyperparameters
    held constant at their initial values, as in random search.
  \item[Random ($\times$ LR)] \emph{Random} with an extra hyperparameter increasing or
  decreasing the learning rate by multiplication after every $T=10$ update steps (emulating a hyperparameter update).
  \item[Random (3-batched)] \emph{Random} reprocessed to retain only the best
        result of every three runs. Allows \emph{Random} to exceed our
        methods' computational budgets; imitates population training.
  \item[Lorraine] \citet{lorraineOptimizingMillionsHyperparameters2020}'s method
        optimising weight decay; other hyperparameters constant.
    \item[Baydin] \citet{baydinOnlineLearningRate2018}'s method optimising
        learning rate only; other hyperparameters constant.
    \item[Ours$^\text{WD+LR}$] Algorithm~\ref{alg:OurAlgorithm} updating weight
    decay and learning rate; other hyperparameters constant.
  \item[Ours$^\text{WD+LR+M}$] Optimising all hyperparameters (adding momentum
        to \emph{Ours$^\text{WD+LR}$})
  \item[Ours$^\text{WD+HDLR+M}$] \emph{Ours$^\text{WD+LR+M}$} with independent learning rates for each model parameter
      \item[Diff-through-Opt] Optimising all hyperparameters using exact
        hypergradients \citep{domkeGenericMethodsOptimizationBased2012}
\end{description}
Any unoptimised hyperparameters are fixed at their random initial values.
Ideally, we seek resilience to poor initialisations, which realistically arise
in unguided hyperparameter selection. Hyperparameters are tuned on the validation
set; this is combined with the training set for our \emph{Random} settings, so
each algorithm observes the same data. We use the UCI/Kin8nm dataset split sizes of
\citet{galDropoutBayesianApproximation2016} and standard 60\%/20\%/20\% splits
for training/validation/test datasets elsewhere, updating hyperparameters every
$T=10$ batches with look-back distance $i = 5$ steps (except \emph{Baydin},
which has no such \emph{meta-hyperparameters}, so updates hyperparameters at every
batch \citep{baydinOnlineLearningRate2018}).

Numerical data is normalised to zero-mean, unit-variance. For efficiency,
the computational graph is detached after each hyperparameter update, so we never
differentiate through hyperparameter updates
--- essentially, back-propagation and momentum histories are truncated. Our approximate hypergradient is
passed to Adam \citep{kingmaAdamMethodStochastic2015} with
meta-learning rate $\kappa = 0.05$ and default
$\beta_{1} = 0.9, \beta_{2} = 0.999$. While these
meta-hyperparameters are not tuned,
previous work indicates performance is progressively less sensitive to higher-order
hyperparameters \citep{franceschiForwardReverseGradientBased2017,
  franceschiBilevelProgrammingHyperparameter2018,
  majumderLearningLearningRate2019}. As some settings
habitually chose unstable high learning rates, we clip these to
$[10^{-10}, 1]$ throughout.

\vspace*{-1.5ex}
\paragraph{UCI Energy: Proof of Concept}
First, we broadly illustrate Algorithm~\ref{alg:OurAlgorithm} on UCI Energy,
using a one-layer multi-layer perceptron (MLP) with 50 hidden units and ReLU
\citep{glorotDeepSparseRectifier2011} activation functions, 
trained under \emph{Ours$^\text{WD+LR}$} for 4\,000 full-batch epochs (see
Appendix~\ref{sec:AppendixProofOfConcept} for details).
Figure~\ref{fig:ToyTrajectories} shows the evolution of learning rate and weight
decay from a variety of initialisations, overlaid on the performance obtained
when all hyperparameters are kept fixed during training. Notice the
trajectories are attracted towards the region of lowest test loss, indicating
our algorithm is capable of useful learning rate and weight decay adjustment.

\subsection{UCI and Kin8nm Datasets: Establishing Robustness}
\label{sec:UCIExperiments}
\begin{table*}
  \centerfigure
  \caption{Final test MSEs after
    training UCI and Kin8nm datasets for 4\,000 full-batch epochs from each of 200 random
    hyperparameter initialisations, showing best and bootstrapped average performance.
    Uncertainties are standard errors; bold values lie in the error bars of the best
    algorithm.
  }\label{tab:UCIResults}
  \resizebox{\linewidth}{!}{
    \begin{tabular}{
    c
    S[table-format=2.2]
    U
    S[table-format=2.2]
    U
    S[table-format=1.3]
    S[table-format=2.1]
    U
    S[table-format=2.2]
    U
    S[table-format=1.2]
    S[table-format=3.2]
    U
    S[table-format=2.3]
    U
    S[table-format=2.1]
    }
    \toprule
    \multirow{2}{*}[\extrarulespace]{Method}
    & \multicolumn{5}{c}{UCI Energy}
    & \multicolumn{5}{c}{Kin8nm ($\times 1\,000$)}
    & \multicolumn{5}{c}{UCI Power} \\
    \cmidrule(lr){2-6} \cmidrule(lr){7-11} \cmidrule(lr){12-16}

    & \multicolumn{2}{c}{Mean} & \multicolumn{2}{c}{Median} & \multicolumn{1}{c}{Best}
    & \multicolumn{2}{c}{Mean} & \multicolumn{2}{c}{Median} & \multicolumn{1}{c}{Best}
    & \multicolumn{2}{c}{Mean} & \multicolumn{2}{c}{Median} & \multicolumn{1}{c}{Best} \\
    \midrule
    \input{Figures/AverageResults_UCI.tex}
    \bottomrule
  \end{tabular}
  }
\end{table*}

Next, we consider the same 50-hidden-unit MLP 
applied to seven standard UCI datasets (Boston Housing, Concrete, Energy, Naval,
Power, Wine, Yacht) and Kin8nm, in a fashion analogous to
\citet{galDropoutBayesianApproximation2016} (we do not consider dropout or
Bayesian formulations). We train 200 hyperparameter initialisations for 4\,000 full-batch epochs.
Table~\ref{tab:UCIResults} shows results for three datasets,
with complete loss evolution and distribution plots in
Appendix~\ref{sec:AllUCIResults}. Some
extreme hyperparameters caused numerical instability and NaN
final losses, which our
averages ignore. NaN results are
not problematic: they indicate extremely poor initialisations, which
should be easier for the user to rectify than merely mediocre
hyperparameters. Error bars are based on 1\,000 sets of bootstrap samples.

Given the sparse distribution of strong hyperparameter initialitions,
random sampling unsurprisingly
achieves generally poor averages. \emph{Random ($\times$ LR)} applies a learning
rate multiplier, uniformly chosen in $[0.95, 1.01]$;
these limits allow extreme initial learning rates to
revert to more typical values after 400 multiplications. This setting's poor
performance shows naïve learning rate schedules cannot
match our algorithms' average improvements. A stronger baseline is \emph{Random
  (3-batched)}, which harshly simulates the greater computational cost of our
methods by retaining only the best of every three \emph{Random} trials
(according to validation loss).
This setting comes close to, but cannot robustly beat, our methods --- a claim
reinforced by final test loss distributions
(Figure~\ref{fig:AllUCICDFs}, Appendix~\ref{sec:AllUCIResults}).

\citet{lorraineOptimizingMillionsHyperparameters2020}'s algorithm is
surprisingly indistinguishable from \emph{Random} in our trials, though it must
account for three random hyperparameters while varying only one
(the weight decay). We surmise that learning
rates and momenta are more important hyperparameters
to select, and poor choices cannot be overcome by intelligent use of weight decay.
\citet{baydinOnlineLearningRate2018}'s algorithm, however, comes closer to the
performance of \emph{Random (3-batched)}, indicating more successful intervention in
its sole optimisable hyperparameter (learning rate). Variance is also much lower than the preceding
algorithms, suggesting greater stability. That said, despite concentrating on a more
important hyperparameter, \emph{Baydin} still suffers from being unable to
control every hyperparameter.

Our scalar algorithms (\emph{Ours$^\text{WD+LR}$} and \emph{Ours$^\text{WD+LR+M}$}) appear generally
more robust to these initialisations, with average losses beating
\emph{Random}, \emph{Lorraine} and \emph{Baydin}.
Figures~\ref{fig:AllUCIEnvelopes} and \ref{fig:AllUCICDFs}
(Appendix~\ref{sec:AllUCIResults}) clearly distinguish these algorithms
from the preceding: we achieve performant results over a wider
space of initial hyperparameters. Given it
considers more hyperparameters, \emph{Ours$^\text{WD+LR+M}$}
predictably outperforms \emph{Ours$^\text{WD+LR}$}, although the difference is less prominent in certain datasets. Unlike pure HPO, 
the non-\emph{Random} algorithms in Table~\ref{tab:UCIResults} vary hyperparameters
\emph{during training}, combining aspects of HPO and
schedule learning. They are thus more flexible than conventional,
static-hyperparameter techniques, even in high dimensions
\citep{lorraineOptimizingMillionsHyperparameters2020}.

\emph{Diff-through-Opt} exactly differentiates
the current loss with respect to the hyperparameters, over
the same $i=5$ look-back window and $T=10$ update interval.
As an exact version of \emph{Ours$^\text{WD+LR+M}$} (though subject to the same
short-horizon bias), it unsurprisingly
matches the other algorithms (Figures~\ref{fig:AllUCIEnvelopes} and
\ref{fig:AllUCICDFs}, Appendix~\ref{sec:AllUCIResults}).
However, our scalar methods' proximity to this exact
baseline is reassuring given our much-reduced memory requirements and generally
comparable error bars. In these experiments, lengthening \emph{Diff-through-Opt}'s look-back horizon
 to all 4\,000 training steps, and repeating those steps for 30 hyperparameter
 updates, did not improve its performance
 (Appendix~\ref{sec:AppendixShortHorizonBias}).

\emph{Ours$^\text{WD+HDLR+M}$} theoretically mitigates short-horizon bias
\citep{wuUnderstandingShortHorizonBias2018a} by adapting appropriately to high-
and low-curvature directions. While this substantially improves performance on
some datasets, it is outperformed by scalar methods on others
(Figures~\ref{fig:AllUCIEnvelopes} and
\ref{fig:AllUCICDFs}, Appendix~\ref{sec:AllUCIResults}). We also see slightly
reduced stability, with more trials diverging to NaN final losses. Intuitively,
the risk of overfitting to the validation set is expected to increase as we
introduce more hyperparameters, which could explain this behaviour --- committing
to large learning rates in a few selected directions, as this method often does,
may prove harmful if the loss surface dynamics change suddenly. However, we
leave detailed investigation of the high-dimensional dynamics to future work.

\subsection{Large-Scale Datasets: Practical Scalability}
\vspace*{-1ex}
\paragraph{Fashion-MNIST: HPO in Multi-Layer Perceptrons}
\label{sec:FashionMNISTExperiments}

\begin{table*}[b]
  \centering
  \caption{Final test $^*$cross-entropy ($^\dagger$perplexity) on larger datasets.
    Bold values are the lowest in class.}\label{tab:LargerScaleResults}
  \resizebox{\linewidth}{!}{
    \begin{tabular}{
    c
    S[table-format=1.3]
    U
    S[table-format=1.3]
    U
    S[table-format=1.3]
    S[table-format=4.0]
    U
    S[table-format=4.0]
    U
    S[table-format=3.0]
    S[table-format=2.3]
    U
    S[table-format=1.3]
    U
    S[table-format=1.3]}
    \toprule
    \multirow{2}{*}[\extrarulespace]{Method}
    & \multicolumn{5}{c}{Fashion-MNIST$^*$}
    & \multicolumn{5}{c}{Penn Treebank$^\dagger$}
    & \multicolumn{5}{c}{CIFAR-10$^*$} \\
    \cmidrule(lr){2-6} \cmidrule(lr){7-11} \cmidrule(lr){12-16}

    & \multicolumn{2}{c}{Mean} & \multicolumn{2}{c}{Median} & \multicolumn{1}{c}{Best}
    & \multicolumn{2}{c}{Mean} & \multicolumn{2}{c}{Median} & \multicolumn{1}{c}{Best}
    & \multicolumn{2}{c}{Mean} & \multicolumn{2}{c}{Median} & \multicolumn{1}{c}{Best} \\
    \midrule
    \input{Figures/AverageResults_LargeScale.tex}
    \bottomrule
  \end{tabular}
  }
\end{table*}

We train the same single 50-unit hidden layer MLP on 10 epochs of Fashion-MNIST
\citep{xiaoFashionMNISTNovelImage2017}, using
50-sample batches. Table~\ref{tab:LargerScaleResults} and
Figure~\ref{fig:FashionMNISTResults} (Appendix~\ref{sec:AppendixLargerScaleResults})
show average test set cross-entropies over 100 initialisations.
Clearly-outlying final losses
(above $10^{3}$) are set to NaN to stop them dominating our error bars.

Echoing Section~\ref{sec:UCIExperiments},
for arbitrary hyperparameter initialisations, our methods generally converge
more robustly to lower losses, even when (as in
Figure~\ref{fig:FashionMNISTResults}, Appendix~\ref{sec:AppendixLargerScaleResults})
NaN solutions are included in our statistics. Importantly, we see
mini-batches provide sufficient gradient information for our HPO task.
\emph{Diff-through-Opt}'s failure to beat our methods is surprising; we
suppose noisy approximate gradients may
regularise our algorithms, preventing them from seeking the short-horizon optimum
so directly, thus mitigating short-horizon bias (see our sensitivity study in
Section~\ref{sec:MiscellaneousStudies}). \emph{Diff-through-Opt} with long look-back horizons does not
improve performance for equal computation (Appendix~\ref{sec:AppendixShortHorizonBias}).
Finally, median and best loss evolution plots are shown in
Figures~\ref{fig:FashionMNISTEvolution} and \ref{fig:FashionMNISTBestEvolution},
the latter including results for a Bayesian Optimisation baseline. For more
details, see Appendices~\ref{sec:BayesianOptimisationBaselines} and
\ref{sec:AppendixLargerScaleResults}.

\vspace*{-1ex}
\paragraph{Penn Treebank: HPO in Recurrent Networks}
\label{sec:PennTreebankExperiments}

Now, we draw inspiration from \citet{lorraineOptimizingMillionsHyperparameters2020}'s
large-scale trials:
a 2-layer, 650-unit LSTM \citep{hochreiterLongShortTermMemory1997} with
learnable embedding, trained on the standard
Penn Treebank-3-subset benchmark dataset \citep{marcusTreebank31999} for 72
epochs. To focus our study,
we omit the dropout, activation regularisation and predefined learning rate schedules used by \citeauthor{lorraineOptimizingMillionsHyperparameters2020}, though we retain
training gradient clipping to a Euclidean norm of $0.25$. Training
considers length-70 subsequences of 40 parallel sequences, using 50 random
hyperparameter initialisations.

Table~\ref{tab:LargerScaleResults} and Figure~\ref{fig:PennTreebankResults}
(Appendix~\ref{sec:AppendixLargerScaleResults}) show final test perplexities. They reflect
our intuition that adjusting progressively
more hyperparameters reduces average test losses, continuing the
trend we have seen thus far. Highly bimodal final loss
distributions for some algorithms cause wide bootstrap-sampled error bars.
Learning rate adjustments show particular gains:
\emph{Our} algorithms and \emph{Diff-through-Opt} perform particularly well in less-optimal
configurations.

\vspace*{-1ex}
\paragraph{CIFAR-10: HPO in Convolutional Networks}
Finally, to demonstrate scalability, we train a ResNet-18
\citep{heDeepResidualLearning2016} on CIFAR-10 \citep{krizhevskyLearningMultipleLayers2009} for 72
epochs. We use the unaugmented dataset (since unbiased data
augmentation over both training and validation datasets exhausts our GPU memory) and
optimise hyperparameters as before, using 100-image batches.

Table~\ref{tab:LargerScaleResults} and Figure~\ref{fig:CIFAR10Results}
(Appendix~\ref{sec:AppendixLargerScaleResults}) show our results. Our
gains are now more marginal, with this setting
apparently robust to its initialisation, though the general ranking of
algorithms remains similar, and we retain useful improvements over our baselines.
However, our best final accuracies fall short of state-of-the-art, suggesting a more
intricate and clever setting may yield further performance gains.

\subsection{Miscellaneous Studies}
\vspace{-0.1cm}
\label{sec:MiscellaneousStudies}
\paragraph{Experiment Runtimes}
\label{sec:ExperimentRuntimes}

\begin{figure*}
  \centering
  \vspace*{-1ex}
  \begin{subfigure}{0.32\textwidth}
    \centering
    \includegraphics[width=\textwidth]{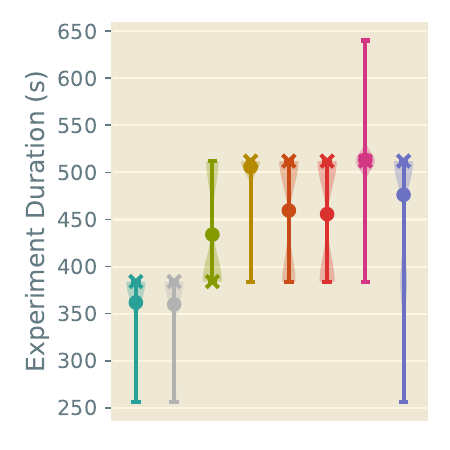}
    \caption{Single-Pass Runtimes}\label{fig:FashionMNISTDuration}
  \end{subfigure}
  \hfill
  \begin{subfigure}{0.32\textwidth}
    \centering
    \includegraphics[width=\textwidth]{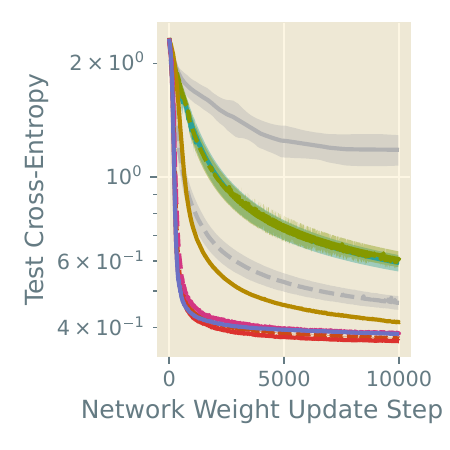}
    \caption{Median Loss Evolution}\label{fig:FashionMNISTEvolution}
  \end{subfigure}
  \hfill
  \begin{subfigure}{0.32\textwidth}
    \centering
    \includegraphics[width=\textwidth]{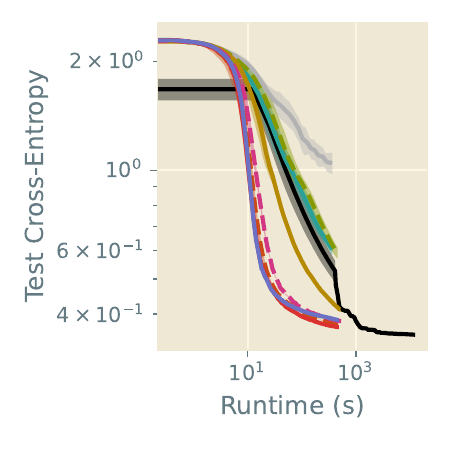}
    \caption{Best Loss Evolution}\label{fig:FashionMNISTBestEvolution}
  \end{subfigure}

  \begin{tabular}{ccccc}
    \linepatch{Cyan} & \linepatch{Grey} & \linepatch[dashed]{Grey} &
                                                                     \linepatch{Green} & \linepatch{Yellow} \\[-1ex]
    Random & Random ($\times$ LR) & Random (3-batched) & Lorraine & Baydin
  \end{tabular}
  \begin{tabular}{ccccc}
    \linepatch{Orange} & \linepatch{Red} & \linepatch{Magenta} & \linepatch{Violet} & \linepatch{Black} \\[-1ex]
    Ours$^\text{WD+LR}$ & Ours$^\text{WD+LR+M}$ & Ours$^\text{WD+HDLR+M}$ & Diff-through-Opt & Bayesian Optimisation
  \end{tabular}

  \DeclareRobustCommand\meankey{%
    \tikz[baseline=-0.75ex, scale=0.1]{
      \draw[fill=black] (0, 0) circle (1);
      \draw[very thick, color=black] (-2.5, 0) -- (2.5, 0);}}
  \DeclareRobustCommand\mediankey{%
    \tikz[baseline=-0.75ex, scale=0.1]{
      \draw[black, very thick] (-1, -1) -- (1, 1) (-1, 1) -- (1, -1);
      \draw[very thick, color=black] (-2.5, 0) -- (2.5, 0);}}
  \begin{tabular}{ccc}
    \meankey~Mean Runtime & \mediankey~Median Runtime & \envelopekey~$\pm$ Standard Error
  \end{tabular}
\vspace{-0.1cm}
  \caption{Illustrations of training a one-layer 50-unit MLP on
    Fashion-MNIST over 100 random hyperparameter initialisations. We include a
    \emph{Bayesian Optimisation} baseline from
    Appendix~\ref{sec:BayesianOptimisationBaselines} and loss evolution plots
    from Appendix~\ref{sec:AppendixLargerScaleResults}.}
  \label{fig:FashionMNISTSummaryPlot}
\end{figure*}

For our larger-scale experiments, we illustrate comparative runtimes in
Figure~\ref{fig:DurationPlots} (Appendix~\ref{sec:LargeScaleRuntimes},
where they are discussed in detail). Note from our Fashion-MNIST sample
(Figure~\ref{fig:FashionMNISTDuration}) that all HPO algorithms have a lower
computational cost than naïvely training two typical fixed hyperparameter initialisations,
despite achieving substantial HPO effect. This compares extremely favourably
to HPO methods relying on repeated retraining.
Additional experiments in Appendix~\ref{sec:ASHAPBTExperiments} combine
\emph{Ours$^\text{WD+LR+M}$} with Asynchronous
Hyperband \citep{liSystemMassivelyParallel2020} or Population-Based Training
\citep{jaderbergPopulationBasedTraining2017}, showing improved and unchanged
performance/time trade-offs, respectively.

\paragraph{UCI Energy: Sensitivity Study}

Finally, we consider a range of update intervals $T$ and look-back distances $i$
on UCI~Energy, performing 400 hyperparameter updates from 100 random
initialisations on each, using \emph{Ours$^\text{WD+LR+M}$}. We plot the median
final test losses for each choice of $T$ and $i$ to the right of
Figure~\ref{fig:MainAnalyticalToyAblation}, and also the median performance
with no HPO (\emph{Random}), which we outperform in every case.
Performance generally improves with larger $T$ ---
likely because the total number of weight updates increases with $T$, since the
number of hyperparameter
updates is fixed at 400. While
increasing $i$ gives more terms in our approximating series
\eqref{eq:BestResponseApproximation}, these extra terms become more and more biased:
they are evaluated at the current weight values instead of at the progressively more
different past weight values.
We theorise this trade-off explains the more complex relationship between $i$
and final performance.
More details are given in Appendix~\ref{sec:AppendixSensitivityStudy}.

\section{Conclusion and Future Work}

We have presented an algorithm for optimising continuous
hyperparameters, specifically those appearing in a differentiable weight update,
including optimiser hyperparameters. Our method requires only a single training
pass, has a motivating true-hypergradient
convergence argument, and demonstrates practical benefits at a range of
experimental scales without greatly sacrificing training time. We also
tackle traditionally-intractable per-parameter learning rate optimisation
on non-trivial datasets. However, this setting surprisingly underperformed its
scalar counterpart; further work is necessary to understand this result.

As future work, our myopic approach could be extended to longer horizons by incorporating the principles of recent
work by \citet{micaelliNongreedyGradientbasedHyperparameter2020}, which presents
a promising research direction. We also depend inconveniently on
meta-hyperparameters, which are not substantially tuned.
Ultimately, we desire systems which are completely independent
of human configuration, thus motivating investigations into the removal of these
settings.

\newpage
\section*{Societal Impact / Ethics Statement}
\label{sec:SocietalImpact}
Our work fits into the broad subfield of automatic machine learning (AutoML),
which aims to use automation to
strip away the tedious work that is necessary
to implement practical ML systems. Our method focuses on automating the
oft-labelled `black art' of optimising hyperparameters. This contributes towards
the democratisation of ML techniques, which we hope will
improve  accessibility to non-experts.

However, our method also has some associated risks. For one, developers of unethical
machine learning applications (e.g.\ for mass surveillance, identity theft or automated weaponry) may
use our techniques to improve their systems' performance. The heavily
metric-dominated nature of our field
creates additional concerns ---
for instance, end-users of our method may not appreciate that optimising naïvely
for training and validation loss alone may result in dangerously poorly-trained
or unethical models if the chosen metric,
developer's intentions and moral good do not align.

More broadly, users may rely excessively on HPO techniques to
optimise their models' performance, which can lead to poor results and inaccurate comparisons if
the HPO strategy is imperfect. Further, reducing the need to
consider hyperparmeter tuning
abstracts away an important component of how machine learning methods work in
practice. Knowledge of this component may then become less accessible,
inhibiting understanding and future research insights from the wider community.

Our datasets are drawn from standard benchmarks,
and thus should not introduce new societal risks. Similarly, our work aims to
decrease the computational burden of HPO, which should mitigate the
environmental impact of ML training --- an ever more important goal in our
increasingly environmentally-conscious society.

\section*{Reproducibility Statement}
All datasets we use are publicly available; for the Penn Treebank
dataset, we provide
a link in Table~\ref{tab:DatasetLicences} (Appendix~\ref{sec:DatasetLicences}). What little data
processing we perform is fully explained in the corresponding subsection of
Section~\ref{sec:Experiments}.

Our mathematical arguments are presented fully and with disclosure of all
assumptions in Section~\ref{sec:VerboseDerivations}.

Source code for all our experiments is provided to reviewers, and is made
available on GitHub ({\small\url{https://github.com/rmclarke/OptimisingWeightUpdateHyperparameters}}). This source code contains a complete
description of our experimental environment, configuration files and
instructions on the reproduction of our experiments.

\section*{Acknowledgements}
We acknowledge computation provided by the Cambridge Service for Data
Driven Discovery (CSD3) operated by the University of Cambridge Research
Computing Service ({\small\url{www.csd3.cam.ac.uk}}), provided by Dell EMC and Intel
using Tier-2 funding from the Engineering and Physical Sciences Research Council
(capital grant EP/P020259/1), and DiRAC funding from the Science and Technology
Facilities Council ({\small\url{www.dirac.ac.uk}}).

We thank Austin Tripp for his feedback on drafts of this paper.

Ross Clarke acknowledges funding from the Engineering and Physical Sciences
Research Council (project reference 2107369, grant EP/S515334/1).

\newpage
\bibliography{ZoteroLibrary}
\bibliographystyle{myicml2021}

\newpage
\appendix
\section{Notes}
\label{sec:AppendixNotes}
\subsection{Experimental Configuration}
\label{sec:ExperimentalConfiguration}
\begin{table}[h]
  \centering
  \caption{System configurations used to run our experiments.}
  \label{tab:ExperimentalHardware}
  \resizebox{\linewidth}{!}{
  \begin{tabular}{cccccc}
    \toprule
    Type & CPU & GPU (NVIDIA) & Python & PyTorch & CUDA \\
    \midrule
    Consumer Desktop & Intel Core i7-3930K & RTX 2080GTX  & 3.9.7 & 1.8.1 & 10.1\\
    Local Cluster & Intel Core i9-10900X & RTX 2080GTX & 3.7.12 & 1.8.1 & 10.1 \\
    \makecell{Cambridge Service for\\Data Driven Discovery (CSD3)*} & AMD EPYC 7763 & Ampere A100 & 3.7.7 & 1.10.1 & 11.1\\
    \bottomrule
  \end{tabular}
  }

  {\small * \url{www.csd3.cam.ac.uk}}
\end{table}
Our experiments are variously performed on one of three sets of hardware, as
detailed in Table~\ref{tab:ExperimentalHardware}. While system configurations
varied between experiments, this does not impair comparability. All Penn Treebank
and CIFAR-10 experiments were performed on the CSD3 Cluster; otherwise, all
experiments comparing runtimes were performed on the Local Cluster. We make use of
GPU acceleration throughout, with the PyTorch
\citep{paszkePyTorchImperativeStyle2019} and Higher
\citep{grefenstetteGeneralizedInnerLoop2019} libraries.

On CIFAR-10 only, a minor bug in our data processing code meant the mean and
standard deviation used to normalise the dataset were incorrect by a factor of
255. As batch normalisation in the ResNet-18 model largely mitigates this error,
and every algorithm was provided with the same mal-normalised data, this does
not threaten the validity of our results. We retain the incorrect transformation
in our published code for reproducibility, but have clearly annotated the
affected lines.

\subsection{Dataset Licences}
\label{sec:DatasetLicences}
\begin{table}[h]
  \centering
  \caption{Licences under which we use datasets in this work.}
  \label{tab:DatasetLicences}
  \begin{tabular}{ccc}
    \toprule
    Dataset & Licence & Source \\
    \midrule
    UCI Boston & \multirow{8}{*}{Unknown; widely used} & \multirow{8}{*}{\citet[Github]{galDropoutBayesianApproximation2016}}\\
    UCI Concrete \\
    UCI Energy \\
    Kin8nm \\
    UCI Naval \\
    UCI Power \\
    UCI Wine \\
    UCI Yacht \\
    \midrule
    Fashion-MNIST & MIT & PyTorch via \verb|torchvision|\\
    Penn Treebank & Proprietary; fair use subset, widely used & *\\
    CIFAR-10 & No licence specified & PyTorch via \verb|torchvision| \\
    \bottomrule
  \end{tabular}

  {\small * \url{http://www.fit.vutbr.cz/\~imikolov/rnnlm/simple-examples.tgz}}
\end{table}
Our datasets are all standard in the ML literature. For
completeness, we outline the licences under which they are used in Table~\ref{tab:DatasetLicences}.

\subsection{Expanded Intuitive Reinterpretation of Iterative Optimisation}
\label{sec:ExpandedIntuitiveReinterpretation}
In this section, we take the opportunity to give a more verbose elaboration on
the ideas presented in
Section~\ref{sec:ReinterpretationOfIterativeOptimisation}, which seeks to
address methodological concerns raised by
\citet{lorraineOptimizingMillionsHyperparameters2020} and reinterpret iterative
optimisation in light of our derived algorithm.

\citet{lorraineOptimizingMillionsHyperparameters2020} base their derivations on
the training loss $\mathcal{L}_{T}$, which does not depend on optimiser
hyperparameters such as the learning rate or momentum coefficient.
Differentiating $\mathcal{L}_{T}$ with respect to these hyperparameters thus
cannot provide any useful information for computing their gradient-based
updates. Similarly, the Implicit Function Theorem cannot be applied to motivate
their approximation in this case, since $\mathcal{L}_{T}$ is not an implicit
function of these hyperparameters. Consequently,
\citeauthor{lorraineOptimizingMillionsHyperparameters2020} declare their method
is methodologically restricted to operate on hyperparameters on which the
training loss depends directly.

In contrast, we work with an arbitrary update $\vec{u}$, which \emph{does} then depend
on the optimiser hyperparameters, allowing us to take the corresponding
derivatives and apply the Implicit Function Theorem. However, the optimisation
objective remains the same, so we should rightly be suspicious that the
methodological issue raised by
\citeauthor{lorraineOptimizingMillionsHyperparameters2020} may still apply. To
address these remaining concerns, we look to recast our development in the form
of \citet{lorraineOptimizingMillionsHyperparameters2020}. We do this by defining
a `pseudo-loss' $\overline{\mathcal{L}}$ such that our weight updates in
\eqref{eq:GeneralWeightUpdate} are equivalent to
\begin{equation}
  \vec{w} \gets \vec{w} - \at{\pderiv{\overline{\mathcal{L}}}{\vec{w}}}{\vec{\lambda}, \vec{w}}.
\end{equation}
We must acknowledge that our optimisation objective is
now for the update $\vec{u}$ to be zero (equivalent to a stationary point of
$\overline{\mathcal{L}}$), rather than the training loss $\mathcal{L}_{T}$ being
at a minimum. But with that caveat, we may apply the machinery of
\citeauthor{lorraineOptimizingMillionsHyperparameters2020} directly: we have
constructed a `loss function' which depends directly on all the hyperparameters,
so may legitimately deploy their strategy. By comparison with
\eqref{eq:GeneralWeightUpdate}, we have that, for any particular problem,
\begin{equation}\label{eq:GeneralPseudoLoss}
  \overline{\mathcal{L}} = \int \vec{u}(\vec{\lambda}, \vec{w}) \,\mathrm{d}\vec{w}.
\end{equation}

Often, in simple cases, the mathematical construction in \eqref{eq:GeneralPseudoLoss}
has a geometric interpretation. Consider the elementary case of vanilla SGD with
learning rate $\eta$, which gives the update rule
$\vec{u}_{\textrm{SGD}} = \eta \pderiv{\mathcal{L}_{T}}{\vec{w}}$. The required
pseudo-loss is then $\overline{\mathcal{L}} = \eta \mathcal{L}_{T}$, which suggests
two equivalent geometric interpretations of the optimisation problem we are
solving.

In the first (conventional) interpretation, we have a fixed loss
surface $\mathcal{L}_{T}$ whose minimum we seek. To optimise $\eta$ in this
problem, we imagine standing on the loss surface at our current point, then
consider the effect of taking different-sized steps across it. Clearly, this
interpretation decouples $\eta$ from the underlying function $\mathcal{L}_{T}$, so
suffers from \citeauthor{lorraineOptimizingMillionsHyperparameters2020}'s
methodological issue.

In the second (our novel) interpretation, we seek the
minimum of the loss surface $\overline{\mathcal{L}}$, which we may traverse only
by taking steps of precisely the magnitude of the local gradient. Now, when we
optimise $\eta$, we are \emph{rescaling} the surface beneath our update step, and
consequently also rescaling the update step size. In this way, we have coupled
$\eta$ and $\mathcal{L}_{T}$ into one objective $\overline{\mathcal{L}}$, which
varies with $\eta$ and so adheres to the methodological restrictions noted by
\citeauthor{lorraineOptimizingMillionsHyperparameters2020}. The key idea is thus
that we have reinterpreted the problem to allow optimiser hyperparameters to
dynamically transform the objective surface, which legitimises our derivation.

When we consider SGD with momentum, the situation becomes more complicated,
because there is not an immediately obvoius closed form for
$\overline{\mathcal{L}}$. Suppose we are descending a steep-sided valley of the
original loss surface $\mathcal{L}_{T}$. The momentum acts to `pull' the update
step in the direction it has accumulated in its buffer (down the axis of the
valley), so it acts to augment gradients in the valley's downhill direction, and
diminishes the effect of gradients in other directions.

If we are working with
the pseudo-loss $\overline{\mathcal{L}}$, we need to achieve the same effect,
but we do not have direct control over the update step size, which is determined
by the local gradient of $\overline{\mathcal{L}}$. So if the update step is
aligned with the accumulated momentum, we must `squash' the loss surface about
our current point in this direction, such that our fixed update step lands at a
point further down the valley in the original, unscaled space of
$\mathcal{L}_{T}$. Accordingly, if the update step is \emph{mis}aligned with the
accumulated momentum, this squashing in the direction of the momentum buffer must be
combined with a `stretching' in the update step direction, so that we make
\emph{less} progress in that direction of the original space. In aggregate, this
is a complex and intricate transformation of the space, such that we bring the
`correct' new point (as specified by the optimiser) to lie underneath the update
step given by $-\pderiv{\overline{\mathcal{L}}}{\vec{w}}$.

To summarise, the key principle we seek to convey in
Section~\ref{sec:ReinterpretationOfIterativeOptimisation} is that of
creating a new pseudo-loss surface $\overline{\mathcal{L}}$, which is a
transformed version of $\mathcal{L}_{T}$, such that the update steps taken by a
hyperparameter-free optimiser over the pseudo-loss $\overline{\mathcal{L}}$ are
equivalent to those taken by the true optimiser over the original training loss
$\mathcal{L}_{T}$.

\subsection{Expanded Discussion of Update Function
  \texorpdfstring{$\vec{u}(\vec{\lambda}, \vec{w})$}{u(λ, w)}}
\label{sec:ExpandedUpdateFunction}
Throughout our derivation, we refer to an arbitrary weight update function
$\vec{u}(\vec{\lambda}, \vec{w})$. While we have not expressed this in its most
general form for clarity, we outline here that this formulation supports a wide
array of non-trivial weight update optimisation algorithms beyond SGD.

$\vec{u}(\vec{\lambda}, \vec{w})$ is precisely the quantity which is added to the
network weights $\vec{w}$ during an update step. For instance, in the PyTorch
framework \citep{paszkePyTorchImperativeStyle2019}, the pre-implemented
optimisers first compute some update tensor $\vec{\Delta w}$, then perform the final
operation $\vec{w} \gets \vec{w} - \vec{\Delta w}$ to update the weights. Thus, we have
$\vec{u}(\vec{\lambda}, \vec{w}) = \vec{\Delta w}$, and automatic differentiation trivially
gives the necessary derivatives to apply our algorithm.

Some optimisers have internal state, for instance the momentum buffer of SGD
with momentum and the two rolling average buffers of Adam. Although this
dependency is not notated in our presentation of $\vec{u}$, it does not
invalidate our development --- the only extra care required is to detach these
buffers from the computational graph after each hyperparameter update. For
PyTorch's implementation of SGD with learning rate $\eta$, momentum coefficient $\mu$
and weight decay coefficient $\xi$, we thus have
\begin{align}
  \vec{v} \gets \mu \vec{v} + \pderiv{\mathcal{L}_{T}}{\vec{w}} + \xi \vec{w}
  &&
  \vec{u}_{\mathrm{momentum}} = \eta \vec{v} \punc{,}
\end{align}
and need simply take care to detach $\vec{v}$ from the computational graph after
each update. For Adam, we may read the definition of $\vec{u}$ straight from its
original paper \citep{kingmaAdamMethodStochastic2015}; in their notation:
\begin{equation}
  \vec{u}_{\mathrm{Adam}} = \alpha \cdot \frac{\hat{m}_{t}}{\sqrt{\hat{v}_{t}} + \epsilon} \punc{,}
\end{equation}
where $t$ indexes time and $\hat{m}_{t}, \hat{v}_{t}$ are internal states, and
we simply detach those states from the computational graph periodically.
Those states depend directly on the additional states $m_{t}, v_{t}$, and
ultimately on the gradient $\nabla_{\theta} f_{t} (\theta_{t-1})$ (which is
$\at{\pderiv{\mathcal{L}_{T}}{\vec{w}}}{\vec{w}_{i-1}}$ in our notation).

\newpage
\section{Additional Results}
\label{sec:AppendixAdditionalResults}

\subsection{UCI Energy: Proof of Concept}
\label{sec:AppendixProofOfConcept}
Our proof of concept graphic (Figure~\ref{fig:ToyTrajectories}) is constructed
using the same experimental
configuration as Section~\ref{sec:UCIExperiments}. Training 500 random
initialisations of the model, weight decay and learning rate for 4\,000 network
weight update steps on \emph{Random} (without any HPO) yields source data for the heatmap
forming the background of Figure~\ref{fig:ToyTrajectories}. The heatmap itself
is the predictive mean of a Gaussian process
\citep{williamsGaussianProcessesRegression1996} using the sum of an RBF and
white noise kernel. We fit the Gaussian process in log (input and output) space
using the defaults of the Scikit-learn implementation
\citep{pedregosaScikitlearnMachineLearning2011}, which automatically select
hyperparameters for the kernels. Any NaN data points are replaced with a large final test loss (150) for the purpose of computing this regression. We then train a
small number of additional initialisations using \emph{Ours$^\text{WD+LR}$} and plot these
trajectories. As always, the training set provides network weight updates, the
validation set provides hyperparameter updates and our results plot final performance
on the test set. While we do not expect the optimal trajectory during training
to coincide with gradients of our background heatmap, the latter provides useful
context.

Generally, given our limited training budget, larger learning rates improve
final performance up to a point, then cause instability if taken to
extremes, and the movement of our trajectories reflects this. Many
trajectories select large weight decay coefficients before doubling back on
themselves to smaller weight decays --- suggestive of our algorithm using the
weight decay to avoid converging to specific minima. Trajectories generally
converge to the valley of good test performance at a learning rate of around
$10^{-1}$ and weight decay below $10^{-2}$, further demonstrating appropriate
behaviour. In short, we are able to
make sensible and interpretable updates to the hyperparameters of this problem,
which lends support to the applicability of our method.

\subsection{UCI Energy: Hypergradient Studies}
\label{sec:AppendixSensitivityStudy}

Our sensitivity studies are based on a similar experimental configuration to
Section~\ref{sec:UCIExperiments}: for each choice of update interval $T$ and
look-back distance $i$, we generate 100 random initialisations of weight decay,
learning rate and momentum.
Each initialisation is trained for 400
hyperparameter steps, so increasing $T$ permits more network weight updates;
averaging over the 100 final test losses yields each result cell in
Figure~\ref{fig:AnalyticalToyAblations} and
Figure~\ref{fig:AnalyticalToyHypergradients}. As in
Section~\ref{sec:UCIExperiments}, we consider the \emph{Ours$^\text{WD+LR+M}$} and
\emph{Diff-through-Opt} algorithms, optimising hyperparameters using Adam with
hyper-learning rate $\kappa = 0.05$ and truncating gradient histories at the previous
hyperparameter update.

\newpage
\subsubsection{Sensitivity of Update Interval and Look-Back Distance}

\begin{figure*}[h]
  \centering
  \begin{subfigure}[b]{0.30\textwidth}
    \centering
    \includegraphics[width=0.5\textwidth]{Figures/Sensitivity_Random.pdf}
    \caption{\emph{Random} ($\times 1000$)}
  \end{subfigure}

  \begin{subfigure}[b]{0.49\textwidth}
    \includegraphics[width=\textwidth]{Figures/Sensitivity_OursLRMomentum.pdf}
    \caption{\emph{Ours$^\text{WD+LR+M}$} ($\times 1000$)}
    \label{fig:OurAnalyticalToyAblation}
  \end{subfigure}
  \hfill
  \begin{subfigure}[b]{0.49\textwidth}
    \includegraphics[width=\textwidth]{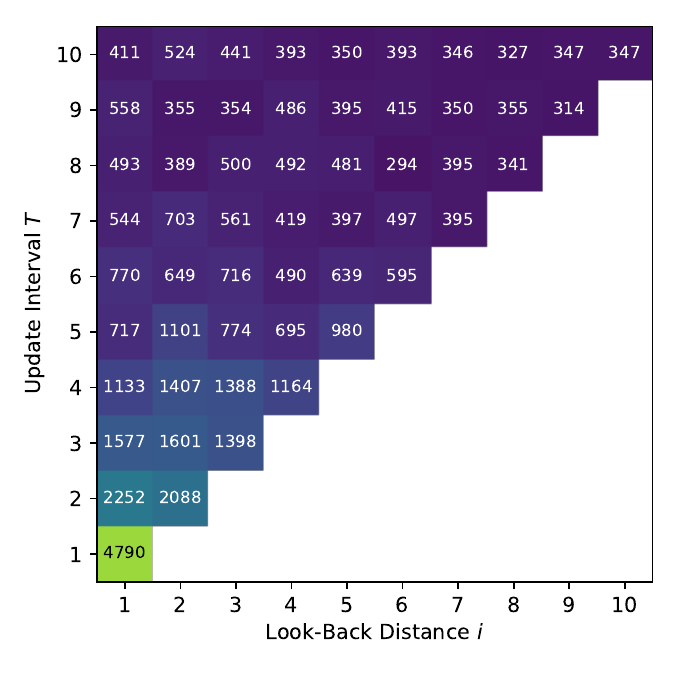}
    \caption{\emph{Diff-through-Opt} ($\times 1000$)}
    \label{fig:DiffThroughOptAnalyticalToyAblation}
  \end{subfigure}
  \caption{Median final test loss over 100 repetitions, after optimising
    UCI Energy for 400 hyperparameter update steps, from a variety of
    update intervals $T$ and look-back distances $i$.}\label{fig:AnalyticalToyAblations}
\end{figure*}

Figure~\ref{fig:AnalyticalToyAblations}
shows our final performance results in this setting: our algorithm's strongest
configurations (Figure~\ref{fig:OurAnalyticalToyAblation}) 
compare favourably with results obtained using the exact hypergradients of
\emph{Diff-through-Opt} (Figure~\ref{fig:DiffThroughOptAnalyticalToyAblation}).
Noting that the conditions necessary to guarantee our algorithm's convergence to
the best-response derivative (specifically that the current network weights are
optimal for the current hyperparameters) are almost certainly not met in
practice, it is unsurprising that \emph{Diff-through-Opt} achieves greater
performance. This difference is much smaller, however, than that between both
HPO algorithms and the \emph{Random} base case; given our substantial efficiency
benefits over \emph{Diff-through-Opt}, we make an acceptable compromise
between accuracy and scalability.
Further, these results justify our choice of $T=10$ and $i=5$ to match the
setting of \citet{lorraineOptimizingMillionsHyperparameters2020}. That said,
recall these figures enforce consistency of number of hyperparameter updates,
not of total computational expense; there remains a case for using large
numbers of less precise hyperparameter updates (small $T$ and $i$), which may
promote faster convergence in practice.

Our results illustrate the trade-off in selecting a look-back distance $i$. 
In principle, increasing $i$ includes more terms in our Neumann series approximation
to the best-response derivative \eqref{eq:BestResponseApproximation}, which
should result in closer convergence to the true derivative. But our derivation
of this expression replaces the sequence $\vec{w}_0, \vec{w}_1, \cdots, \vec{w}_{i-1}$
from \eqref{eq:UnrolledUpdate} with $\vec{w}_i$ at all time steps. Because the $j$th 
summation step of \eqref{eq:UnrolledUpdate} involves a product of terms in $\vec{w}_{i-1}$ 
back to $\vec{w}_{i-1-j}$, the approximation of each summation step in 
\eqref{eq:BestResponseApproximation} becomes less and less accurate as $j$ increases, 
since we expect earlier model weights to become more different from $\vec{w}_i$.
A larger $i$ causes us to continue the summation up to larger $j$, so introduces
progressively less accurate terms into our approximations. Thus, increasing $i$
is only of value while this inaccuracy is counteracted by the improved convergence
of our summation, giving the behaviour we see here.

\subsubsection{Accuracy of Hypergradients}

\begin{figure*}[h]
  \centering
  \begin{subfigure}[b]{0.49\textwidth}
    \includegraphics[width=\textwidth]{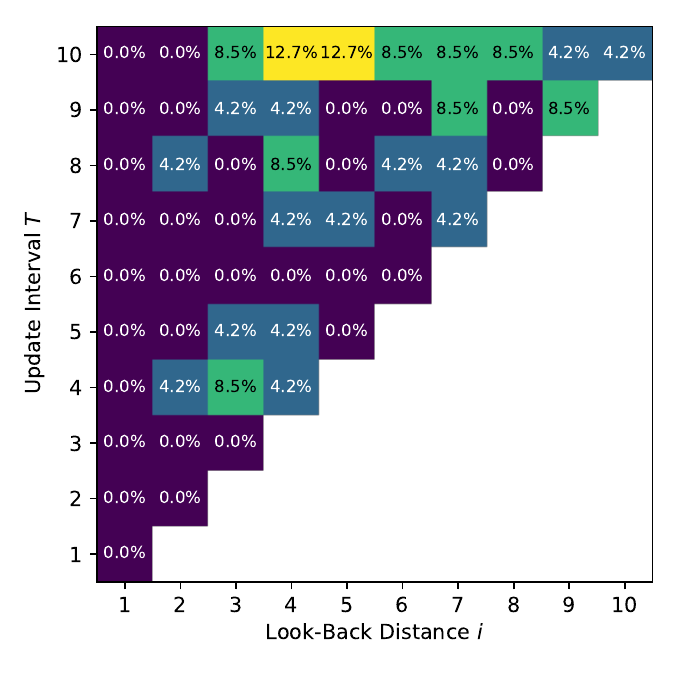}
    \caption{Learning Rate}
  \end{subfigure}
  \hfill
  \begin{subfigure}[b]{0.49\textwidth}
    \includegraphics[width=\textwidth]{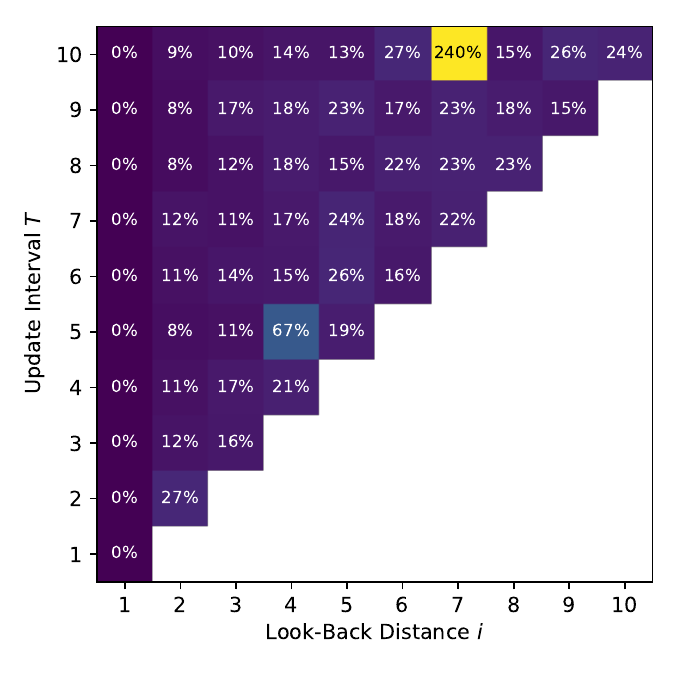}
    \caption{Weight Decay}
  \end{subfigure}
  \caption{Mean absolute error in our approximate hypergradients, based on the
    first hyperparameter updates of 50 repetitions for each cell, using UCI
    Energy. We vary the update interval $T$ and look-back distance $i$ to
    investigate the sensitivity of the problem to these meta-hyperparameters.
    Hypergradients are shown separately for each
    hyperparameter.}\label{fig:AnalyticalToyHypergradients}
\end{figure*}

We use this same experimental setting to evaluate the accuracy of our
approximate hypergradients, but disable learning rate clipping so as to allow
the natural behaviour of each algorithm to be displayed. . Our approximate and
exact settings are only
guaranteed to be trying to compute the same hypergradient at the very first
hyperparameter update, so we consider this time step only. Using those
first hyperparameter updates from 50 random initialisations of the hyperparameters
and model, we compute the absolute error of the approximate
hypergradient (\emph{Ours$^\text{WD+LR+M}$}) as a percentage of the exact hypergradient
(\emph{Diff-through-Opt}), then plot the means of these errors in Figure~\ref{fig:AnalyticalToyHypergradients}.

Notice our learning rate hypergradients do not exhibit any particularly strong trends over the
meta-hyperparameters we consider, but our approximate weight decay hypergradients
seem to benefit from a shorter look-back distance. Some
of the latter average
proportional errors are large; a closer examination of the error distributions reveals
these cells have outlying results at high errors. We hypothesise that the
generally very small weight decay coefficients present more opportunity for
numerical computational error than the generally larger learning rates, in which
case hypergradients for different hyperparameters may have markedly different accuracies.
Indeed, the apparently quantised errors in the learning rate hypergradients
indicate precision truncation is influencing the results of this study.
However, our previous coarser studies suggest that, despite being approximate,
our method does not suffer catastrophically from inaccurate hypergradients.

\clearpage
\subsection{Experiment Details}
\label{sec:ExperimentDetails}

\begin{table}[h]
  \centering
  \caption{Transformation and initialisation parameters used in our
    hyperparameter optimisation.}
  \label{tab:HyperparameterTransformations}
  \begin{tabular}{ccS[table-format=-2]S[table-format=-1]cS[table-format=1e-1]S[table-format=1.2]}
    \toprule
    \multirow{2}{*}[\extrarulespace]{Hyperparameter} & \multirow{2}{*}[\extrarulespace]{\makecell{Initialisation\\Space}} & \multicolumn{2}{c}{\makecell{Initialisation\\Range}} &
                                                                   \multirow{2}{*}[\extrarulespace]{\makecell{Optimisation\\
                                                                   Space}} & \multicolumn{2}{c}{Natural
                                                                           Range}\\
    \cmidrule(lr){3-4} \cmidrule(lr){6-7}
    & & {Min} & {Max} & & {Min} & {Max} \\
    \midrule
    Learning Rate & Logarithmic & -6 & -1 & Logarithmic & 1e-6 & 0.1 \\
    Weight Decay & Logarithmic & -7 & -2 & Logarithmic & 1e-7 & 0.01 \\
    Momentum & Natural & 0 & 1  & Sigmoidal & 0 & 1 \\
    \bottomrule
  \end{tabular}
\end{table}

Since the hyperparameters associated with SGD only take meaningful values on
subsets of the real line, we apply gradient descent to transformed versions of
the hyperparameters. For non-negative hyperparameters, such as learning rate and
weight decay, we work with the base-10 logarithm of the actual value; for momentum, which
lies in the unit interval $[0, 1]$, we work with
the inverse-sigmoid of the actual value. These transformed search spaces mean a
naïve implementation of gradient descent will not update hyperparameters to
nonsensical values.

Our hyperparameter initialisations are drawn uniformly from a broad range,
intended to be a superset of typical values to permit analysis of the realistic
case where we have little intuition for the optimal choices. Thus, our
hyperparameter ranges include typical values for a variety of applications,
while giving each algorithm the opportunity to briefly stretch into more extreme
values should these be useful on occasion. Where we have applied
a logarithmic transformation to a hyperparameter (learning rate and weight
decay), we draw samples uniformly in
logarithmic space; where inverse-sigmoid transforms are used (momentum), we draw
samples in natural space to ensure adequate exploration of
values near 0 and 1. Table~\ref{tab:HyperparameterTransformations} summarises
our initialisation and optimisation strategy.

Throughout, we seek robustness to a range of different initialisations, which
is indicated by average final performance across a variety of initial hyperparameter
choices. We include metrics of best performance to show each algorithm is capable of 
reaching the same vicinity as the best possible case, but this is not the focus of our 
work. Intuitively, a more naïve algorithm (e.g.\ \emph{Random}) could stumble across
the best possible final model weights, in which case no other algorithm would be able
to beat it; this would not necessarily mean we would prefer the \emph{Random} algorithm 
in a general setting.

\subsection{UCI and Kin8nm Datasets}
\label{sec:AllUCIResults}

\begin{figure*}
  \centering
  \begin{subfigure}{0.32\textwidth}
    \centering
    \includegraphics[width=\textwidth]{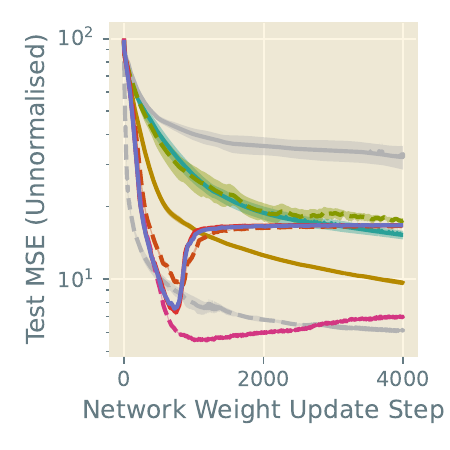}
    \caption{UCI Boston}\label{fig:UCIBostonEnvelope}
  \end{subfigure}
  \hfill
  \begin{subfigure}{0.32\textwidth}
    \centering
    \includegraphics[width=\textwidth]{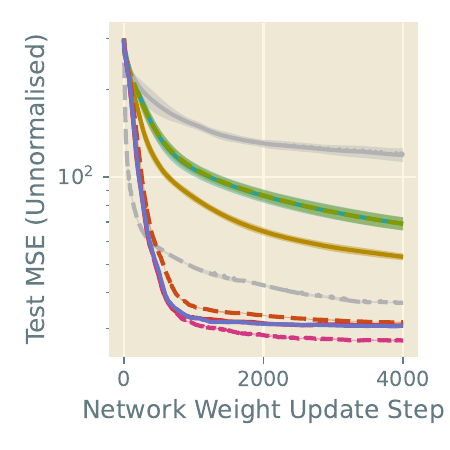}
    \caption{UCI Concrete}\label{fig:UCIConcreteEnvelope}
  \end{subfigure}
  \hfill
  \begin{subfigure}{0.32\textwidth}
    \centering
    \includegraphics[width=\textwidth]{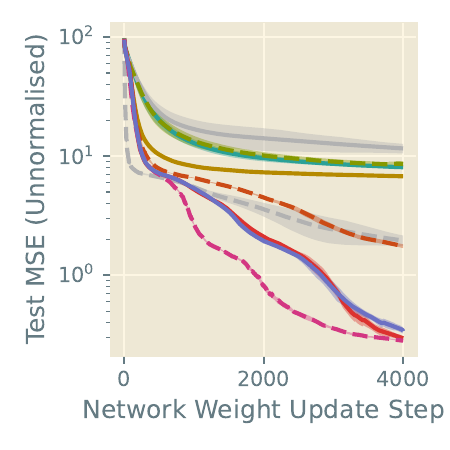}
    \caption{UCI Energy}\label{fig:UCIEnergyEnvelope}
  \end{subfigure}
  \hfill
  \begin{subfigure}{0.32\textwidth}
    \centering
    \includegraphics[width=\textwidth]{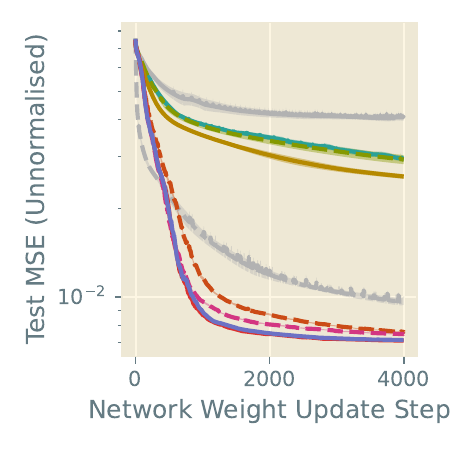}
    \caption{Kin8nm}\label{fig:UCIKin8nmEnvelope}
  \end{subfigure}
  \hfill
  \begin{subfigure}{0.32\textwidth}
    \centering
    \includegraphics[width=\textwidth]{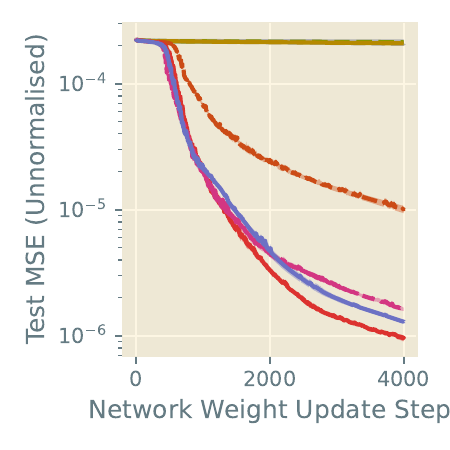}
    \caption{UCI Naval}\label{fig:UCINavalEnvelope}
  \end{subfigure}
  \hfill
  \begin{subfigure}{0.32\textwidth}
    \centering
    \includegraphics[width=\textwidth]{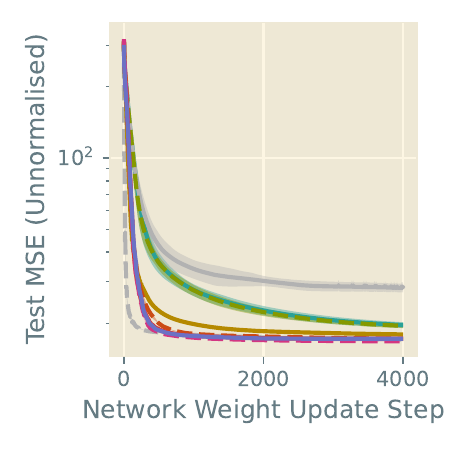}
    \caption{UCI Power}\label{fig:UCIPowerEnvelope}
  \end{subfigure}
  \hfill
  \begin{subfigure}{0.32\textwidth}
    \centering
    \includegraphics[width=\textwidth]{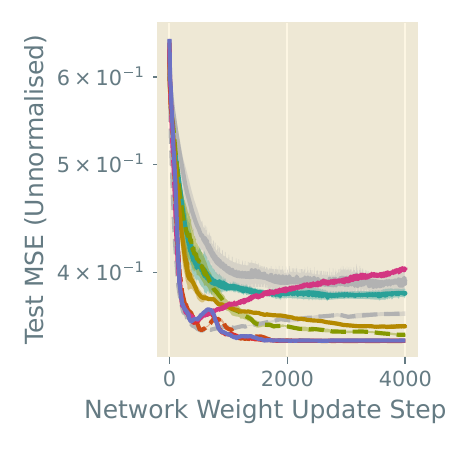}
    \caption{UCI Wine}\label{fig:UCIWineEnvelope}
  \end{subfigure}
  \hfill
  \begin{subfigure}{0.32\textwidth}
    \centering
    \includegraphics[width=\textwidth]{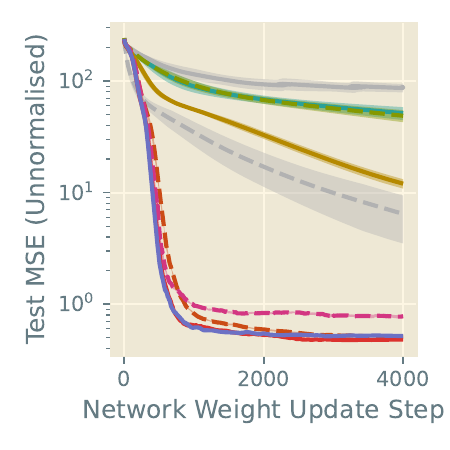}
    \caption{UCI Yacht}\label{fig:UCIYachtEnvelope}
  \end{subfigure}
  \hfill
  \begin{subfigure}{0.32\textwidth}
    \centering
    \begin{tabular}{cl}
      \linepatch{Cyan} & Random \\
      \linepatch{Grey} & Random ($\times$ LR)\\
      \linepatch[dashed]{Grey} & Random (3-batched)\\
      \linepatch[dashed]{Green} & Lorraine \\
      \linepatch{Yellow} & Baydin \\
      \linepatch[dashed]{Orange} & Ours$^\text{WD+LR}$ \\
      \linepatch{Red} & Ours$^\text{WD+LR+M}$ \\
      \linepatch[dashed]{Magenta} & Ours$^\text{WD+HDLR+M}$ \\
      \linepatch{Violet} & Diff-through-Opt \\[1ex]
      \envelopekey & $\pm$ Standard Error
    \end{tabular}
  \end{subfigure}
  \caption{Evolution of test MSEs during training of UCI and Kin8nm datasets for
    4\,000 full-batch epochs from each of 200 random initialisations of SGD
    learning rate, weight decay and momentum. We bootstrap sample our results to
    construct a median test MSE estimator at each update step and its standard error.}
  \label{fig:AllUCIEnvelopes}
\end{figure*}

\begin{figure*}
  \centering
  \begin{subfigure}{0.32\textwidth}
    \centering
    \includegraphics[width=\textwidth]{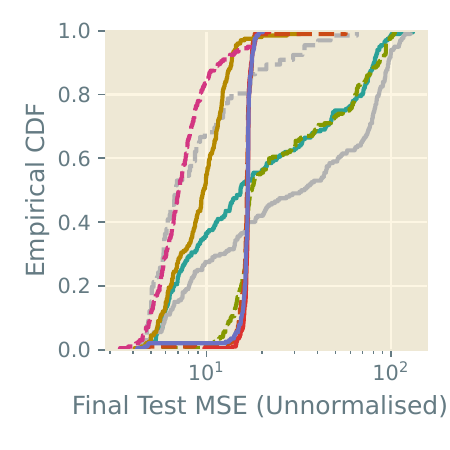}
    \caption{UCI Boston}\label{fig:UCIBostonResults}
  \end{subfigure}
  \hfill
  \begin{subfigure}{0.32\textwidth}
    \centering
    \includegraphics[width=\textwidth]{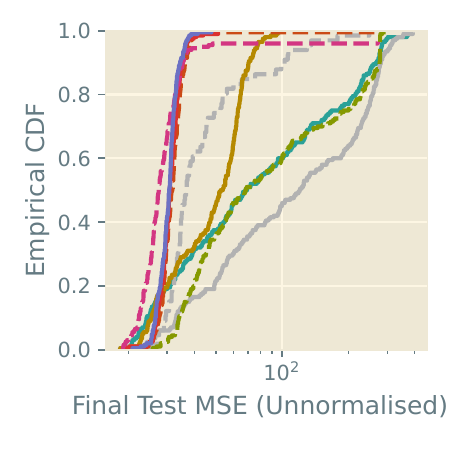}
    \caption{UCI Concrete}\label{fig:UCIConcreteResults}
  \end{subfigure}
  \hfill
  \begin{subfigure}{0.32\textwidth}
    \centering
    \includegraphics[width=\textwidth]{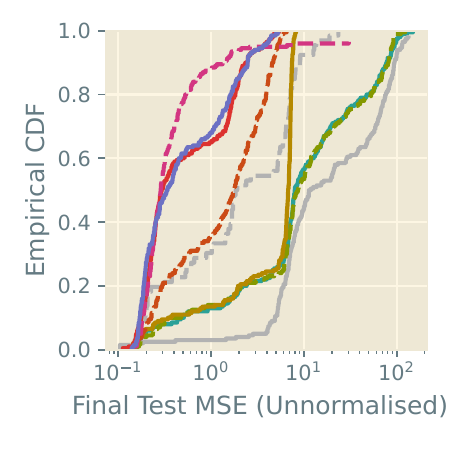}
    \caption{UCI Energy}
  \end{subfigure}
  \hfill
  \begin{subfigure}{0.32\textwidth}
    \centering
    \includegraphics[width=\textwidth]{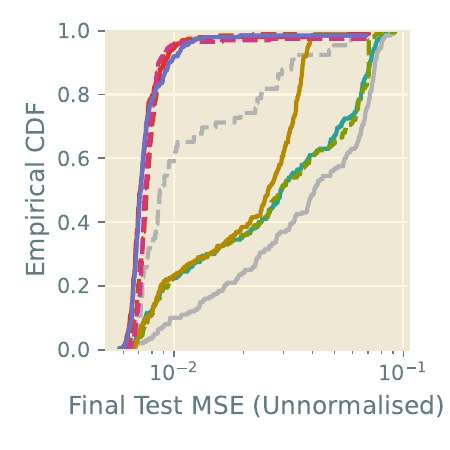}
    \caption{Kin8nm}
  \end{subfigure}
  \hfill
  \begin{subfigure}{0.32\textwidth}
    \centering
    \includegraphics[width=\textwidth]{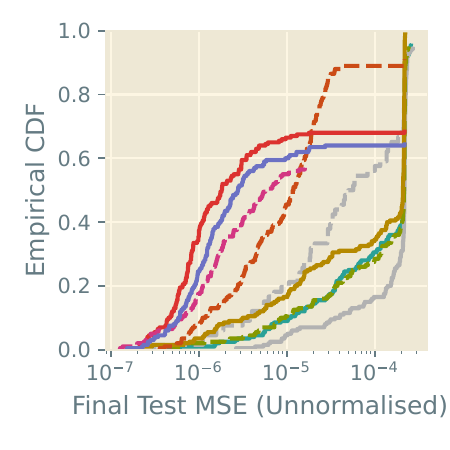}
    \caption{UCI Naval}\label{fig:UCINavalResults}
  \end{subfigure}
  \hfill
  \begin{subfigure}{0.32\textwidth}
    \centering
    \includegraphics[width=\textwidth]{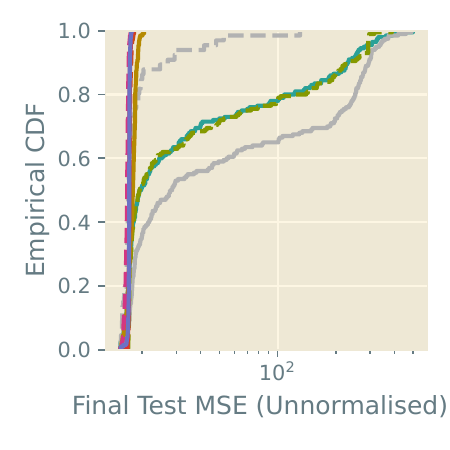}
    \caption{UCI Power}
  \end{subfigure}
  \hfill
  \begin{subfigure}{0.32\textwidth}
    \centering
    \includegraphics[width=\textwidth]{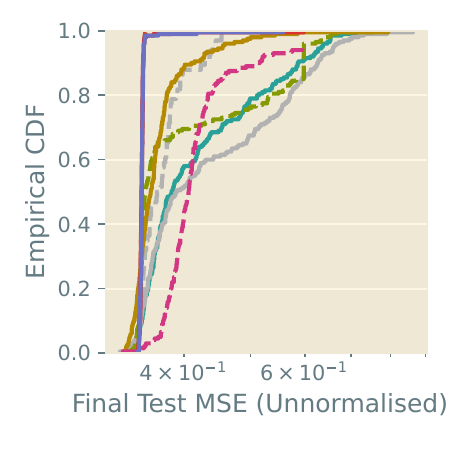}
    \caption{UCI Wine}\label{fig:UCIWineResults}
  \end{subfigure}
  \hfill
  \begin{subfigure}{0.32\textwidth}
    \centering
    \includegraphics[width=\textwidth]{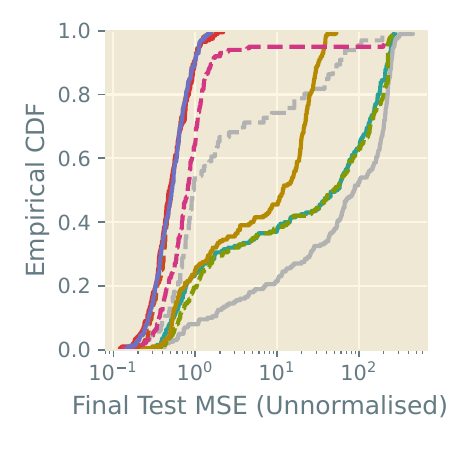}
    \caption{UCI Yacht}\label{fig:UCIYachtResults}
  \end{subfigure}
  \hfill
  \begin{subfigure}{0.32\textwidth}
    \centering
    \begin{tabular}{cl}
      \linepatch{Cyan} & Random \\
      \linepatch{Grey} & Random ($\times$ LR)\\
      \linepatch[dashed]{Grey} & Random (3-batched)\\
      \linepatch[dashed]{Green} & Lorraine \\
      \linepatch{Yellow} & Baydin \\
      \linepatch[dashed]{Orange} & Ours$^\text{WD+LR}$ \\
      \linepatch{Red} & Ours$^\text{WD+LR+M}$ \\
      \linepatch[dashed]{Magenta} & Ours$^\text{WD+HDLR+M}$ \\
      \linepatch{Violet} & Diff-through-Opt \\
    \end{tabular}
  \end{subfigure}
  \caption{Empirical CDFs of final test losses, after training UCI and Kin8nm datasets for
    4\,000 full-batch epochs from each of 200 random initialisations of SGD
    learning rate, weight decay and momentum. Runs ending with NaN
    losses account for the CDFs not reaching $1.0$. Higher curves are better;
    note the logarithmic horizontal scales.}
  \label{fig:AllUCICDFs}
\end{figure*}

\begin{figure*}
  \centering
  \begin{subfigure}{0.32\textwidth}
    \centering
    \includegraphics[width=\textwidth]{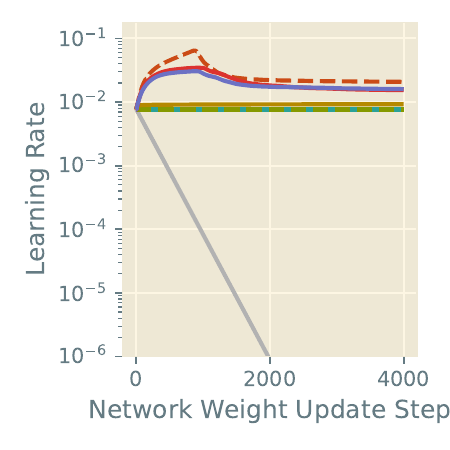}
    \caption{UCI Boston}
  \end{subfigure}
  \hfill
  \begin{subfigure}{0.32\textwidth}
    \centering
    \includegraphics[width=\textwidth]{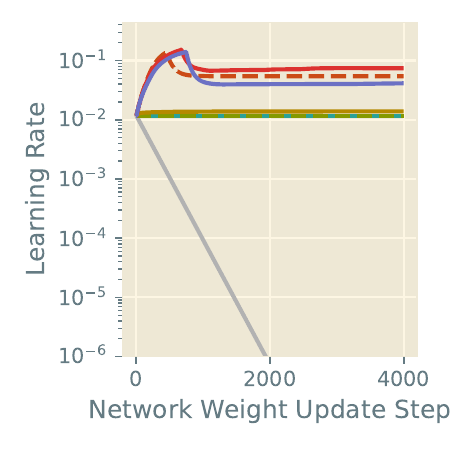}
    \caption{UCI Concrete}
  \end{subfigure}
  \hfill
  \begin{subfigure}{0.32\textwidth}
    \centering
    \includegraphics[width=\textwidth]{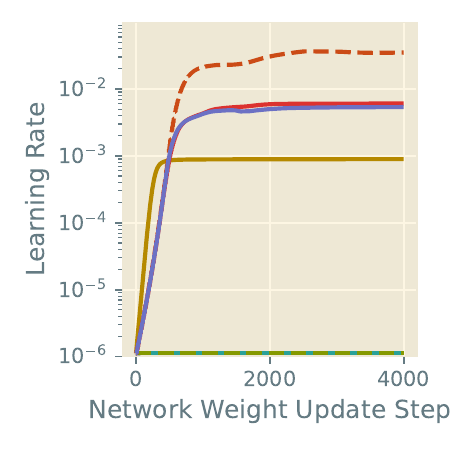}
    \caption{UCI Energy}
  \end{subfigure}
  \hfill
  \begin{subfigure}{0.32\textwidth}
    \centering
    \includegraphics[width=\textwidth]{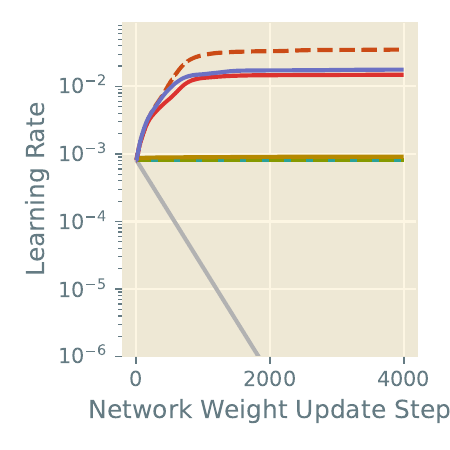}
    \caption{Kin8nm}
  \end{subfigure}
  \hfill
  \begin{subfigure}{0.32\textwidth}
    \centering
    \includegraphics[width=\textwidth]{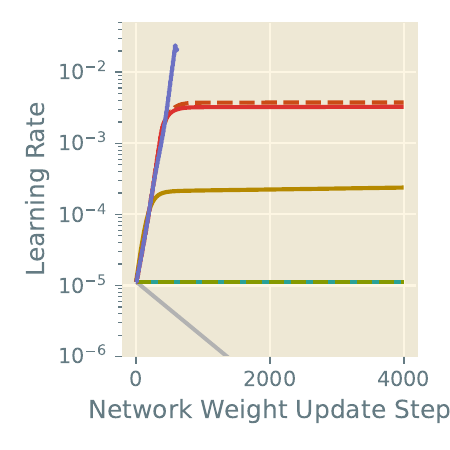}
    \caption{UCI Naval}
  \end{subfigure}
  \hfill
  \begin{subfigure}{0.32\textwidth}
    \centering
    \includegraphics[width=\textwidth]{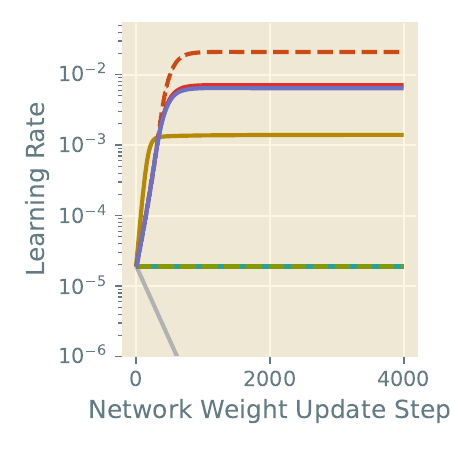}
    \caption{UCI Power}
  \end{subfigure}
  \hfill
  \begin{subfigure}{0.32\textwidth}
    \centering
    \includegraphics[width=\textwidth]{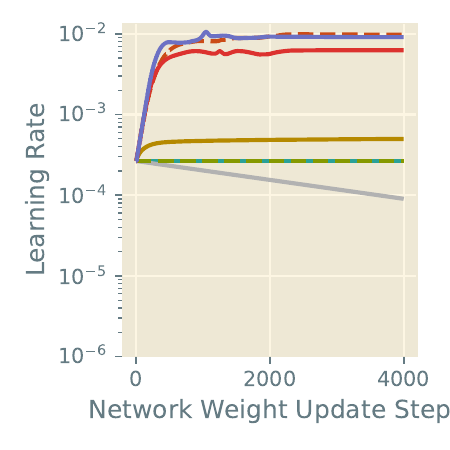}
    \caption{UCI Wine}
  \end{subfigure}
  \hfill
  \begin{subfigure}{0.32\textwidth}
    \centering
    \includegraphics[width=\textwidth]{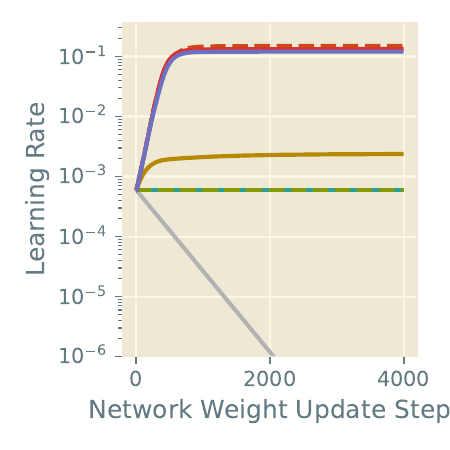}
    \caption{UCI Yacht}
  \end{subfigure}
  \hfill
  \begin{subfigure}{0.32\textwidth}
    \centering
    \begin{tabular}{cl}
      \linepatch{Cyan} & Random \\
      \linepatch{Grey} & Random ($\times$ LR)\\
      \linepatch[dashed]{Green} & Lorraine \\
      \linepatch{Yellow} & Baydin \\
      \linepatch[dashed]{Orange} & Ours$^\text{WD+LR}$ \\
      \linepatch{Red} & Ours$^\text{WD+LR+M}$ \\
      \linepatch{Violet} & Diff-through-Opt \\
    \end{tabular}
  \end{subfigure}
  \caption{Sample learning rate evolutions after training UCI and Kin8nm datasets for 4\,000
    full-batch epochs; one of 200 random initialisations is shown for each
    dataset, chosen randomly from our results.}
  \label{fig:AllUCISampleLRs}
\end{figure*}

Results from all UCI datasets we considered and Kin8nm are presented in
Figures~\ref{fig:AllUCIEnvelopes} and \ref{fig:AllUCICDFs}.
These show many of the features already discussed in
Section~\ref{sec:UCIExperiments}. We account for experimental runs ending in NaN
losses by appropriately scaling the maximum value of the CDFs in
Figure~\ref{fig:AllUCICDFs} to be less than
$1.0$. For each dataset, one random configuration is
selected, and the resulting learning rate evolutions plotted in
Figure~\ref{fig:AllUCISampleLRs}. Note that UCI Naval generated a large number
of NaN runs, which explains the prominent failure of certain algorithms to reach
the top of our CDF plots.

\clearpage
\subsection{UCI, Kin8nm and Fashion-MNIST: Bayesian Optimisation Baselines}
\label{sec:BayesianOptimisationBaselines}
\begin{table}[h]
  \caption{Repeat of Table~\ref{tab:UCIResults}: final test MSE
    on selected UCI and Kin8nm datasets. Now includes 32 repetitions of 30-sample Bayesian
    Optimisation, which is not considered in our emboldening.}
  \label{tab:BayesianOptimisationUCIResults}
  \centering
  \resizebox{\linewidth}{!}{
    \begin{tabular}{
    c
    S[table-format=2.2]
    U
    S[table-format=2.2]
    U
    S[table-format=1.4]
    S[table-format=2.1]
    U
    S[table-format=2.2]
    U
    S[table-format=1.2]
    S[table-format=3.2]
    U
    S[table-format=2.3]
    U
    S[table-format=2.1]
    }
    \toprule
    \multirow{2}{*}[\extrarulespace]{Method}
    & \multicolumn{5}{c}{UCI Energy}
    & \multicolumn{5}{c}{Kin8nm ($\times 1\,000$)}
    & \multicolumn{5}{c}{UCI Power} \\
    \cmidrule(lr){2-6} \cmidrule(lr){7-11} \cmidrule(lr){12-16}

    & \multicolumn{2}{c}{Mean} & \multicolumn{2}{c}{Median} & \multicolumn{1}{c}{Best}
    & \multicolumn{2}{c}{Mean} & \multicolumn{2}{c}{Median} & \multicolumn{1}{c}{Best}
    & \multicolumn{2}{c}{Mean} & \multicolumn{2}{c}{Median} & \multicolumn{1}{c}{Best} \\
    \midrule
    \input{Figures/AverageResults_UCI.tex}
    \midrule
    \input{Figures/AverageResults_UCI_BayesOpt.tex}
    \bottomrule
  \end{tabular}
  }
\end{table}

\begin{table}[h]
  \caption{Partial repeat of Table~\ref{tab:LargerScaleResults}: final test cross-entropy
    on Fashion-MNIST. Now includes 32 repetitions of 30-sample Bayesian
    Optimisation, which is not considered in our emboldening.}
  \label{tab:BayesianOptimisationFashionMNISTResults}
  \centering
    \begin{tabular}{
    c
    S[table-format=1.4]
    U
    S[table-format=1.4]
    U
    S[table-format=1.3]
    }
    \toprule
    \multirow{2}{*}[\extrarulespace]{Method}
    & \multicolumn{5}{c}{Fashion-MNIST} \\
    \cmidrule(lr){2-6}

    & \multicolumn{2}{c}{Mean} & \multicolumn{2}{c}{Median} & \multicolumn{1}{c}{Best} \\
    \midrule
    \input{Figures/AverageResults_Fashion-MNIST_NonBayesOpt.tex}
    \midrule
    \input{Figures/AverageResults_Fashion-MNIST_BayesOpt.tex}
    \bottomrule
  \end{tabular}
\end{table}

To contextualise our methods against the best possible performance, we also
perform 32 repetitions of 30-sample Bayesian Optimisation (BO) on selected UCI
datasets, Kin8nm and Fashion-MNIST, using the same procedure as
\emph{Random} to train each sampled configuration (but now with an separate
validation set used to compare configurations), and record these in
Tables~\ref{tab:BayesianOptimisationUCIResults} and
\ref{tab:BayesianOptimisationFashionMNISTResults}. We use the Python BO
implementation of \citet{nogueiraBayesianOptimizationOpen2014}, retaining the default
configuration that the first five of our 30 initialisations be drawn randomly
from the space. All runtimes are measured based on 8-way
parallel execution per GPU. Although BO
involves a much greater computational burden than the other methods we consider,
it should more robustly estimate the optimal hyperparameters.

Our average performance falls short of the BO benchmark, but we
come substantially closer than the \emph{Random} and \emph{Lorraine} baselines,
and similarly as close as \emph{Diff-through-Opt}. On Fashion-MNIST especially, our
methods are very competitive, achieving performance in the vicinity of BO with
30 times less computation. Considering our best runs,
we seem to suffer only a minor performance penalty (compared to the best
hyperparameters we found) by adopting Algorithm~\ref{alg:OurAlgorithm}.

\begin{figure*}
  \centering
  \begin{subfigure}[m]{0.32\textwidth}
    \includegraphics[width=\textwidth]{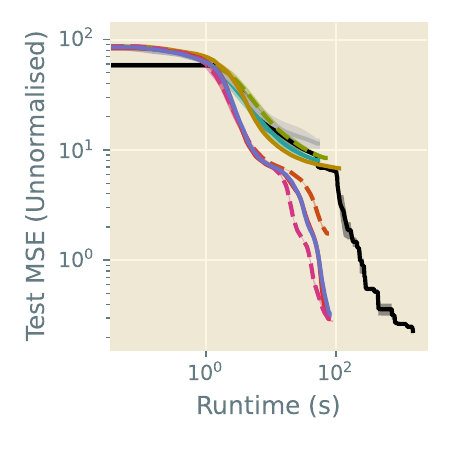}
    \caption{UCI Energy}\label{fig:UCIEnergyRuntimes}
  \end{subfigure}
  \hfill
  \begin{subfigure}[m]{0.32\textwidth}
    \includegraphics[width=\textwidth]{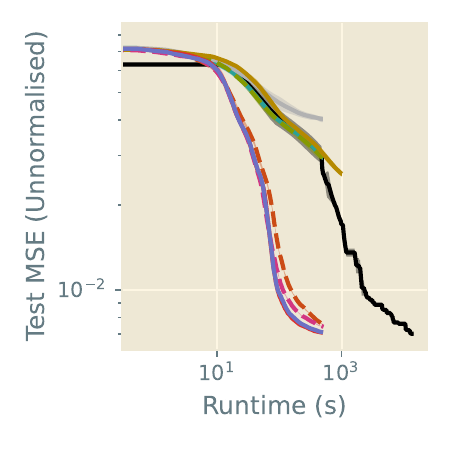}
    \caption{Kin8nm}
  \end{subfigure}
  \hfill
  \begin{subfigure}[m]{0.32\textwidth}
    \includegraphics[width=\textwidth]{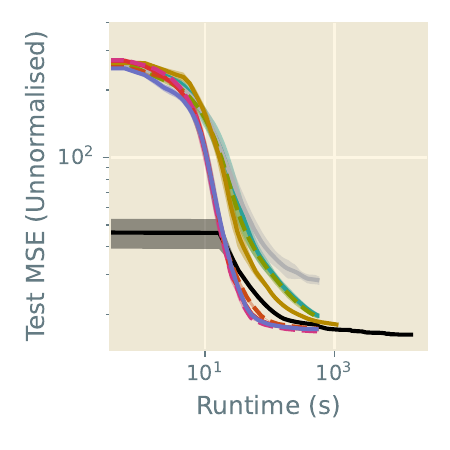}
    \caption{UCI Power}
  \end{subfigure}

  \begin{subfigure}[m]{0.49\textwidth}
    \centering
    \includegraphics[width=0.65306\textwidth]{Figures/HPOCurves_Fashion-MNIST.pdf}
    \caption{Fashion-MNIST}
  \end{subfigure}
  \begin{subfigure}{0.49\linewidth}
    \centering
    \begin{tabular}{cl}
      \linepatch{Cyan} & Random \\
      \linepatch{Grey} & Random ($\times$ LR) \\
      \linepatch[dashed]{Green} & Lorraine \\
      \linepatch{Yellow} & Baydin \\
      \linepatch[dashed]{Orange} & Ours$^\text{WD+LR}$ \\
      \linepatch{Red} & Ours$^\text{WD+LR+M}$ \\
      \linepatch[dashed]{Magenta} & Ours$^\text{WD+HDLR+M}$ \\
      \linepatch{Violet} & Diff-through-Opt \\
      \linepatch{Black} & Bayesian Optimisation \\[1ex]
      \envelopekey & $\pm$ Standard Error
    \end{tabular}
  \end{subfigure}
  \caption{Evolution of incumbent best test MSE with time, during training of
    selected UCI datasets/Kin8nm/Fashion-MNIST from 200/200/100 random hyperparameter initialisations, with
    Bayesian Optimisation for comparison. Shaded envelopes are bootstrapped
    standard errors of the best loss found at that time. Note the logarithmic horizontal axes.}
  \label{fig:HPOLearningCurves}
\end{figure*}

As an additional comparison, we show in Figure~\ref{fig:HPOLearningCurves} the
evolution of incumbent best test loss over wall time during our previous training of
200 (UCI and Kin8nm)/100 (Fashion-MNIST) initialisations. Clearly we would expect BO's
greater computation to
result in smaller final losses, which we observe, but our methods are capable of
reaching performant results much faster than BO. This underscores the principal
benefit of our algorithm: with only a single training episode and a relatively
small runtime penalty, we achieve a meaningful HPO effect, which reaches the
vicinity of the ideal performance found by much slower and more computationally
onerous methods.

\clearpage
\subsection{Large-Scale Datasets}
\label{sec:AppendixLargerScaleResults}
\begin{figure*}[h]
  \centering
  \begin{subfigure}[b]{0.32\textwidth}
    \includegraphics[width=\textwidth]{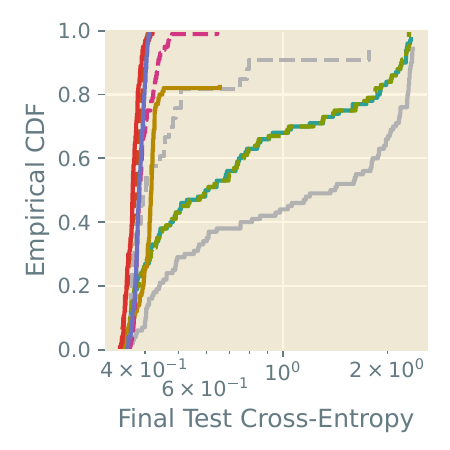}
    \caption{Fashion-MNIST using an MLP}\label{fig:FashionMNISTResults}
  \end{subfigure}
  \hfill
  \begin{subfigure}[b]{0.32\textwidth}
    \includegraphics[width=\textwidth]{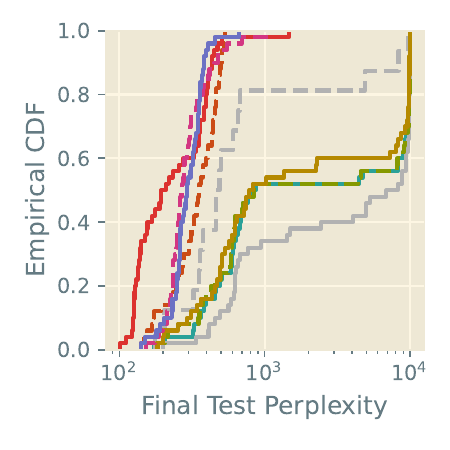}
    \caption{Penn Treebank using an LSTM}\label{fig:PennTreebankResults}
  \end{subfigure}
  \hfill
  \begin{subfigure}[b]{0.32\textwidth}
    \includegraphics[width=\textwidth]{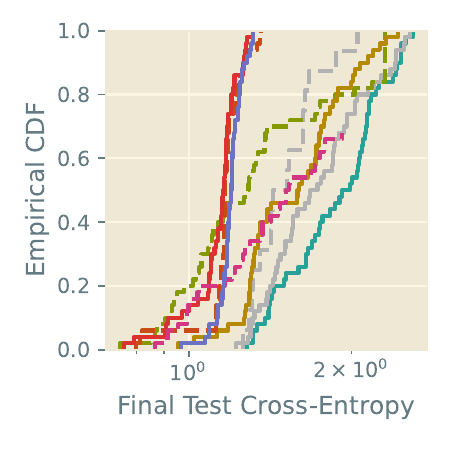}
    \caption{CIFAR-10 using a ResNet-18}\label{fig:CIFAR10Results}
  \end{subfigure}
  \caption{Empirical CDFs of final test losses, trained on larger-scale
    datasets for 50$^{\text{(a)}}$/72$^{\text{(b, c)}}$ epochs from
    100$^{\text{(a)}}$/50$^{\text{(b, c)}}$ random SGD hyperparameter
    initialisations. Key as in Figure~\ref{fig:LargerScaleErrors}; notes as in
    Figure~\ref{fig:AllUCICDFs}.}\label{fig:LargerScaleResults}
\end{figure*}

\begin{figure*}[h]
  \centering
  \begin{subfigure}[m]{0.32\textwidth}
    \includegraphics[width=\textwidth]{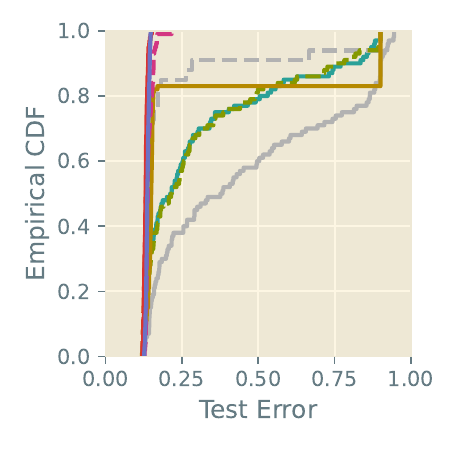}
    \caption{Fashion-MNIST using an MLP}\label{fig:FashionMNISTErrorResults}
  \end{subfigure}
  \hfill
  \begin{subfigure}[m]{0.32\textwidth}
    \includegraphics[width=\textwidth]{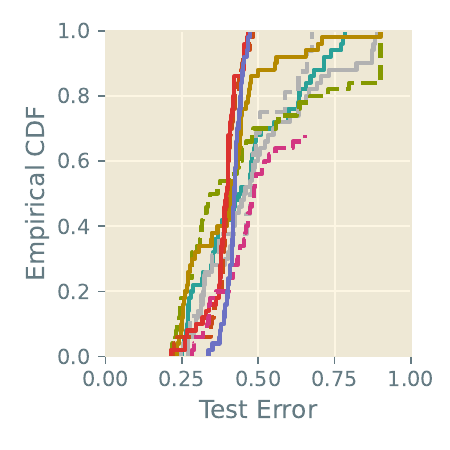}
    \caption{CIFAR-10 using a ResNet-18}
  \end{subfigure}
  \hfill
  \begin{subfigure}{0.32\textwidth}
    \begin{tabular}{cl}
      \linepatch{Cyan} & Random \\
      \linepatch{Grey} & Random ($\times$ LR)\\
      \linepatch[dashed]{Grey} & Random (3-batched)\\
      \linepatch[dashed]{Green} & Lorraine \\
      \linepatch{Yellow} & Baydin \\
      \linepatch[dashed]{Orange} & Ours$^\text{WD+LR}$ \\
      \linepatch{Red} & Ours$^\text{WD+LR+M}$ \\
      \linepatch[dashed]{Magenta} & Ours$^\text{WD+HDLR+M}$ \\
      \linepatch{Violet} & Diff-through-Opt \\
    \end{tabular}
  \end{subfigure}
  \caption{Empirical CDFs of final test errors after training on larger-scale
    datasets for 50$^{\text{(a)}}$/72$^{\text{(b)}}$ epochs from each of
    100$^{\text{(a)}}$/50$^{\text{(b)}}$ random initialisations of SGD
    hyperparameters. Notes as in Figure~\ref{fig:AllUCICDFs}.}
  \label{fig:LargerScaleErrors}
\end{figure*}

For Fashion-MNIST, Penn Treebank and CIFAR-10,
Figure~\ref{fig:LargerScaleResults} shows the
empirical distributions of final test performance, contextualising the values
shown in Table~\ref{tab:LargerScaleResults}. In addition,
Figure~\ref{fig:LargerScaleErrors} and
Table~\ref{tab:LargerScaleErrors} show our results in terms of error
percentages for Fashion-MNIST and CIFAR-10.

\begin{table*}[h]
  \centering  \caption{Quantitative analysis of final test errors from Figure~\ref{fig:LargerScaleErrors}.
    Bold values are the lowest in class.}\label{tab:LargerScaleErrors}
  \resizebox{\linewidth}{!}{
  \begin{tabular}{
    c
    S[table-format=1.4]
    U
    S[table-format=1.4]
    U
    S[table-format=1.3]
    S[table-format=1.3]
    U
    S[table-format=1.3]
    U
    S[table-format=1.3]}
    \toprule
    \multirow{2}{*}[\extrarulespace]{Method}
    & \multicolumn{5}{c}{Fashion-MNIST}
    & \multicolumn{5}{c}{CIFAR-10} \\
    \cmidrule(lr){2-6} \cmidrule(lr){7-11}

    & \multicolumn{2}{c}{Mean} & \multicolumn{2}{c}{Median} & \multicolumn{1}{c}{Best}
    & \multicolumn{2}{c}{Mean} & \multicolumn{2}{c}{Median} & \multicolumn{1}{c}{Best} \\
    \midrule
    \input{Figures/AverageErrors_LargeScale.tex}
    \bottomrule
  \end{tabular}
}
\end{table*}

\clearpage
\subsubsection{Runtimes}
\label{sec:LargeScaleRuntimes}

\begin{figure*}[h]
  \centering
  \begin{subfigure}{0.32\textwidth}
    \centering
    \includegraphics[width=\textwidth]{Figures/Runtime_Fashion-MNIST.pdf}
    \caption{Fashion-MNIST}
  \end{subfigure}
  \hfill
  \begin{subfigure}{0.32\textwidth}
    \centering
    \includegraphics[width=\textwidth]{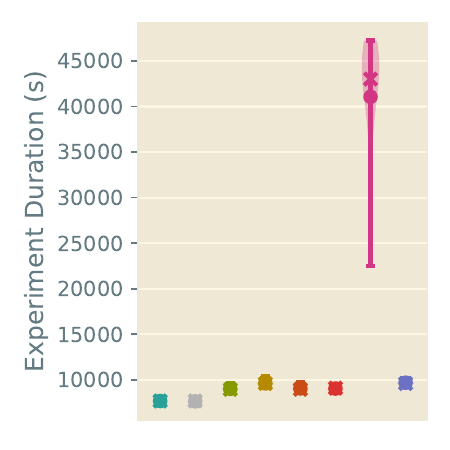}
    \caption{Penn Treebank}\label{fig:PennTreebankDuration}
  \end{subfigure}
  \hfill
  \begin{subfigure}{0.32\textwidth}
    \centering
    \includegraphics[width=\textwidth]{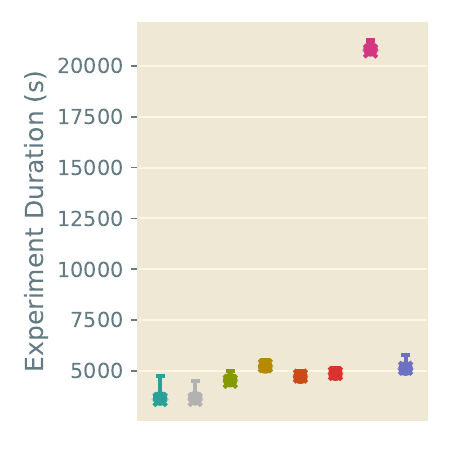}
    \caption{CIFAR-10}\label{fig:CIFAR10Duration}
  \end{subfigure}

  \begin{tabular}{cccc}
    \linepatch{Cyan} & \linepatch{Grey} & \linepatch{Green} & \linepatch{Yellow} \\[-1ex] 
    Random & \makecell{Random\\($\times$ LR)} & Lorraine & Baydin
  \end{tabular}
  \begin{tabular}{cccc}
    \linepatch{Orange} &  \linepatch{Red} & \linepatch{Magenta} & \linepatch{Violet} \\[-1ex]
    \makecell{Ours$^\text{WD+LR}$} & \makecell{Ours$^\text{WD+LR+M}$} & \makecell{Ours$^\text{WD+HDLR+M}$} & \makecell{Diff-\\through-Opt}
  \end{tabular}

  \DeclareRobustCommand\meankey{%
    \tikz[baseline=-0.75ex, scale=0.1]{
      \draw[fill=black] (0, 0) circle (1);
      \draw[very thick, color=black] (-2.5, 0) -- (2.5, 0);}}
  \DeclareRobustCommand\mediankey{%
    \tikz[baseline=-0.75ex, scale=0.1]{
      \draw[black, very thick] (-1, -1) -- (1, 1) (-1, 1) -- (1, -1);
      \draw[very thick, color=black] (-2.5, 0) -- (2.5, 0);}}
  \begin{tabular}{cc}
    \meankey~Mean Runtime & \mediankey~Median Runtime
  \end{tabular}

  \caption{Single-pass runtime distributions for our larger-scale experiments.
    Worst-case complexities are dominated by second derivative computations,
    but remain competitive with conventional multi-pass HPO
    techniques, which scale our Random results by the number of configurations sampled.}
  \label{fig:DurationPlots}
\end{figure*}

For our larger-scale experiments, we illustrate comparative runtimes in
Figure~\ref{fig:DurationPlots}. More complex model architectures, with more learnable parameters,
suffer a greater computational cost from taking second derivatives with respect
to each model weight, which appears more prominently in our Penn~Treebank and
CIFAR-10 experiments. However, our extension of
\citet{lorraineOptimizingMillionsHyperparameters2020}'s method to
additional hyperparameters doesn't dramatically increase runtimes:
second derivative computations themselves are responsible for most added
complexity. In all three cases, our scalar algorithms have runtimes within that
of naïvely training two fixed hyperparameter initialisations (such as those
considered by \emph{Random}), comparing extremely favourably to
HPO methods requiring repeated retraining. Naturally, the massive increase in
the number of hyperparameters optimised by \emph{Ours$^\text{WD+HDLR+M}$} causes a
prominent jump in runtimes, but even this algorithm remains within a practical
factor of the \emph{Random} baseline.

\clearpage
\subsubsection{Batch Normalisation and Fashion-MNIST}

\begin{figure*}[h]
  \centering
  \begin{subfigure}[m]{0.32\textwidth}
    \includegraphics[width=\textwidth]{Figures/CDF_Fashion-MNIST.pdf}
    \caption{Without Batch Normalisation}
  \end{subfigure}
  \hfill
  \begin{subfigure}[m]{0.32\textwidth}
    \includegraphics[width=\textwidth]{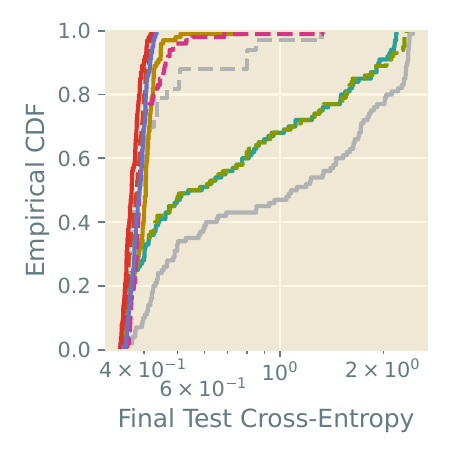}
    \caption{With Batch Normalisation}
  \end{subfigure}
  \hfill
  \begin{subfigure}[m]{0.32\textwidth}
    \begin{tabular}{cl}
      \linepatch{Cyan} & Random \\
      \linepatch{Grey} & Random ($\times$ LR)\\
      \linepatch[dashed]{Grey} & Random (3-batched)\\
      \linepatch[dashed]{Green} & Lorraine \\
      \linepatch{Yellow} & Baydin \\
      \linepatch[dashed]{Orange} & Ours$^\text{WD+LR}$ \\
      \linepatch{Red} & Ours$^\text{WD+LR+M}$ \\
      \linepatch[dashed]{Magenta} & Ours$^\text{WD+HDLR+M}$ \\
      \linepatch{Violet} & Diff-through-Opt \\
    \end{tabular}
  \end{subfigure}
  \caption{Comparing CDFs of final test cross-entropy from our Fashion-MNIST experiments
    of Section~\ref{sec:FashionMNISTExperiments}: (a) as before, without batch
    normalisation; (b) with batch normalisation in our simple MLP model.}\label{fig:FashionMNISTBatchNormComparison}
\end{figure*}

\begin{table*}[h]
  \centering
  \caption{Quantitative analysis of final test cross-entropies from
    Figure~\ref{fig:FashionMNISTBatchNormComparison} --- investigating the use of
    batch normalisation on Fashion-MNIST. Bold values are the lowest in
    class.}\label{tab:FashionMNISTBatchNormComparison}
  \resizebox{\linewidth}{!}{
  \begin{tabular}{
    c
    S[table-format=5.3]
    U
    S[table-format=1.3]
    U
    S[table-format=1.3]
    S[table-format=1.3]
    U
    S[table-format=1.3]
    U
    S[table-format=1.3]}
    \toprule
    \multirow{2}{*}[\extrarulespace]{Method}
    & \multicolumn{5}{c}{Without Batch Normalisation}
    & \multicolumn{5}{c}{With Batch Normalisation} \\
    \cmidrule(lr){2-6} \cmidrule(lr){7-11}

    & \multicolumn{2}{c}{Mean} & \multicolumn{2}{c}{Median} & \multicolumn{1}{c}{Best}
    & \multicolumn{2}{c}{Mean} & \multicolumn{2}{c}{Median} & \multicolumn{1}{c}{Best} \\
    \midrule
    \input{Figures/AverageResults_Fashion-MNIST.tex}
    \bottomrule
  \end{tabular}
  }
\end{table*}

As batch normalisation rescales the parameter gradients for each training step,
it is capable of further influencing how final performance is governed by
learning rates and their variations. To study this effect, we repeat the
Fashion-MNIST experiments described in Section~\ref{sec:FashionMNISTExperiments}
using batch normalisation in our simple MLP model. Our results are plotted
alongside the non-batch normalised CDFs in
Figure~\ref{fig:FashionMNISTBatchNormComparison}, with averages summarised in
Table~\ref{tab:FashionMNISTBatchNormComparison}.

Subjectively, performance does not seem to dramatically change when batch
normalisation is invoked, with CDFs having broadly the same shape. Considering
means and medians as our figures of merit
(Table~\ref{tab:FashionMNISTBatchNormComparison}) suggests the same general
pattern: generally, our algorithms benefit slightly from the absence of batch
normalisation, and other algorithms benefit slightly from its presence, but
there is not a unanimous trend. We conclude that, in the context of this work,
batch normalisation does not substantially improve or hinder performance.

\clearpage
\subsection{Effect of Short-Horizon Bias}
\label{sec:AppendixShortHorizonBias}
Attempts to quantify the effect of short-horizon bias in our experiments are
frustrated by the need to perform computationally-intensive longer-horizon
experiments, which has limited the scope of evaluations we can perform. However,
we are able to study the potential improvements of longer horizons in three
experimental settings.

\subsubsection{Truncated Problem, Full Horizon}
\label{sec:LongDiffThroughOptStandalone}

\begin{figure*}[h]
  \centering
  \begin{subfigure}{0.32\textwidth}
    \centering
    \includegraphics[width=\textwidth]{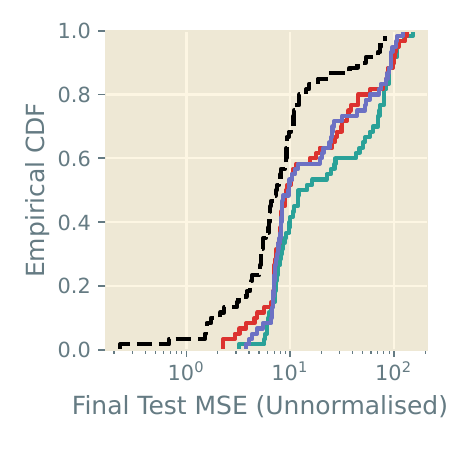}
    \caption{CDF: UCI Energy}
  \end{subfigure}
  \hfill
  \begin{subfigure}{0.32\textwidth}
    \centering
    \includegraphics[width=\textwidth]{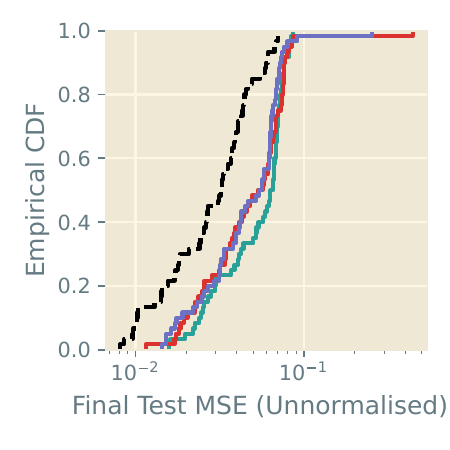}
    \caption{CDF: Kin8nm}
  \end{subfigure}
  \hfill
  \begin{subfigure}{0.32\textwidth}
    \centering
    \includegraphics[width=\textwidth]{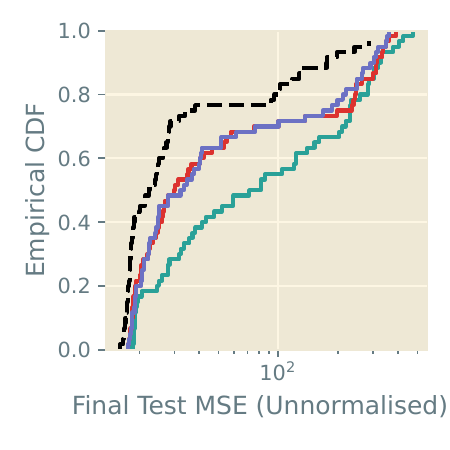}
    \caption{CDF: UCI Power}
  \end{subfigure}
  \hfill
  \begin{subfigure}{0.32\textwidth}
    \centering
    \includegraphics[width=\textwidth]{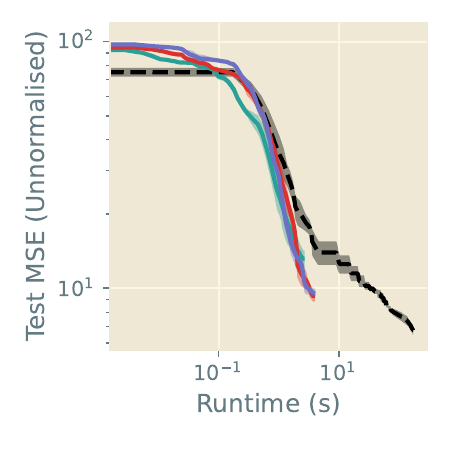}
    \caption{Evolution: UCI Energy}
  \end{subfigure}
  \hfill
  \begin{subfigure}{0.32\textwidth}
    \centering
    \includegraphics[width=\textwidth]{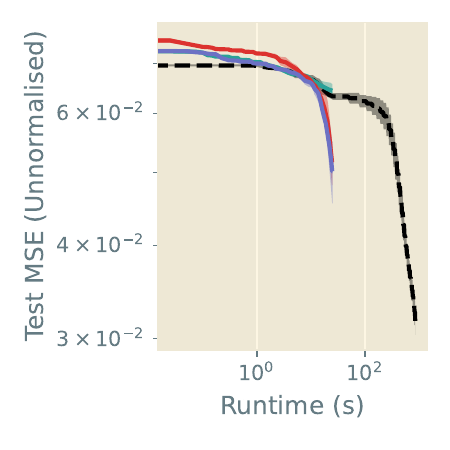}
    \caption{Evolution: Kin8nm}
  \end{subfigure}
  \hfill
  \begin{subfigure}{0.32\textwidth}
    \centering
    \includegraphics[width=\textwidth]{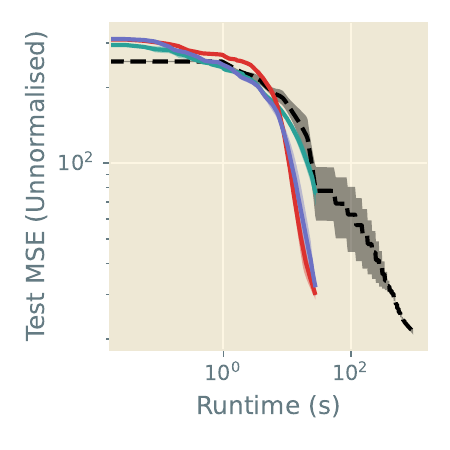}
    \caption{Evolution: UCI Power}
  \end{subfigure}
  \hfill
  \begin{subfigure}{0.32\textwidth}
    \centering
    \includegraphics[width=\textwidth]{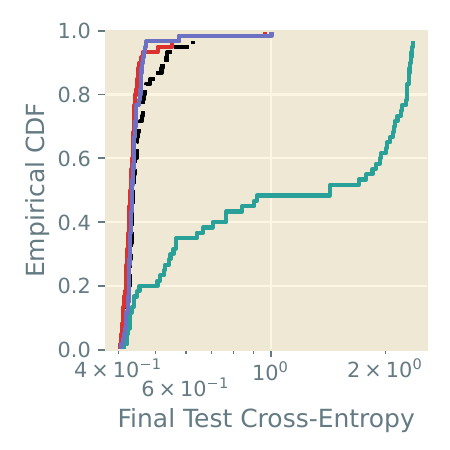}
    \caption{CDF: Fashion-MNIST (Loss)}
  \end{subfigure}
  \hfill
  \begin{subfigure}{0.32\textwidth}
    \centering
    \includegraphics[width=\textwidth]{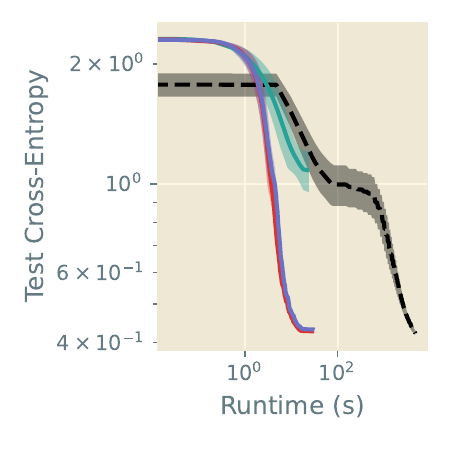}
    \caption{Evol'n: Fashion-MNIST (Loss)}
  \end{subfigure}
  \hfill
  \begin{subfigure}{0.32\textwidth}
    \centering
    \includegraphics[width=\textwidth]{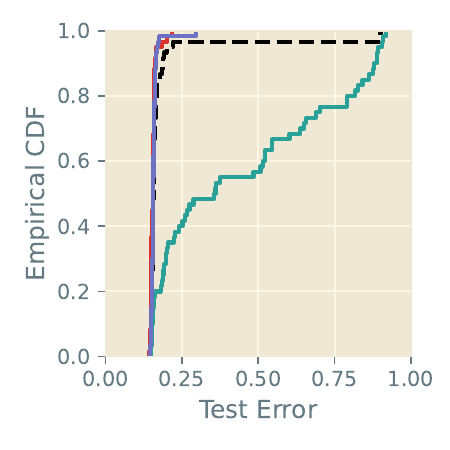}
    \caption{CDF: Fashion-MNIST (Error)}
  \end{subfigure}
  
    \centering
    \begin{tabular}{ccccc}
      \linepatch{Cyan} & \linepatch{Red} & \linepatch{Violet} &
                                                                \linepatch[dashed]{Black} & \envelopekey \\
      Random & Ours$^\text{WD+LR+M}$ & Diff-through-Opt &
                                                       Long Diff-through-Opt & Standard Error
    \end{tabular}
  \caption{CDFs of final test losses and evolutions of bootstrapped median
    losses from 60 random initialisations on a shorter problem of 200 (UCI and Kin8nm) /
    1000 (Fashion-MNIST) network weight updates, featuring
    full-horizon \emph{Long Diff-through-Opt} as described in
    Appendix~\ref{sec:LongDiffThroughOptStandalone}.}
  \label{fig:LongDiffThroughOptStandaloneResults}
\end{figure*}

\begin{table*}[h]
  \centering
  \caption{Quantitative analysis of UCI and Kin8nm final test performance from
    Figure~\ref{fig:LongDiffThroughOptStandaloneResults}, including full-horizon
    \emph{Long Diff-through-Opt} on shorter problems as described in
    Appendix~\ref{sec:LongDiffThroughOptStandalone}.
    Bold values are the lowest in class.}\label{tab:LongDiffThroughOptStandalone_UCIResults}
  \resizebox{\linewidth}{!}{
  \begin{tabular}{
    c
    S[table-format=2.0]
    U
    S[table-format=2.1]
    U
    S[table-format=1.3]
    S[table-format=2.0]
    U
    S[table-format=2.0]
    U
    S[table-format=2.2]
    S[table-format=3.0]
    U
    S[table-format=2.0]
    U
    S[table-format=2.1]}
    \toprule
    \multirow{2}{*}[\extrarulespace]{Method}
    & \multicolumn{5}{c}{UCI Energy}
    & \multicolumn{5}{c}{Kin8nm ($\times 1\,000$)}
    & \multicolumn{5}{c}{UCI Power} \\
    \cmidrule(lr){2-6} \cmidrule(lr){7-11} \cmidrule(lr){12-16}

    & \multicolumn{2}{c}{Mean} & \multicolumn{2}{c}{Median} & \multicolumn{1}{c}{Best}
    & \multicolumn{2}{c}{Mean} & \multicolumn{2}{c}{Median} & \multicolumn{1}{c}{Best}
    & \multicolumn{2}{c}{Mean} & \multicolumn{2}{c}{Median} & \multicolumn{1}{c}{Best} \\
    \midrule
    \input{Figures/AverageResults_UCI_LongDiffThroughOpt_Standalone.tex}
    \bottomrule
  \end{tabular}
  }
\end{table*}

\begin{table*}[h]
  \centering
  \caption{Quantitative analysis of Fashion-MNIST final test performance from
    Figure~\ref{fig:LongDiffThroughOptStandaloneResults}, including full-horizon
    \emph{Long Diff-through-Opt} on a shorter problem as described in
    Appendix~\ref{sec:LongDiffThroughOptStandalone}.
    Bold values are the lowest in class.}\label{tab:LongDiffThroughOptStandalone_FashionMNISTResults}
  \resizebox{\linewidth}{!}{
  \begin{tabular}{
    c
    S[table-format=1.3]
    U
    S[table-format=1.3]
    U
    S[table-format=1.3]
    S[table-format=1.3]
    U
    S[table-format=1.4]
    U
    S[table-format=1.3]}
    \toprule
    \multirow{2}{*}[\extrarulespace]{Method}
    & \multicolumn{5}{c}{Final Test Cross-Entropy} 
    & \multicolumn{5}{c}{Final Test Error} \\
    \cmidrule(lr){2-6} \cmidrule(lr){7-11}

    & \multicolumn{2}{c}{Mean} & \multicolumn{2}{c}{Median} & \multicolumn{1}{c}{Best}
    & \multicolumn{2}{c}{Mean} & \multicolumn{2}{c}{Median} & \multicolumn{1}{c}{Best} \\
    \midrule
    \input{Figures/AverageMixed_Fashion-MNIST_LongDiffThroughOpt_Standalone.tex}
    \bottomrule
  \end{tabular}
  }
\end{table*}

Firstly, we consider the four study datasets UCI Energy/Power, Kin8nm and
Fashion-MNIST, shortening training so it only lasts for 200 (UCI and Kin8nm) / 1000
(Fashion-MNIST) network weight update steps. This shorter training episode means
we cannot compare to our results in Section~\ref{sec:Experiments}, but does
permit full-horizon differentiation-through-optimisation. \emph{Long
  Diff-through-Opt} does exactly this, performing 30 (UCI and Kin8nm) / 50 (Fashion-MNIST)
hyperparameter updates using the maximum look-back horizon and restarting
training with the new hyperparameters after each update. In this sense, we use
30 or 50 times as much computation as in our other algorithms (here:
\emph{Random}, \emph{Ours$^\text{WD+LR+M}$} and \emph{Diff-through-Opt}). The latter
update their hyperparameters at the same intervals as before, giving 20 (UCI and
Kin8nm) /
100 (Fashion-MNIST) updates overall. We take 60 random hyperparameter
initialisations and run each of these four methods on them.

Our results are shown in Figure~\ref{fig:LongDiffThroughOptStandaloneResults}
and Tables~\ref{tab:LongDiffThroughOptStandalone_UCIResults} and
\ref{tab:LongDiffThroughOptStandalone_FashionMNISTResults}. They show a stark
contrast between datasets --- on UCI and Kin8nm trials, \emph{Long Diff-through-Opt}
manages, as expected, to achieve substantially improved final performance over
its short-horizon competition, but on Fashion-MNIST, performance is much more
similar between HPO algorithms. \citet{wuUnderstandingShortHorizonBias2018a}
intuit short-horizon bias by describing the simple setting of a quadratic with
different curvature magnitudes in each dimension, and we expect the simpler UCI/Kin8nm
datasets to match this clean depiction of the optimisation space more closely
than Fashion-MNIST. Thus, we speculate the space of non-trivial datasets, such
as Fashion-MNIST, is sufficiently complicated that merely eliminating
short-horizon bias is insufficient to prevent the optimiser from getting stuck
in similarly-performant local optima.

\clearpage
\subsubsection{Fashion-MNIST, Medium Horizon}
\label{sec:LongDiffThroughOptMedium}

\begin{figure*}[h]
  \centering
  \hspace*{0.08\textwidth}
  \begin{subfigure}{0.32\textwidth}
    \centering
    \includegraphics[width=\textwidth]{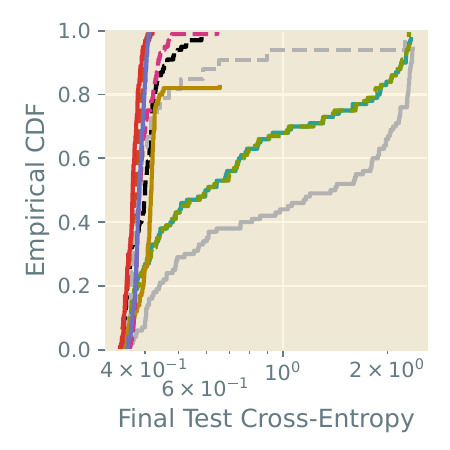}
    \caption{CDF (Loss)}
  \end{subfigure}
  \hfill
  \begin{subfigure}{0.32\textwidth}
    \centering
    \includegraphics[width=\textwidth]{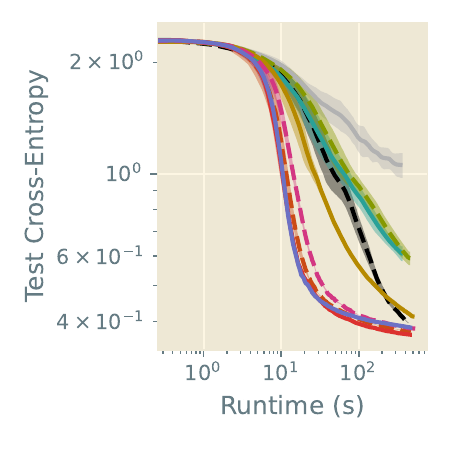}
    \caption{Evolution (Loss)}
  \end{subfigure}
  \hspace{0.08\textwidth}

  \hspace*{0.08\textwidth}
  \begin{subfigure}{0.32\textwidth}
    \centering
    \includegraphics[width=\textwidth]{Figures/CDF_Error_Fashion-MNIST_LongDiffThroughOpt_Standalone.pdf}
    \caption{CDF (Error)}
  \end{subfigure}
  \hspace*{0.08\textwidth}
  \begin{subfigure}{0.5\textwidth}
    \centering
    \begin{tabular}{cl}
      \linepatch{Cyan} & Random \\
      \linepatch{Grey} & Random ($\times$ LR) \\
      \linepatch[dashed]{Green} & Lorraine \\
      \linepatch{Yellow} & Baydin \\
      \linepatch[dashed]{Orange} & Ours$^\text{WD+LR}$ \\
      \linepatch{Red} & Ours$^\text{WD+LR+M}$ \\
      \linepatch[dashed]{Magenta} & Ours$^\text{WD+HDLR+M}$ \\
      \linepatch{Violet} & Diff-through-Opt \\
      \linepatch[dashed]{Black} & \makecell[l]{Medium Diff-through-Opt} \\[1ex]
      \envelopekey & $\pm$ Standard Error
    \end{tabular}
  \end{subfigure}
  \caption{CDFs of final test losses and evolutions of bootstrapped median
    losses from 100 random initialisations on Fashion-MNIST, featuring
    \emph{Medium Diff-through-Opt} with a 200-step look-back horizon as
    described in Appendix~\ref{sec:LongDiffThroughOptMedium}.}
  \label{fig:LongDiffThroughOptMediumResults}
\end{figure*}

\begin{table*}[h]
  \centering
  \caption{Quantitative analysis of Fashion-MNIST final test performance from
    Figure~\ref{fig:LongDiffThroughOptMediumResults}, including 
    \emph{Medium Diff-through-Opt} with a 200-step look-back horizon as described in
    Appendix~\ref{sec:LongDiffThroughOptMedium}.
    Bold values are the lowest in class.}\label{tab:LongDiffThroughOptMediumResults}
  \resizebox{\linewidth}{!}{
  \begin{tabular}{
    c
    S[table-format=1.3]
    U
    S[table-format=1.3]
    U
    S[table-format=1.3]
    S[table-format=1.4]
    U
    S[table-format=1.4]
    U
    S[table-format=1.3]}
    \toprule
    \multirow{2}{*}[\extrarulespace]{Method}
    & \multicolumn{5}{c}{Final Test Cross-Entropy} 
    & \multicolumn{5}{c}{Final Test Error} \\
    \cmidrule(lr){2-6} \cmidrule(lr){7-11}

    & \multicolumn{2}{c}{Mean} & \multicolumn{2}{c}{Median} & \multicolumn{1}{c}{Best}
    & \multicolumn{2}{c}{Mean} & \multicolumn{2}{c}{Median} & \multicolumn{1}{c}{Best} \\
    \midrule
    \input{Figures/AverageMixed_Fashion-MNIST_LongDiffThroughOpt_Medium.tex}
    \bottomrule
  \end{tabular}
  }
\end{table*}

Next, we note our Fashion-MNIST experiments of Section~\ref{sec:Experiments} run
for 10\,000 network weight steps, which can accommodate a longer look-back
horizon for the purposes of our study. We introduce a new algorithm,
\emph{Medium Diff-through-Opt}, which is identical to our previous Fashion-MNIST
\emph{Diff-through-Opt} configuration, except the look-back horizon is now 200
steps, and the number of hyperparameter updates reduced to 50 to keep the same
computational burden. This setting explores the trade-off between hypergradient
accuracy and number of hypergradients computed. We apply this new configuration
to the same 100 iterations as before, plotting combined results in
Figure~\ref{fig:LongDiffThroughOptMediumResults} and
Table~\ref{tab:LongDiffThroughOptMediumResults}. 

Again, we see that Fashion-MNIST seems resilient to the short-horizon bias, with
this medium-horizon configuration unable to substantially improve on our
existing results. Naturally, since \emph{Medium Diff-through-Opt} performs fewer
hyperparameter updates, its performance may suffer if the hyperparameters have
not converged, which we expect to be the likely cause here. However, it is
reassuring that our methods are capable of performing well despite their
short-horizon foundations.

\clearpage
\subsubsection{UCI Energy, Full Horizon}
\label{sec:LongDiffThroughOptFull}

\begin{figure*}[h]
  \centering
  \hfill
  \begin{subfigure}{0.32\textwidth}
    \centering
    \includegraphics[width=\textwidth]{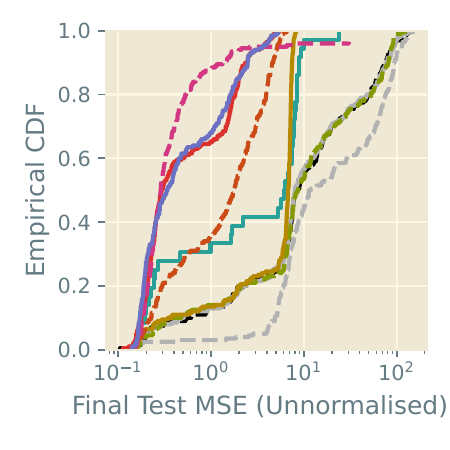}
    \caption{CDF (Loss)}
  \end{subfigure}
  \hfill
  \begin{subfigure}{0.32\textwidth}
    \centering
    \includegraphics[width=\textwidth]{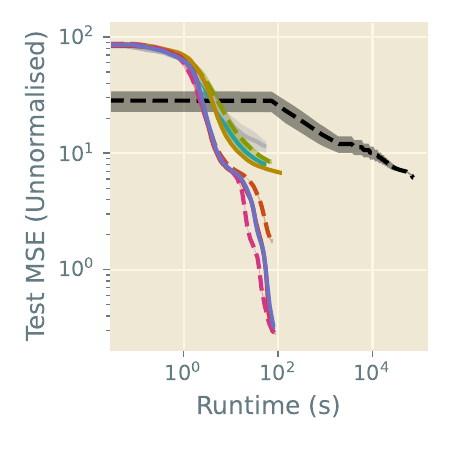}
    \caption{Evolution (Loss)}
  \end{subfigure}
  \hfill
  \begin{subfigure}{0.32\textwidth}
    \begin{tabular}{cl}
      \linepatch{Cyan} & Random \\
      \linepatch{Grey} & Random ($\times$ LR) \\
      \linepatch[dashed]{Green} & Lorraine \\
      \linepatch{Yellow} & Baydin \\
      \linepatch[dashed]{Orange} & Ours$^\text{WD+LR}$ \\
      \linepatch{Red} & Ours$^\text{WD+LR+M}$ \\
      \linepatch[dashed]{Magenta} & Ours$^\text{WD+HDLR+M}$ \\
      \linepatch{Violet} & Diff-through-Opt \\
      \linepatch[dashed]{Black} & \makecell[l]{Full Diff-through-Opt} \\[1ex]
      \envelopekey & $\pm$ Standard Error
    \end{tabular}
  \end{subfigure}
  \caption{CDF of final test losses and evolution of bootstrapped median
    losses from 200 random initialisations on UCI Energy, featuring
    36 initialisations of \emph{Full Diff-through-Opt} with a 4\,000-step look-back horizon as
    described in Appendix~\ref{sec:LongDiffThroughOptFull}.}
  \label{fig:LongDiffThroughOptFullResults}
\end{figure*}

\begin{table*}[h]
  \centering
  \caption{Quantitative analysis of UCI Energy final test performance from
    Figure~\ref{fig:LongDiffThroughOptFullResults}, including 
    \emph{Full Diff-through-Opt} with a 4\,000-step look-back horizon as described in
    Appendix~\ref{sec:LongDiffThroughOptFull}.
    Bold values are the lowest in class.}\label{tab:LongDiffThroughOptFullResults}
  \begin{tabular}{
    c
    S[table-format=1.3]
    U
    S[table-format=1.3]
    U
    S[table-format=1.3]}
    \toprule
    \multirow{2}{*}[\extrarulespace]{Method}
    & \multicolumn{5}{c}{Final Test MSE} \\
    \cmidrule(lr){2-6} 

    & \multicolumn{2}{c}{Mean} & \multicolumn{2}{c}{Median} & \multicolumn{1}{c}{Best} \\
    \midrule
    \input{Figures/AverageResults_UCI_Energy_LongDiffThroughOpt_Full.tex}
    \bottomrule
  \end{tabular}
\end{table*}

Finally, we consider UCI Energy, whose size is more amenable to full-horizon
studies. Our new setting here, \emph{Full Diff-through-Opt}, extends the
\emph{Diff-through-Opt} configuration from Section~\ref{sec:Experiments} by
using a full look-back horizon of 4\,000 steps, restarting training after each
hyperparameter update and performing 30 such updates. This permits a
full-horizon study on the same problem as before, using a similar computational
footprint to our Bayesian Optimisation trials of Appendix~\ref{sec:BayesianOptimisationBaselines}
(that is, 30 times the cost of our original experiments).
We consider 36 random initialisations for this algorithm, and combine the
results with our original UCI Energy experiments, showing both in
Figure~\ref{fig:LongDiffThroughOptFullResults} and
Table~\ref{tab:LongDiffThroughOptFullResults}.

Curiously, while \emph{Full Diff-through-Opt} outclasses naïve \emph{Random}
approaches, it is very uncompetitive with its short-horizon counterparts,
achieving much higher average final losses (the larger variance being, of
course, expected due to the smaller number of repetitions). The principal
difference between \emph{Full Diff-through-Opt} and \emph{Diff-through-Opt},
apart from the different look-back distances, is the number of hyperparameter
updates performed, since these are far more expensive in the former case. While
more hyperparameter updates would likely cause \emph{Full Diff-through-Opt}'s
performance to improve, the computational impact would be massive, with such a
setting unlikely to be desired in practical applications. Again, we conclude that
it is simplistic to claim short-horizon bias renders an algorithm ill-suited to
HPO --- in reality, it is one of many factors worthy of consideration in the
selection of an approach. In this case, the larger number of updates we are able
to perform by sacrificing horizon length leads to favourable performance
improvements, even though this optimisation is technically `biased' as a result.

\clearpage
\subsection{UCI Energy: Further Analysis}
In light of our main results, we perform several additional experiments to
investigate particular aspects of our algorithm. These are conducted on the
UCI~Energy dataset, using the same configuration as in
Section~\ref{sec:UCIExperiments} except where otherwise noted.

\subsubsection{Validation Dataset Size}

\begin{table*}
  \centerfigure
  \caption{Final test MSEs after training UCI Energy as in
    Table~\ref{tab:UCIResults}, with dataset splits modified to the indicated
    training/validation/test proportions. Uncertainties are standard errors, and
    the values may be directly compared to Table~\ref{tab:UCIResults}.
  }\label{tab:ValidationProportionResults}
  \resizebox{\linewidth}{!}{
    \begin{tabular}{
    c
    S[table-format=1.2]
    U
    S[table-format=1.2]
    U
    S[table-format=1.3]
    S[table-format=1.2]
    U
    S[table-format=1.3]
    U
    S[table-format=1.3]
    S[table-format=1.2]
    U
    S[table-format=1.3]
    U
    S[table-format=1.3]
    S[table-format=1.2]
    U
    S[table-format=1.2]
    U
    S[table-format=1.3]
    }
    \toprule
    \multirow{2}{*}[\extrarulespace]{Method}
    & \multicolumn{5}{c}{Default (80\%/10\%/10\%)}
    & \multicolumn{5}{c}{67.5\%/22.5\%/10\%}
    & \multicolumn{5}{c}{56.25\%/33.75\%/10\%}
    & \multicolumn{5}{c}{45\%/45\%/10\%} \\
    \cmidrule(lr){2-6} \cmidrule(lr){7-11} \cmidrule(lr){12-16} \cmidrule(lr){17-21}

    & \multicolumn{2}{c}{Mean} & \multicolumn{2}{c}{Median} & \multicolumn{1}{c}{Best}
    & \multicolumn{2}{c}{Mean} & \multicolumn{2}{c}{Median} & \multicolumn{1}{c}{Best}
    & \multicolumn{2}{c}{Mean} & \multicolumn{2}{c}{Median} & \multicolumn{1}{c}{Best}
    & \multicolumn{2}{c}{Mean} & \multicolumn{2}{c}{Median} & \multicolumn{1}{c}{Best} \\
    \midrule
    \input{Figures/AverageResults_UCI_Energy_ValidationProportions.tex}
    \bottomrule
  \end{tabular}
  }
\end{table*}

Noting that results for \emph{Ours$^\text{WD+HDLR+M}$} did not always
outperform \emph{Ours$^\text{WD+LR+M}$} as might have been expected, and
considering the risk of overfitting to validation data, we
consider the role the validation dataset might have to play on final test performance.
The conventional training/validation/test dataset split sizes we use throughout
this paper produce a training set which is substantially larger than validation
or test. In \emph{Our$^\text{WD+HDLR+M}$} case, we have approximately the same
number of hyperparameters as model weights, which is not the case in classical
training. Thus, we posit that choosing a larger validation dataset may
permit improved hyperparameter tuning, addressing this performance gap.

In Table~\ref{tab:ValidationProportionResults} we show results over 200 random
initialisations, with bootstrapped error bars computed as in
Section~\ref{sec:UCIExperiments}. We fix the test set size at our default
10\% of the available data (per
\citet{galDropoutBayesianApproximation2016}), and vary the balance between
training and validation data as shown.

Both the scalar and high-dimensional settings clearly benefit from a larger
validation dataset, with a 67.5\%/22.5\%/10\% split significantly improving
performance. But \emph{Ours$^\text{WD+HDLR+M}$} seems to exhibit less reliable,
higher-variance behaviour at even larger validation splits, with
\emph{Ours$^\text{WD+LR+M}$} also suffering worse performance in these settings.
The differences between best-case performance in these algorithms suggest there
is yet more to be understood about \emph{Our$^\text{WD+HDLR+M}$} setting.
However, it is clear that with larger numbers of hyperparameters, we must take
care to balance the competing demands for training and validation data.

\clearpage
\subsubsection{Asynchronous Hyperband and Population-Based Training}
\label{sec:ASHAPBTExperiments}

\begin{figure*}
  \centering
  \begin{subfigure}{0.48\textwidth}
    \centering
    \includegraphics[width=\textwidth]{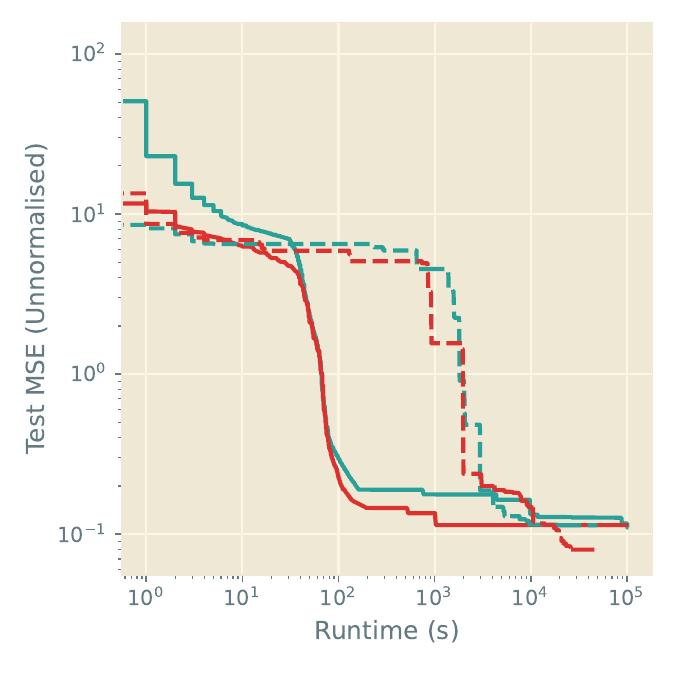}
  \end{subfigure}
  \hfill
  \begin{subfigure}{0.5\textwidth}
    \centering
    \begin{tabular}{cl}
      \linepatch{Cyan} & ASHA + Random \\
      \linepatch{Red} & ASHA + Ours$^\text{WD+LR+M}$ \\
      \linepatch[dashed]{Cyan} & PBT + Random \\
      \linepatch[dashed]{Red} & PBT + Ours$^\text{WD+LR+M}$ \\
    \end{tabular}
  \end{subfigure}
  \caption{Evolution of incumbent best test MSE with time, during training of
    UCI~Energy on ASHA (114\,286 initialisations) and PBT (200 training paths),
    with the threshold for `fully-trained' set at 4\,000 iterations. Note the
    logarithmic axes.}
  \label{fig:ASHAPBTResults}
\end{figure*}

\begin{table*}
  \centerfigure
  \caption{Extended reprise of Table~\ref{tab:UCIResults}: final test MSEs after
    training UCI~Energy for 4\,000 iterations, averaged over 200
    initialisations. Also shown are statistics for the initialisations which survive to
    full training on ASHA (out of 114\,286 considered) and the final
    models produced by PBT (out of 200 training paths). Uncertainties are
    standard errors; bold values lie in the error bars of the best algorithm.
  }\label{tab:ASHAPBTResults}
    \begin{tabular}{
    c
    S[table-format=1.3]
    U
    S[table-format=1.4]
    U
    S[table-format=1.4]
    }
    \toprule
    \multirow{2}{*}[\extrarulespace]{Method}
    & \multicolumn{5}{c}{UCI Energy} \\
    \cmidrule(lr){2-6}

    & \multicolumn{2}{c}{Mean} & \multicolumn{2}{c}{Median} & \multicolumn{1}{c}{Best} \\
    \midrule
    \input{Figures/AverageResults_uci_energy.tex}
    \midrule
    \input{Figures/AverageResults_ASHA.tex}
    \midrule
    \input{Figures/AverageResults_PBT.tex}
    \bottomrule
  \end{tabular}
\end{table*}

Finally, we consider two further non-gradient-based HPO strategies: Asynchronous
Hyperband (ASHA; \citet{liSystemMassivelyParallel2020}) and Population-Based
Training (PBT; \citet{jaderbergPopulationBasedTraining2017}). Both approaches
exploit parallel computation to consider a population of hyperparameter
selections, and the performance of each individual selection during
training is used to focus computational effort on the most promising trials.
Broadly speaking, ASHA does this by terminating poorly-performing trials early, while
PBT locally perturbs trials or replaces them entirely with better-performing
checkpoints. In this sense, ASHA and PBT are orthogonal to the type of algorithm we
propose in this paper, which seeks to improve the performance obtained by single
initialisations trained in isolation. However, our algorithm can readily be
combined with ASHA or PBT: by training each proposed hyperparameter choice using our algorithm
instead of fixing the hyperparameters as is conventional, we exploit our method's benefits for each
individual run, thus providing ASHA/PBT with a stronger population of trials
from which to make optimisation decisions. We demonstrate this synergy in the
experiments of this section.

Throughout, we make use of the Ray Tune \citep{liawTuneResearchPlatform2018}
library's implementation of these strategies, utilising the default settings
unless noted, and consider the UCI~Energy dataset using the same setup as
in Section~\ref{sec:UCIExperiments}. For the \emph{Random}
setting, we do not combine the training and validation datasets, instead training
with the former and reserving the latter for the evaluation of different
initialisations by ASHA/PBT. For \emph{Our$^\text{WD+LR+M}$} setting, the validation set is
used both for gradient-based hyperparameter updates during training and for
configuration evaluation by ASHA/PBT. Otherwise, both settings are the same as in
Section~\ref{sec:Experiments}, and again we quote final losses on a held-out
test set, with 1\,000 full-dataset bootstrap samples used to compute standard
errors. We also leave the hyperparameter search space unchanged from before
(see Section~\ref{sec:ExperimentDetails}).

\paragraph{Asynchronous Hyperband}
For ASHA, we seek to match the $4\,000 \times 200$ weight update steps performed in
Section~\ref{sec:UCIExperiments}, with 4\,000 steps still required for full training.
Under the default settings of Ray Tune (no grace period, reduction factor 4, one
bracket), we obtain computational budgets of
$(1, 4, 16, 64, 256, 1\,024, 4\,000)$ iterations, with the number of initialisations
surviving to each budget step correspondingly decreasing by a factor of 4 each
time. Non-parallelised Hyperband \citep{liHyperbandNovelBanditbased2017} uses
the same total computational effort for each budget step; applying this fact to
the smallest computational budget gives that we must begin with
$\frac{4\,000 \times 200}{7 \times 1} \approx 114\,286$ initialisations for computational
comparability. While the asynchronous modifications of ASHA introduce additional
stochasticity, we assume the expected workload to be the same as the
non-parallelised case. We run eight workers in
parallel to mimic the experimental setting of our earlier studies.

From the results in Table~\ref{tab:ASHAPBTResults}, we see our hypothesis
largely confirmed. By virtue of curating the population of initialisations,
rather than considering each one in isolation, ASHA trivially gives improved
average performance over all earlier methods.
However, combining ASHA with \emph{Ours$^\text{WD+LR+M}$} yields a significantly
more performant population of trials on average, which leads to low-valued,
low-variance mean and median final test performance. In both ASHA settings, the
best final losses in the resulting populations are towards the lower end of
those found in our previous results, but do not beat the especially small loss
found by \emph{Random (3-batched)}. From this, we suppose that the methods we consider are
nearing the performance limits of the UCI~Energy dataset on this model when trained in this
way. But regardless, we clearly see that deploying our
algorithm in a particular context tends to improve average-case performance,
indicating a positive influence on training.

\paragraph{Population-Based Training}
For PBT, we again attempt to match the earlier computational burden by training 200
initialisations for 4\,000 iterations using 8-way parallelism, acknowledging
that PBT's checkpoint restoration will cause most runs to perform more than
4\,000 weight update steps. We set the perturbation interval to 100 steps, so
each initialisation has 40 opportunities to be mutated or reset by PBT. In
\emph{PBT + Ours$^\text{WD+LR+M}$}, it is unclear how to accommodate PBT's
hyperparameter mutations, as our algorithm will have updated the hyperparameters
PBT associates with each trial. Where we detect PBT has mutated the
hyperparameters, we compute the ratio between the new and old values as seen by
PBT, then apply that same ratio to mutate the true current value of the
hyperparameters.

Our PBT results (Table~\ref{tab:ASHAPBTResults}) largely continue the pattern we
have discussed. Combining our algorithm with PBT achieves substantial improvements
in mean and median performance, giving the lowest values seen in any of our
experiments on UCI~Energy. PBT uses additional internal machinery to manage the
population of hyperparameter initialisations, and we found this caused the total training time
to considerably increase over the non-population-based methods of
Section~\ref{sec:UCIExperiments}, so the performance improvements we see here
are perhaps unsurprising (see also Figure~\ref{fig:ASHAPBTResults} and our
discussion of runtimes below). However, in situations where we have already committed to the
additional cost of PBT, replacing the fixed-hyperparameter internal training
routine with our algorithm clearly results in a superior population of models,
improving the quality of the models available for downstream applications. Note
that the quality of the search over hyperparameter initialisations affects our
overall best loss found, so some variation in the Best column of
Table~\ref{tab:ASHAPBTResults} is to be expected.

\paragraph{Runtime Analysis}
Figure~\ref{fig:ASHAPBTResults} shows a runtime analysis of these results on
ASHA and PBT. We see clearly that the superior performance of
\emph{Ours$^\text{WD+LR+M}$} in isolated training of specific hyperparameter
initialisations causes ASHA to reach low losses faster. In PBT, there is little
difference in performance vs.\ runtime when \emph{Random} or
\emph{Ours$^\text{WD+LR+M}$} is used as the internal learning strategy.
Subjectively, this is likely because the overhead of saving checkpoints and
restarting from them dominated the overall cost of running PBT. In both
cases, PBT is substantially slower than ASHA to reach the most performant
losses. Note by comparison with Figure~\ref{fig:UCIEnergyRuntimes} that our
methods alone are competitive with ASHA when we compare the evolution of
incumbent best performance over time. Note also
that ASHA and PBT do not have access to the intermediate test losses plotted in
Figure~\ref{fig:ASHAPBTResults}, so the smallest test losses found are not
necessarily achievable as an output of these methods.

\clearpage
\section{Complete Derivation}
\label{sec:VerboseDerivations}
Here, we reprise the derivations of Section~\ref{sec:Derivations} in more
verbose form. Our subsections mirror those used previously.

\subsection{Implicit Function Theorem in Bilevel Optimisation}
Recall that we consider some model parameterised by parameters $\vec{w}$, which
we train to minimise a training loss $\mathcal{L}_{T}$. Our training algorithm
is governed by hyperparameters $\vec{\lambda}$, which we simultaneously train to
minimise a validation loss $\mathcal{L}_{V}$. We target optimal
$\vec{w}^{*}$ and $\vec{\lambda}^{*}$, defined by the \emph{bilevel optimisation problem}
\begin{align}
  ^\text{(a)}\  \vec{\lambda}^{*} = \argmin_{\vec{\lambda}} \mathcal{L}_{V}(\vec{\lambda}, \vec{w}^{*}(\vec{\lambda})) \punc{,}
  && \text{such that} &&
  ^\text{(b)}\  \vec{w}^{*}(\vec{\lambda}) = \argmin_{\vec{w}} \mathcal{L}_{T} (\vec{\lambda}, \vec{w}) \punc{.}
  \relabel{eq:BilevelOptimisation}
\end{align}
(\ref{eq:BilevelOptimisation}b) simply reflects a standard model training
procedure, with fixed $\vec{\lambda}$, so we need only consider how to choose the
$\vec{\lambda}^{*}$ from (\ref{eq:BilevelOptimisation}a).

Suppose we seek a gradient-based hyperparameter optimisation method for
its performance potential. We then require the total derivative of the
validation objective $\mathcal{L}_{V}$ with respect to the hyperparameters
$\vec{\lambda}$, which may be obtained by the chain rule:
\begin{equation}
  \deriv{\mathcal{L}_{V}}{\vec{\lambda}} = \pderiv{\mathcal{L}_{V}}{\vec{\lambda}} + \pderiv{\mathcal{L}_{V}}{\vec{w}^{*}} \pderiv{\vec{w}^{*}}{\vec{\lambda}} \punc{.}
  \relabel{eq:TotalDerivative}
\end{equation}

An exact expression for $\pderiv{\vec{w}^{*}}{\vec{\lambda}}$ can be obtained by
applying Cauchy's Implicit Function Theorem
(Theorem~\ref{thm:ImplicitFunctionTheorem}). Suppose that for some $\vec{\lambda}'$ we
have identified a locally optimal $\vec{w}'$, and assume continuous
differentiability of $\pderiv{\mathcal{L}_{T}}{\vec{w}}$. Then, there exists an
open subset of hyperparameter space containing $\vec{\lambda}'$, over which we can
define a function $\vec{w}^{*}(\vec{\lambda})$ relating locally optimal $\vec{w}$ for
each $\vec{\lambda}$ by
\begin{equation}
  \pderiv{\vec{w}^{*}}{\vec{\lambda}} = - \left( \frac{\partial^{2} \mathcal{L}_{T}}{\partial \vec{w} \partial \vec{w}\trans} \right)^{-1} \frac{\partial^{2} \mathcal{L}_{T}}{\partial \vec{w} \partial \vec{\lambda}\trans} \punc{.}
  \relabel{eq:ImplicitFunctionTheorem}
\end{equation}
Of course, the inverse Hessian term in \eqref{eq:ImplicitFunctionTheorem}
becomes rapidly intractable for models at realistic scale.

\subsection{Approximate Best-Response Derivative}
\label{sec:AppendixApproximateBestResponseDerivative}
To proceed, we require an approximate expression for $\pderiv{\vec{w}^{*}}{\vec{\lambda}}$.
Suppose model training proceeds iteratively by the rule
\begin{equation}
  \vec{w}_{i}(\vec{\lambda}) = \vec{w}_{i-1}(\vec{\lambda}) - \vec{u}(\vec{\lambda}, \vec{w}_{i-1}(\vec{\lambda})) \punc{,}
  \relabel{eq:GeneralWeightUpdate}
\end{equation}
where $i$ is the current iteration and $\vec{u}$ is an arbitrary function
differentiable in both its arguments. Note that the trivial choice
\begin{equation}
  \vec{u}(\vec{\lambda}, \vec{w}_{i-1}(\vec{\lambda})) = \vec{w}_{i-1}(\vec{\lambda}) - \vec{f}(\vec{\lambda}, \vec{w}_{i-1}(\vec{\lambda}))
\end{equation}
allows arbitrary differentiable weight updates $\vec{f}$ to be subsumed by this
framework.

Differentiating with respect to $\vec{\lambda}$ at the current hyperparameters $\vec{\lambda}'$, we have:
\begin{align}
  \at{ \pderiv{\vec{w}_{i}}{\vec{\lambda}} }{\vec{\lambda}'}
  &= \at{\pderiv{\vec{w}_{i-1}}{\vec{\lambda}} }{\vec{\lambda}'} - \at{ \deriv{\vec{u}}{\vec{\lambda}} }{\vec{\lambda}', \vec{w}_{i-1}(\vec{\lambda}')} \\
  &= \at{\pderiv{\vec{w}_{i-1}}{\vec{\lambda}} }{\vec{\lambda}'} - \at{\pderiv{\vec{u}}{\vec{w}} \pderiv{\vec{w}_{i-1}}{\vec{\lambda}} + \pderiv{\vec{u}}{\vec{\lambda}} }{\vec{\lambda}', \vec{w}_{i-1}(\vec{\lambda}')} \\
  &= \at{ \left( \vec{I} - \pderiv{\vec{u}}{\vec{w}} \right) \pderiv{\vec{w}_{i-1}}{\vec{\lambda}} - \pderiv{\vec{u}}{\vec{\lambda}} }{\vec{\lambda}', \vec{w}_{i-1}(\vec{\lambda}')} \\
  &= - \at{ \pderiv{\vec{u}}{\vec{\lambda}} }{\vec{\lambda}', \vec{w}_{i-1}(\vec{\lambda}')} + \at{ \vec{I} - \pderiv{\vec{u}}{\vec{w}} }{\vec{\lambda}', \vec{w}_{i-1}(\vec{\lambda}')} \at{ \left( \vec{I} - \pderiv{\vec{u}}{\vec{w}} \right) \pderiv{\vec{w}_{i-2}}{\vec{\lambda}} - \pderiv{\vec{u}}{\vec{\lambda}} }{\vec{\lambda}', \vec{w}_{i-2}(\vec{\lambda}')} \\
  &= - \sum_{0 \leq j < i} \left( \left( \prod_{0 \leq k < j} \at{ \vec{I} - \pderiv{\vec{u}}{\vec{w}} }{\vec{\lambda}', \vec{w}_{i-1-k}(\vec{\lambda}')} \right) \at{ \pderiv{\vec{u}}{\vec{\lambda}} }{\vec{\lambda}', \vec{w}_{i-1-j}(\vec{\lambda}')} \right)
    \relabel{eq:UnrolledUpdate}
\end{align}

To combine terms in \eqref{eq:UnrolledUpdate}, we must make approximations
such that all bracketed expressions are evaluated at the same
$\vec{\lambda}, \vec{w}$. For network weights $\vec{w}$, we set
$\vec{w}_{0} = \vec{w}_{i} = \vec{w}^{*}$ for all iterations $i$, effectively
assuming we have been at the optimal $\vec{w}$ for some time; when we come to
consider finite truncations of the summation, this approximation will be
especially appealing. With this approximation, we may
collapse the product and write
\begin{align}
  \at{\pderiv{\vec{w}_{i}}{\vec{\lambda}}}{\vec{\lambda}'}
  &\approx - \sum_{0 \leq j < i} \left[ \left( \prod_{0 \leq k < j} \left( \vec{I} -
    \pderiv{\vec{u}}{\vec{w}} \right) \right)
     \pderiv{\vec{u}}{\vec{\lambda}}
    \right]_{\vec{\lambda}', \vec{w}_{i}(\vec{\lambda}')}\\
  &= - \at{ \sum_{0 \leq j < i} \left( \vec{I} - \pderiv{\vec{u}}{\vec{w}} \right)^{j} \pderiv{\vec{u}}{\vec{\lambda}}}{\vec{\lambda}', \vec{w}_{i}(\vec{\lambda}')} \punc{.}
\relabel{eq:BestResponseApproximation}
\end{align}

As $i$ increases, our training algorithm should converge towards a locally
optimal $\vec{w}^{*}$, and the summation in \eqref{eq:BestResponseApproximation}
will accrue more terms. Instead, we reinterpret $i$ as specifying the finite
number of terms we consider as an approximation to the whole sum. With this
redefined $i$, we arrive at our approximate best-response Jacobian
\begin{equation}
  \at{\pderiv{\vec{w}^{*}}{\vec{\lambda}}}{\vec{\lambda}'}
  \approx - \at{ \sum_{0 \leq j < i} \left( \vec{I} - \pderiv{\vec{u}}{\vec{w}} \right)^{j} \pderiv{\vec{u}}{\vec{\lambda}}}{\vec{\lambda}', \vec{w}^{*}(\vec{\lambda}')} \punc{.}
  \relabel{eq:BestResponseApproximation}
\end{equation}

\subsection{Convergence to Best-Response Derivative}
To informally argue the convergence properties of our method, note that
\eqref{eq:BestResponseApproximation} is a truncated Neumann series:
\begin{equation}
  \at{ \pderiv{\vec{w}_{i}}{\vec{\lambda}} }{\vec{\lambda}'}
  \approx - \at{ \sum_{0 \leq j < i} \left( \vec{I} - \pderiv{\vec{u}}{\vec{w}} \right)^j \pderiv{\vec{u}}{\vec{\lambda}} }{\vec{\lambda}', \vec{w}^*(\vec{\lambda}')} \punc{.}
\end{equation}

Analogously to its scalar equivalent --- the geometric series --- this series
converges as $i \to \infty$ if the multiplicative term
$(\vec{I} - \pderiv{\vec{u}}{\vec{w}})$ is `contractive' in some sense. For our
purposes, it suffices to assume a Banach space and, for the operator norm
$\norm{\cdot}$, that $\norm{\vec{I} - \pderiv{\vec{u}}{\vec{w}}} < 1$. In this case,
we can directly apply the closed-form limit expression for the Neumann series:
\begin{align}
  \lim_{i \to \infty} \at{ \pderiv{\vec{w}_{i}}{\vec{\lambda}} }{\vec{\lambda}'}
  &\approx -\lim_{i \to \infty} \at{ \sum_{0 \leq j < i} \left( \vec{I} - \pderiv{\vec{u}}{\vec{w}} \right)^j \pderiv{\vec{u}}{\vec{\lambda}} }{\vec{\lambda}', \vec{w}^*(\vec{\lambda}')} \\
  &= - \at{ \left(\vec{I} - \left( \vec{I} - \pderiv{\vec{u}}{\vec{w}} \right) \right)^{-1} \pderiv{\vec{u}}{\vec{\lambda}} }{\vec{\lambda}', \vec{w}^*(\vec{\lambda}')} \\
  &= - \at{ \left( \pderiv{\vec{u}}{\vec{w}} \right)^{-1} \pderiv{\vec{u}}{\vec{\lambda}} }{\vec{\lambda}', \vec{w}^*(\vec{\lambda}')} \punc{.}
    \relabel{eq:InfiniteLimit}
\end{align}

Note that \eqref{eq:InfiniteLimit} is exactly the result of the Implicit
Function Theorem \eqref{eq:ImplicitFunctionTheorem} applied to our more general
weight update $\vec{u}$. Concretely, suppose for some $\vec{\lambda}''$ and $\vec{w}''$
that $\vec{u}(\vec{\lambda}'', \vec{w}'') = \vec{0}$; in words, that our training
process does not update the network weights at this point in
weight-hyperparameter space, and assume $\vec{u}$ is continuously differentiable
with invertible Jacobian. Then, over some subset of hyperparameter space
containing $\vec{\lambda}''$, we can say: $\vec{w}^{*}(\vec{\lambda})$ exists, we have
$\vec{u}(\vec{\lambda}, \vec{w}^{*}(\vec{\lambda})) = \vec{0}$ throughout the subspace and
$\pderiv{\vec{w}^{*}}{\vec{\lambda}}$ is given by \eqref{eq:InfiniteLimit} exactly.

\subsection{Summary of Differences from \citet{lorraineOptimizingMillionsHyperparameters2020}}
In their derivation, which ours closely parallels,
\citet{lorraineOptimizingMillionsHyperparameters2020} replace
\eqref{eq:GeneralWeightUpdate} with a hard-coded SGD update step; in our
notation, they fix
$\vec{u}(\vec{\lambda}, \vec{w}) = \vec{u}_\textrm{SGD}(\vec{\lambda}, \vec{w}) = \eta \pderiv{\mathcal{L}_{T}}{\vec{w}}$.
They then isolate the learning rate $\eta$, insisting
all hyperparameters $\vec{\lambda}$ must be dependent variables of the training
loss $\mathcal{L}_{T}$, before further differentiating the training
loss in their derivation. Consequently, their algorithm cannot optimise optimiser
hyperparameters such as learning rates and momentum factors, since these come
into play only \emph{after} the training loss has been computed and differentiated.
While a careful choice of transformed loss function may allow certain optimiser hyperparameters
to be handled by the framework of \citet{lorraineOptimizingMillionsHyperparameters2020} (e.g.\ 
setting $\overline{\mathcal{L}} = \eta \mathcal{L}_T$ to incorporate a learning rate),
this manual rederivation is far from trivial in more complex cases (e.g.\ to incorporate momentum).

Instead, we define a \emph{general} weight update step
\eqref{eq:GeneralWeightUpdate}, without assuming any particular form. We may
then differentiate the general update $\vec{u}$ directly,
encompassing any hyperparameters appearing within it but outside the training
loss, which \citet{lorraineOptimizingMillionsHyperparameters2020} would miss.
Crucially, this includes the hyperparameters of most gradient-based
optimisers, including SGD. Thus, we lift
\citeauthor{lorraineOptimizingMillionsHyperparameters2020}'s results from expressions
about $\pderiv{\mathcal{L}_{T}}{\vec{w}}$ to expressions about $\vec{u}$, then
update hyperparameters inaccessible to the former. Concretely: we
optimise weight decay coefficients, learning rates and momentum factors in
settings where \citeauthor{lorraineOptimizingMillionsHyperparameters2020}
can only manage the weight decay alone.

\end{document}

%% file: Figures/AverageResults_UCI.tex
Random & 24 & $\pm\,$2 & 8.3 & $\pm\,$0.7 & 0.124 & 36 & $\pm\,$2 & 30 & $\pm\,$2 & \bfseries 5.70 & 70 & $\pm\,$7 & 20 & $\pm\,$1 & \bfseries 15.4\\
Random ($\times$ LR) & 35 & $\pm\,$3 & 13 & $\pm\,$3 & 0.113 & 44 & $\pm\,$2 & 41 & $\pm\,$3 & 6.18 & 106 & $\pm\,$8 & 29 & $\pm\,$5 & 16.0\\
Random (3-batched) & 5 & $\pm\,$1 & 2 & $\pm\,$1 & \bfseries 0.104 & 15 & $\pm\,$1 & 10 & $\pm\,$1 & 6.24 & 17.9 & $\pm\,$0.4 & 17.0 & $\pm\,$0.1 & 15.5\\
Lorraine & 24 & $\pm\,$2 & 8.7 & $\pm\,$0.9 & 0.145 & 36 & $\pm\,$2 & 30 & $\pm\,$3 & 6.32 & 69 & $\pm\,$6 & 19.7 & $\pm\,$0.9 & 15.6\\
Baydin & 5.5 & $\pm\,$0.2 & 6.8 & $\pm\,$0.1 & 0.144 & 23.4 & $\pm\,$0.8 & 26 & $\pm\,$1 & 6.00 & 17.80 & $\pm\,$0.07 & 17.94 & $\pm\,$0.07 & 15.4\\
Ours$^\text{WD+LR}$ & 2.1 & $\pm\,$0.1 & 1.8 & $\pm\,$0.2 & 0.130 & \bfseries 8.1 & $\pm\,$0.2 & 7.60 & $\pm\,$0.07 & 6.21 & 17.26 & $\pm\,$0.02 & 17.21 & $\pm\,$0.01 & 15.6\\
Ours$^\text{WD+LR+M}$ & 0.96 & $\pm\,$0.08 & 0.30 & $\pm\,$0.03 & 0.112 & \bfseries 8.1 & $\pm\,$0.4 & \bfseries 7.10 & $\pm\,$0.08 & 5.90 & 17.19 & $\pm\,$0.01 & 17.192 & $\pm\,$0.004 & 15.6\\
Ours$^\text{WD+HDLR+M}$ & \bfseries 0.6 & $\pm\,$0.2 & \bfseries 0.28 & $\pm\,$0.01 & 0.139 & \bfseries 8.1 & $\pm\,$0.3 & 7.47 & $\pm\,$0.06 & 6.04 & \bfseries 16.69 & $\pm\,$0.03 & \bfseries 16.76 & $\pm\,$0.02 & 15.4\\
Diff-through-Opt & 0.93 & $\pm\,$0.08 & 0.34 & $\pm\,$0.04 & 0.132 & \bfseries 7.9 & $\pm\,$0.3 & \bfseries 7.13 & $\pm\,$0.07 & 5.88 & 17.15 & $\pm\,$0.02 & 17.184 & $\pm\,$0.006 & 15.5\\

%% file: Figures/AverageResults_LargeScale.tex
Random & 0.92 & $\pm\,$0.07 & 0.60 & $\pm\,$0.08 & 0.340 & 4700 & $\pm\,$700 & 3000 & $\pm\,$3000 & 169 & 1.90 & $\pm\,$0.05 & 1.9 & $\pm\,$0.1 & 0.819\\
Random ($\times$ LR) & 1.29 & $\pm\,$0.08 & 1.2 & $\pm\,$0.3 & 0.339 & 5700 & $\pm\,$600 & 7000 & $\pm\,$2000 & 200 & 1.77 & $\pm\,$0.05 & 1.70 & $\pm\,$0.09 & 0.847\\
Random (3-batched) & 0.8 & $\pm\,$0.1 & 0.49 & $\pm\,$0.09 & 0.349 & 1300 & $\pm\,$600 & 470 & $\pm\,$80 & 171 & 1.60 & $\pm\,$0.07 & 1.55 & $\pm\,$0.09 & 0.817\\
Lorraine & 0.95 & $\pm\,$0.07 & 0.61 & $\pm\,$0.08 & 0.343 & 4700 & $\pm\,$600 & 3000 & $\pm\,$3000 & 170 & 1.41 & $\pm\,$0.07 & 1.26 & $\pm\,$0.07 & \bfseries 0.688\\
Baydin & 0.409 & $\pm\,$0.004 & 0.413 & $\pm\,$0.002 & 0.337 & 4200 & $\pm\,$600 & 2000 & $\pm\,$2000 & 149 & 1.60 & $\pm\,$0.05 & 1.6 & $\pm\,$0.1 & 0.756\\
Ours$^\text{WD+LR}$ & 0.373 & $\pm\,$0.002 & 0.375 & $\pm\,$0.003 & \bfseries 0.336 & 360 & $\pm\,$20 & 360 & $\pm\,$30 & 139 & 1.17 & $\pm\,$0.02 & 1.170 & $\pm\,$0.005 & 0.715\\
Ours$^\text{WD+LR+M}$ & \bfseries 0.369 & $\pm\,$0.002 & \bfseries 0.369 & $\pm\,$0.001 & 0.339 & \bfseries 270 & $\pm\,$30 & \bfseries 210 & $\pm\,$50 & \bfseries 100 & \bfseries 1.13 & $\pm\,$0.02 & \bfseries 1.16 & $\pm\,$0.01 & 0.712\\
Ours$^\text{WD+HDLR+M}$ & 0.398 & $\pm\,$0.004 & 0.386 & $\pm\,$0.003 & 0.350 & 310 & $\pm\,$20 & 270 & $\pm\,$10 & 150 & 1.33 & $\pm\,$0.05 & 1.32 & $\pm\,$0.07 & 0.822\\
Diff-through-Opt & 0.386 & $\pm\,$0.001 & 0.385 & $\pm\,$0.002 & 0.355 & \bfseries 300 & $\pm\,$10 & 290 & $\pm\,$20 & 114 & 1.195 & $\pm\,$0.009 & 1.197 & $\pm\,$0.007 & 0.967\\

%% file: Figures/AverageResults_UCI_BayesOpt.tex
Bayesian Optimisation & 0.4 & $\pm\,$0.1 & 0.23 & $\pm\,$0.03 & 0.106 & 7.3 & $\pm\,$0.2 & 6.9 & $\pm\,$0.2 & 6.22 & 16.08 & $\pm\,$0.09 & 16.2 & $\pm\,$0.2 & 15.0\\

%% file: Figures/AverageResults_Fashion-MNIST_NonBayesOpt.tex
Random & 0.92 & $\pm\,$0.07 & 0.60 & $\pm\,$0.08 & 0.340\\
Random ($\times$ LR) & 1.29 & $\pm\,$0.08 & 1.2 & $\pm\,$0.3 & 0.339\\
Random (3-batched) & 0.8 & $\pm\,$0.1 & 0.49 & $\pm\,$0.09 & 0.349\\
Lorraine & 0.95 & $\pm\,$0.07 & 0.61 & $\pm\,$0.08 & 0.343\\
Baydin & 0.409 & $\pm\,$0.004 & 0.413 & $\pm\,$0.002 & 0.337\\
Ours$^\text{WD+LR}$ & 0.373 & $\pm\,$0.002 & 0.375 & $\pm\,$0.003 & \bfseries 0.336\\
Ours$^\text{WD+LR+M}$ & \bfseries 0.369 & $\pm\,$0.002 & \bfseries 0.369 & $\pm\,$0.001 & 0.339\\
Ours$^\text{WD+HDLR+M}$ & 0.398 & $\pm\,$0.004 & 0.386 & $\pm\,$0.003 & 0.350\\
Diff-through-Opt & 0.386 & $\pm\,$0.001 & 0.385 & $\pm\,$0.002 & 0.355\\

%% file: Figures/AverageResults_Fashion-MNIST_BayesOpt.tex
Bayesian Optimisation & 0.3501 & $\pm\,$0.0005 & 0.3505 & $\pm\,$0.0007 & 0.344\\

%% file: Figures/AverageErrors_LargeScale.tex
Random & 0.31 & $\pm\,$0.02 & 0.21 & $\pm\,$0.02 & 0.116 & 0.46 & $\pm\,$0.02 & 0.44 & $\pm\,$0.03 & 0.243\\
Random ($\times$ LR) & 0.46 & $\pm\,$0.03 & 0.37 & $\pm\,$0.06 & \bfseries 0.115 & 0.50 & $\pm\,$0.03 & 0.46 & $\pm\,$0.03 & 0.244\\
Random (3-batched) & 0.20 & $\pm\,$0.03 & 0.141 & $\pm\,$0.009 & 0.119 & 0.41 & $\pm\,$0.03 & \bfseries 0.39 & $\pm\,$0.07 & 0.245\\
Lorraine & 0.32 & $\pm\,$0.02 & 0.22 & $\pm\,$0.03 & 0.120 & 0.46 & $\pm\,$0.03 & \bfseries 0.36 & $\pm\,$0.04 & 0.206\\
Baydin & 0.27 & $\pm\,$0.03 & 0.1499 & $\pm\,$0.0008 & 0.117 & 0.40 & $\pm\,$0.02 & 0.41 & $\pm\,$0.02 & 0.220\\
Ours$^\text{WD+LR}$ & 0.1329 & $\pm\,$0.0007 & 0.133 & $\pm\,$0.001 & 0.118 & \bfseries 0.393 & $\pm\,$0.008 & \bfseries 0.399 & $\pm\,$0.004 & 0.207\\
Ours$^\text{WD+LR+M}$ & \bfseries 0.1318 & $\pm\,$0.0006 & \bfseries 0.1316 & $\pm\,$0.0006 & 0.119 & \bfseries 0.386 & $\pm\,$0.008 & \bfseries 0.396 & $\pm\,$0.005 & \bfseries 0.204\\
Ours$^\text{WD+HDLR+M}$ & 0.143 & $\pm\,$0.001 & 0.1384 & $\pm\,$0.0008 & 0.122 & 0.44 & $\pm\,$0.02 & 0.45 & $\pm\,$0.02 & 0.268\\
Diff-through-Opt & 0.1376 & $\pm\,$0.0005 & 0.1372 & $\pm\,$0.0008 & 0.125 & 0.420 & $\pm\,$0.004 & 0.422 & $\pm\,$0.004 & 0.331\\

%% file: Figures/AverageResults_Fashion-MNIST.tex
Random & 0.92 & $\pm\,$0.07 & 0.60 & $\pm\,$0.08 & 0.340 & 0.88 & $\pm\,$0.06 & 0.6 & $\pm\,$0.1 & \bfseries 0.333\\
Random ($\times$ LR) & 1.29 & $\pm\,$0.08 & 1.2 & $\pm\,$0.3 & 0.339 & 1.21 & $\pm\,$0.08 & 1.1 & $\pm\,$0.2 & 0.338\\
Random (3-batched) & 0.8 & $\pm\,$0.1 & 0.49 & $\pm\,$0.09 & 0.349 & 0.42 & $\pm\,$0.02 & \bfseries 0.37 & $\pm\,$0.01 & 0.340\\
Lorraine & 0.95 & $\pm\,$0.07 & 0.61 & $\pm\,$0.08 & 0.343 & 0.89 & $\pm\,$0.06 & 0.58 & $\pm\,$0.09 & 0.343\\
Baydin & 0.409 & $\pm\,$0.004 & 0.413 & $\pm\,$0.002 & 0.337 & 0.404 & $\pm\,$0.005 & 0.404 & $\pm\,$0.002 & 0.335\\
Ours$^\text{WD+LR}$ & 0.373 & $\pm\,$0.002 & 0.375 & $\pm\,$0.003 & \bfseries 0.336 & 0.380 & $\pm\,$0.002 & 0.381 & $\pm\,$0.004 & 0.338\\
Ours$^\text{WD+LR+M}$ & \bfseries 0.369 & $\pm\,$0.002 & \bfseries 0.369 & $\pm\,$0.001 & 0.339 & \bfseries 0.371 & $\pm\,$0.002 & \bfseries 0.368 & $\pm\,$0.003 & 0.337\\
Ours$^\text{WD+HDLR+M}$ & 0.398 & $\pm\,$0.004 & 0.386 & $\pm\,$0.003 & 0.350 & 0.41 & $\pm\,$0.01 & 0.388 & $\pm\,$0.002 & 0.348\\
Diff-through-Opt & 0.386 & $\pm\,$0.001 & 0.385 & $\pm\,$0.002 & 0.355 & 0.388 & $\pm\,$0.002 & 0.388 & $\pm\,$0.004 & 0.348\\

%% file: Figures/AverageResults_UCI_LongDiffThroughOpt_Standalone.tex
Random & 40 & $\pm\,$5 & 17 & $\pm\,$9 & 3.19 & 56 & $\pm\,$3 & 63 & $\pm\,$4 & 15.8 & 130 & $\pm\,$20 & 80 & $\pm\,$30 & 18.5\\
Ours$^\text{WD+LR+M}$ & 30 & $\pm\,$5 & 10 & $\pm\,$3 & 2.24 & 57 & $\pm\,$7 & 54 & $\pm\,$7 & 11.5 & 100 & $\pm\,$20 & 31 & $\pm\,$6 & 17.6\\
Diff-through-Opt & 30 & $\pm\,$4 & 10 & $\pm\,$3 & 3.76 & 52 & $\pm\,$4 & 53 & $\pm\,$6 & 12.9 & 100 & $\pm\,$10 & 33 & $\pm\,$6 & 17.6\\
Long Diff-through-Opt & \bfseries 14 & $\pm\,$3 & \bfseries 7.3 & $\pm\,$0.9 & \bfseries 0.206 & \bfseries 32 & $\pm\,$2 & \bfseries 31 & $\pm\,$3 & \bfseries 7.70 & \bfseries 49 & $\pm\,$8 & \bfseries 22 & $\pm\,$2 & \bfseries 15.9\\

%% file: Figures/AverageMixed_Fashion-MNIST_LongDiffThroughOpt_Standalone.tex
Random & 1.3 & $\pm\,$0.1 & 1.2 & $\pm\,$0.4 & 0.408 & 0.45 & $\pm\,$0.04 & 0.35 & $\pm\,$0.09 & 0.146\\
Ours$^\text{WD+LR+M}$ & \bfseries 0.443 & $\pm\,$0.009 & \bfseries 0.430 & $\pm\,$0.003 & 0.401 & \bfseries 0.156 & $\pm\,$0.002 & \bfseries 0.155 & $\pm\,$0.001 & 0.142\\
Diff-through-Opt & \bfseries 0.45 & $\pm\,$0.01 & 0.434 & $\pm\,$0.002 & 0.406 & 0.159 & $\pm\,$0.002 & 0.1565 & $\pm\,$0.0007 & 0.145\\
Long Diff-through-Opt & \bfseries 0.450 & $\pm\,$0.005 & 0.436 & $\pm\,$0.003 & \bfseries 0.398 & 0.19 & $\pm\,$0.02 & 0.1574 & $\pm\,$0.0009 & \bfseries 0.140\\

%% file: Figures/AverageMixed_Fashion-MNIST_LongDiffThroughOpt_Medium.tex
Random & 0.92 & $\pm\,$0.07 & 0.60 & $\pm\,$0.08 & 0.340 & 0.31 & $\pm\,$0.02 & 0.21 & $\pm\,$0.02 & 0.116\\
Random ($\times$ LR) & 1.29 & $\pm\,$0.08 & 1.2 & $\pm\,$0.3 & 0.339 & 0.46 & $\pm\,$0.03 & 0.37 & $\pm\,$0.06 & \bfseries 0.115\\
Random (3-batched) & 0.8 & $\pm\,$0.1 & 0.49 & $\pm\,$0.09 & 0.349 & 0.20 & $\pm\,$0.03 & 0.141 & $\pm\,$0.009 & 0.119\\
Lorraine & 0.95 & $\pm\,$0.07 & 0.61 & $\pm\,$0.08 & 0.343 & 0.32 & $\pm\,$0.02 & 0.22 & $\pm\,$0.03 & 0.120\\
Baydin & 0.409 & $\pm\,$0.004 & 0.413 & $\pm\,$0.002 & 0.337 & 0.27 & $\pm\,$0.03 & 0.1499 & $\pm\,$0.0008 & 0.117\\
Ours$^\text{WD+LR}$ & 0.373 & $\pm\,$0.002 & 0.375 & $\pm\,$0.003 & \bfseries 0.336 & 0.1329 & $\pm\,$0.0007 & 0.133 & $\pm\,$0.001 & 0.118\\
Ours$^\text{WD+LR+M}$ & \bfseries 0.369 & $\pm\,$0.002 & \bfseries 0.369 & $\pm\,$0.001 & 0.339 & \bfseries 0.1318 & $\pm\,$0.0006 & \bfseries 0.1316 & $\pm\,$0.0006 & 0.119\\
Ours$^\text{WD+HDLR+M}$ & 0.398 & $\pm\,$0.004 & 0.386 & $\pm\,$0.003 & 0.350 & 0.143 & $\pm\,$0.001 & 0.1384 & $\pm\,$0.0008 & 0.122\\
Diff-through-Opt & 0.386 & $\pm\,$0.001 & 0.385 & $\pm\,$0.002 & 0.355 & 0.1376 & $\pm\,$0.0005 & 0.1372 & $\pm\,$0.0008 & 0.125\\
Medium Diff-through-Opt & 0.400 & $\pm\,$0.005 & 0.399 & $\pm\,$0.004 & 0.338 & 0.16 & $\pm\,$0.01 & 0.141 & $\pm\,$0.002 & 0.116\\

%% file: Figures/AverageResults_UCI_Energy_LongDiffThroughOpt_Full.tex
Random & 24 & $\pm\,$2 & 8.3 & $\pm\,$0.7 & 0.124\\
Random ($\times$ LR) & 35 & $\pm\,$3 & 13 & $\pm\,$3 & 0.113\\
Random (3-batched) & 5 & $\pm\,$1 & 2 & $\pm\,$1 & \bfseries 0.104\\
Lorraine & 24 & $\pm\,$2 & 8.7 & $\pm\,$0.9 & 0.145\\
Baydin & 5.5 & $\pm\,$0.2 & 6.8 & $\pm\,$0.1 & 0.144\\
Ours$^\text{WD+LR}$ & 2.1 & $\pm\,$0.1 & 1.8 & $\pm\,$0.2 & 0.130\\
Ours$^\text{WD+LR+M}$ & 0.96 & $\pm\,$0.08 & 0.30 & $\pm\,$0.03 & 0.112\\
Ours$^\text{WD+HDLR+M}$ & \bfseries 0.6 & $\pm\,$0.2 & \bfseries 0.28 & $\pm\,$0.01 & 0.139\\
Diff-through-Opt & 0.93 & $\pm\,$0.08 & 0.34 & $\pm\,$0.04 & 0.132\\
Full Diff-through-Opt & 5.2 & $\pm\,$0.8 & 6 & $\pm\,$2 & 0.138\\

%% file: Figures/AverageResults_UCI_Energy_ValidationProportions.tex
Ours$^\text{WD+LR+M}$ & 0.96 & $\pm\,$0.08 & 0.30 & $\pm\,$0.03 & 0.112 & 0.81 & $\pm\,$0.09 & 0.34 & $\pm\,$0.03 & 0.127 & 1.01 & $\pm\,$0.08 & 0.47 & $\pm\,$0.05 & 0.184 & 1.04 & $\pm\,$0.09 & 0.52 & $\pm\,$0.03 & 0.204\\
Ours$^\text{WD+HDLR+M}$ & 0.6 & $\pm\,$0.2 & 0.28 & $\pm\,$0.01 & 0.139 & 0.35 & $\pm\,$0.05 & 0.234 & $\pm\,$0.005 & 0.143 & 0.8 & $\pm\,$0.5 & 0.260 & $\pm\,$0.006 & 0.160 & 0.9 & $\pm\,$0.5 & 0.33 & $\pm\,$0.01 & 0.181\\

%% file: Figures/AverageResults_uci_energy.tex
Random & 24 & $\pm\,$2 & 8.3 & $\pm\,$0.7 & 0.124\\
Random ($\times$ LR) & 35 & $\pm\,$3 & 13 & $\pm\,$3 & 0.113\\
Random (3-batched) & 5 & $\pm\,$1 & 2 & $\pm\,$1 & 0.104\\
Lorraine & 24 & $\pm\,$2 & 8.7 & $\pm\,$0.9 & 0.145\\
Baydin & 5.5 & $\pm\,$0.2 & 6.8 & $\pm\,$0.1 & 0.144\\
Ours$^\text{WD+LR}$ & 2.1 & $\pm\,$0.1 & 1.8 & $\pm\,$0.2 & 0.130\\
Ours$^\text{WD+LR+M}$ & 0.96 & $\pm\,$0.08 & 0.30 & $\pm\,$0.03 & 0.112\\
Ours$^\text{WD+HDLR+M}$ & 0.6 & $\pm\,$0.2 & 0.28 & $\pm\,$0.01 & 0.139\\
Diff-through-Opt & 0.93 & $\pm\,$0.08 & 0.34 & $\pm\,$0.04 & 0.132\\

%% file: Figures/AverageResults_ASHA.tex
ASHA + Random & 0.29 & $\pm\,$0.09 & 0.191 & $\pm\,$0.004 & 0.131\\
ASHA + Ours$^\text{WD+LR+M}$ & 0.160 & $\pm\,$0.004 & 0.162 & $\pm\,$0.003 & 0.104\\

%% file: Figures/AverageResults_PBT.tex
PBT + Random & 0.18 & $\pm\,$0.03 & 0.1328 & $\pm\,$0.0007 & \bfseries 0.0934\\
PBT + Ours$^\text{WD+LR+M}$ & \bfseries 0.134 & $\pm\,$0.005 & \bfseries 0.126 & $\pm\,$0.003 & 0.119\\